\documentclass{article}

\usepackage{microtype}
\usepackage{graphicx}
\usepackage{booktabs} 
\usepackage{multirow}

\usepackage{hyperref}

\usepackage[accepted]{icml2026}

\usepackage{amsmath}
\usepackage{amssymb}
\usepackage{mathtools}
\usepackage{amsthm}

\usepackage[capitalize,noabbrev]{cleveref}

\theoremstyle{plain}

\theoremstyle{definition}

\theoremstyle{remark}

\usepackage[textsize=tiny]{todonotes}

\usepackage{soul}
\usepackage{enumitem}
\usepackage{color-edits}

\addauthor{ss}{oc-orange-9}

\DeclareRobustCommand{\styledtext}[3]{\textcolor{#1}{\highlightcolor{#2}{#3}}}

\definecolor{oc-gray-0}{HTML}{F8F9FA}
\definecolor{oc-gray-1}{HTML}{F1F3F5}
\definecolor{oc-gray-2}{HTML}{E9ECEF}
\definecolor{oc-gray-3}{HTML}{DEE2E6}
\definecolor{oc-gray-4}{HTML}{CED4DA}
\definecolor{oc-gray-5}{HTML}{ADB5BD}
\definecolor{oc-gray-6}{HTML}{868E96}
\definecolor{oc-gray-7}{HTML}{495057}
\definecolor{oc-gray-8}{HTML}{343A40}
\definecolor{oc-gray-9}{HTML}{212529}

\definecolor{oc-red-0}{HTML}{FFF5F5}
\definecolor{oc-red-1}{HTML}{FFE3E3}
\definecolor{oc-red-2}{HTML}{FFC9C9}
\definecolor{oc-red-3}{HTML}{FFA8A8}
\definecolor{oc-red-4}{HTML}{FF8787}
\definecolor{oc-red-5}{HTML}{FF6B6B}
\definecolor{oc-red-6}{HTML}{FA5252}
\definecolor{oc-red-7}{HTML}{F03E3E}
\definecolor{oc-red-8}{HTML}{E03131}
\definecolor{oc-red-9}{HTML}{C92A2A}

\definecolor{ai2pink}{HTML}{E64980}

\definecolor{oc-pink-0}{HTML}{FFF0F6}
\definecolor{oc-pink-1}{HTML}{FFDEEB}
\definecolor{oc-pink-2}{HTML}{FCC2D7}
\definecolor{oc-pink-3}{HTML}{FAA2C1}
\definecolor{oc-pink-4}{HTML}{F783AC}
\definecolor{oc-pink-5}{HTML}{F06595}
\definecolor{oc-pink-6}{HTML}{E64980}
\definecolor{oc-pink-7}{HTML}{D6336C}
\definecolor{oc-pink-8}{HTML}{C2255C}
\definecolor{oc-pink-9}{HTML}{A61E4D}

\definecolor{oc-grape-0}{HTML}{F8F0FC}
\definecolor{oc-grape-1}{HTML}{F3D9FA}
\definecolor{oc-grape-2}{HTML}{EEBEFA}
\definecolor{oc-grape-3}{HTML}{E599F7}
\definecolor{oc-grape-4}{HTML}{DA77F2}
\definecolor{oc-grape-5}{HTML}{CC5DE8}
\definecolor{oc-grape-6}{HTML}{BE4BDB}
\definecolor{oc-grape-7}{HTML}{AE3EC9}
\definecolor{oc-grape-8}{HTML}{9C36B5}
\definecolor{oc-grape-9}{HTML}{862E9C}

\definecolor{oc-violet-0}{HTML}{F3F0FF}
\definecolor{oc-violet-1}{HTML}{E5DBFF}
\definecolor{oc-violet-2}{HTML}{D0BFFF}
\definecolor{oc-violet-3}{HTML}{B197FC}
\definecolor{oc-violet-4}{HTML}{9775FA}
\definecolor{oc-violet-5}{HTML}{845EF7}
\definecolor{oc-violet-6}{HTML}{7950F2}
\definecolor{oc-violet-7}{HTML}{7048E8}
\definecolor{oc-violet-8}{HTML}{6741D9}
\definecolor{oc-violet-9}{HTML}{5F3DC4}

\definecolor{oc-indigo-0}{HTML}{EDF2FF}
\definecolor{oc-indigo-1}{HTML}{DBE4FF}
\definecolor{oc-indigo-2}{HTML}{BAC8FF}
\definecolor{oc-indigo-3}{HTML}{91A7FF}
\definecolor{oc-indigo-4}{HTML}{748FFC}
\definecolor{oc-indigo-5}{HTML}{5C7CFA}
\definecolor{oc-indigo-6}{HTML}{4C6EF5}
\definecolor{oc-indigo-7}{HTML}{4263EB}
\definecolor{oc-indigo-8}{HTML}{3B5BDB}
\definecolor{oc-indigo-9}{HTML}{364FC7}

\definecolor{oc-blue-0}{HTML}{E7F5FF}
\definecolor{oc-blue-1}{HTML}{D0EBFF}
\definecolor{oc-blue-2}{HTML}{A5D8FF}
\definecolor{oc-blue-3}{HTML}{74C0FC}
\definecolor{oc-blue-4}{HTML}{4DABF7}
\definecolor{oc-blue-5}{HTML}{339AF0}
\definecolor{oc-blue-6}{HTML}{228BE6}
\definecolor{oc-blue-7}{HTML}{1C7ED6}
\definecolor{oc-blue-8}{HTML}{1971C2}
\definecolor{oc-blue-9}{HTML}{1864AB}

\definecolor{oc-cyan-0}{HTML}{E3FAFC}
\definecolor{oc-cyan-1}{HTML}{C5F6FA}
\definecolor{oc-cyan-2}{HTML}{99E9F2}
\definecolor{oc-cyan-3}{HTML}{66D9E8}
\definecolor{oc-cyan-4}{HTML}{3BC9DB}
\definecolor{oc-cyan-5}{HTML}{22B8CF}
\definecolor{oc-cyan-6}{HTML}{15AABF}
\definecolor{oc-cyan-7}{HTML}{1098AD}
\definecolor{oc-cyan-8}{HTML}{0C8599}
\definecolor{oc-cyan-9}{HTML}{0B7285}

\definecolor{oc-teal-0}{HTML}{E6FCF5}
\definecolor{oc-teal-1}{HTML}{C3FAE8}
\definecolor{oc-teal-2}{HTML}{96F2D7}
\definecolor{oc-teal-3}{HTML}{63E6BE}
\definecolor{oc-teal-4}{HTML}{38D9A9}
\definecolor{oc-teal-5}{HTML}{20C997}
\definecolor{oc-teal-6}{HTML}{12B886}
\definecolor{oc-teal-7}{HTML}{0CA678}
\definecolor{oc-teal-8}{HTML}{099268}
\definecolor{oc-teal-9}{HTML}{087F5B}

\definecolor{oc-green-0}{HTML}{EBFBEE}
\definecolor{oc-green-1}{HTML}{D3F9D8}
\definecolor{oc-green-2}{HTML}{B2F2BB}
\definecolor{oc-green-3}{HTML}{8CE99A}
\definecolor{oc-green-4}{HTML}{69DB7C}
\definecolor{oc-green-5}{HTML}{51CF66}
\definecolor{oc-green-6}{HTML}{40C057}
\definecolor{oc-green-7}{HTML}{37B24D}
\definecolor{oc-green-8}{HTML}{2F9E44}
\definecolor{oc-green-9}{HTML}{2B8A3E}

\definecolor{oc-lime-0}{HTML}{F4FCE3}
\definecolor{oc-lime-1}{HTML}{E9FAC8}
\definecolor{oc-lime-2}{HTML}{D8F5A2}
\definecolor{oc-lime-3}{HTML}{C0EB75}
\definecolor{oc-lime-4}{HTML}{A9E34B}
\definecolor{oc-lime-5}{HTML}{94D82D}
\definecolor{oc-lime-6}{HTML}{82C91E}
\definecolor{oc-lime-7}{HTML}{74B816}
\definecolor{oc-lime-8}{HTML}{66A80F}
\definecolor{oc-lime-9}{HTML}{5C940D}

\definecolor{oc-yellow-0}{HTML}{FFF9DB}
\definecolor{oc-yellow-1}{HTML}{FFF3BF}
\definecolor{oc-yellow-2}{HTML}{FFEC99}
\definecolor{oc-yellow-3}{HTML}{FFE066}
\definecolor{oc-yellow-4}{HTML}{FFD43B}
\definecolor{oc-yellow-5}{HTML}{FCC419}
\definecolor{oc-yellow-6}{HTML}{FAB005}
\definecolor{oc-yellow-7}{HTML}{F59F00}
\definecolor{oc-yellow-8}{HTML}{F08C00}
\definecolor{oc-yellow-9}{HTML}{E67700}

\definecolor{oc-orange-0}{HTML}{FFF4E6}
\definecolor{oc-orange-1}{HTML}{FFE8CC}
\definecolor{oc-orange-2}{HTML}{FFD8A8}
\definecolor{oc-orange-3}{HTML}{FFC078}
\definecolor{oc-orange-4}{HTML}{FFA94D}
\definecolor{oc-orange-5}{HTML}{FF922B}
\definecolor{oc-orange-6}{HTML}{FD7E14}
\definecolor{oc-orange-7}{HTML}{F76707}
\definecolor{oc-orange-8}{HTML}{E8590C}
\definecolor{oc-orange-9}{HTML}{D9480F}

\newcommand{\highlightcolor}[2]{%
  \sethlcolor{#1}%
  \hl{#2}%
}

\newlength{\myparskiplength}
\newlength{\myparindentlength}
\parindent = \myparindentlength
\newcommand{\modelname}{DR Tulu\xspace}

\definecolor{prompt}{HTML}{5f84e4}
\definecolor{img}{HTML}{820100}

\usepackage{fancyvrb}
\usepackage{fvextra}

\DefineVerbatimEnvironment{VerbatimWrap}{Verbatim}{
breaklines=true, 
breakindent=0pt, 
breaksymbol={},
fontsize=\small,
}

\usepackage{url}
\usepackage{tabularx}
\usepackage{color-edits}
\usepackage{comment}
\newtoggle{comment}
\usepackage{arydshln}
\usepackage[dvipsnames,svgnames,table]{xcolor} 
\definecolor{lightgray}{gray}{0.9} 
\usepackage{pifont}
\definecolor{darkgreen}{RGB}{0,100,0}  
\newcommand{\cmark}{\textcolor{darkgreen}{\ding{51}}}%
\newcommand{\xmark}{\textcolor{red}{\ding{55}}}%
\toggletrue{comment}

\usepackage{wrapfig}   
\usepackage{caption}   
\usepackage{subcaption}

\usepackage{float}                     
\usepackage{fontawesome5}              
\usepackage{amssymb} 

\newcommand{\heart}{\ensuremath\heartsuit} 

\renewcommand{\algorithmiccomment}[1]{\hfill\textcolor{teal}{\(\triangleright\) #1}}

\usepackage[most]{tcolorbox}
\usepackage{xspace}

\definecolor{pal-coral}{HTML}{F8807F}   
\definecolor{pal-apricot}{HTML}{FFBB74} 
\definecolor{pal-olivine}{HTML}{A5C476} 
\definecolor{pal-mint}{HTML}{B3EFCD}    
\definecolor{pal-aqua}{HTML}{81D7D4}    
\definecolor{Ai2Pink}{HTML}{F0529C}
\colorlet{Ai2PinkLight}{Ai2Pink!70!white}

\tcbset{
  token/.style={
    enhanced, online, nobeforeafter, sharp corners,
    boxsep=0pt, left=3pt, right=3pt, top=1pt, bottom=1pt,
    boxrule=0.4pt,
    tcbox raise base,
    valign=center
  }
}

\newcommand{\Think}{\styledtext{oc-cyan-8}{oc-cyan-0}{\texttt{ think }\xspace}}
\newcommand{\CallTool}{\styledtext{oc-pink-8}{oc-pink-0}{\texttt{ tool }\xspace}}
\newcommand{\Answer}{\styledtext{oc-yellow-8}{oc-yellow-0}{\texttt{ answer }\xspace}}
\newcommand{\Cite}{\styledtext{oc-lime-8}{oc-lime-0}{\texttt{ cite }\xspace}}

\newcommand{\ThinkTok}{\textcolor{oc-gray-7}{\texttt{<think></think>}}}
\newcommand{\CallToolTok}[1]{\textcolor{oc-gray-7}{\texttt{<call\_tool></call\_tool>}}}
\newcommand{\AnswerTok}{\textcolor{oc-gray-7}{\texttt{<answer></answer>}}}
\newcommand{\CiteTok}[1]{\textcolor{oc-gray-7}{\texttt{<cite id="#1"></cite>}}}

\newcommand{\opendata}[1][1em]{%
  \raisebox{-0.2\height}{\includegraphics[height=#1]{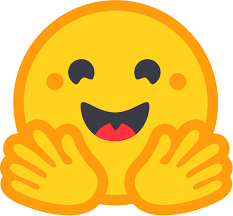}}%
}
\newcommand{\opencode}[1][1em]{%
  \raisebox{-0.2\height}{\includegraphics[height=#1]{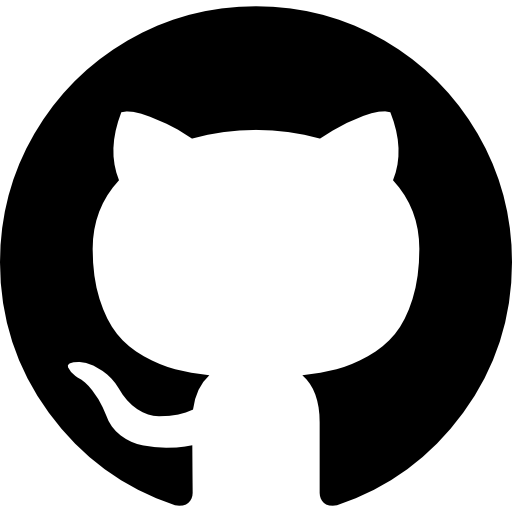}}%
}
\newcommand{\oai}[1][1em]{%
  \raisebox{-0.2\height}{\includegraphics[height=#1]{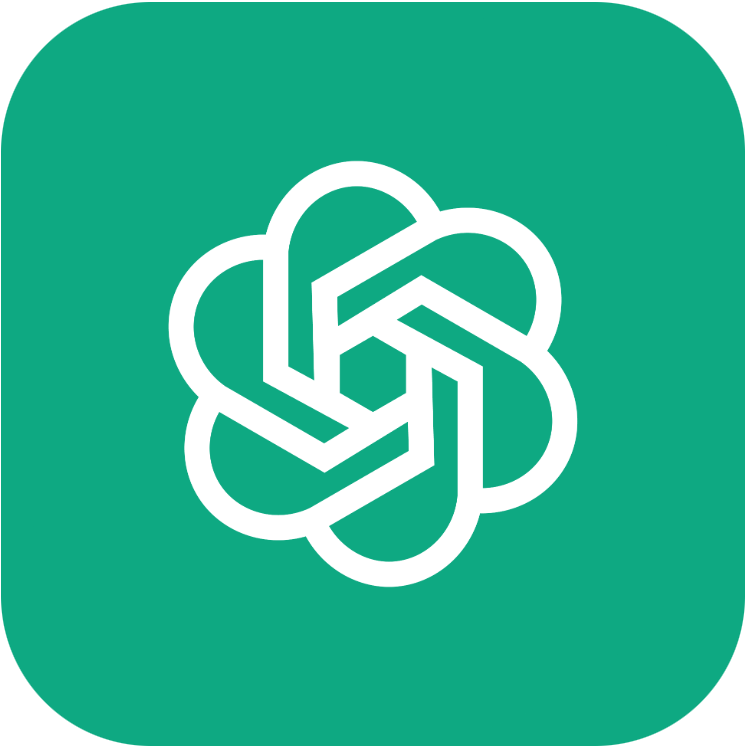}}%
}
\newcommand{\perplexity}[1][1em]{%
  \raisebox{-0.2\height}{\includegraphics[height=#1]{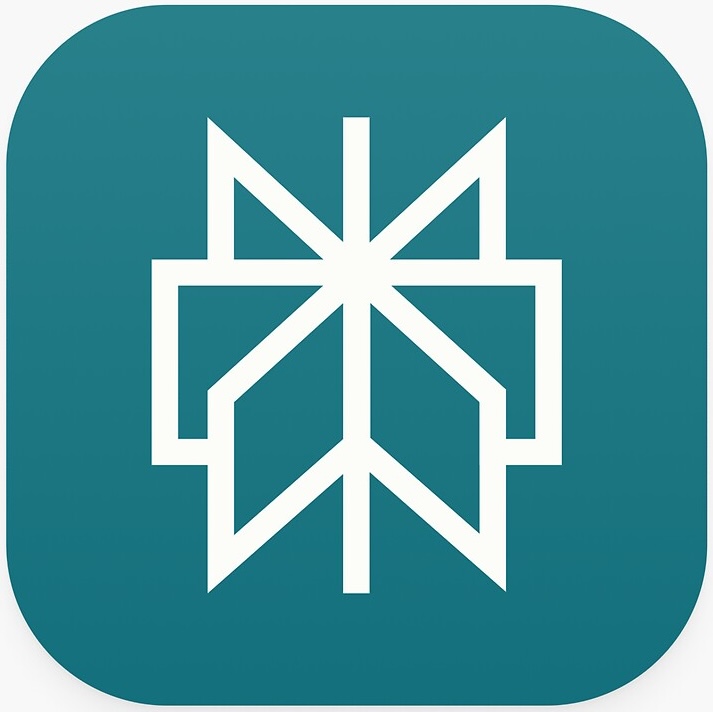}}%
}
\newcommand{\claude}[1][1em]{%
  \raisebox{-0.2\height}{\includegraphics[height=#1]{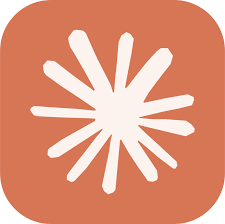}}%
}
\newcommand{\gemini}[1][1em]{%
  \raisebox{-0.2\height}{\includegraphics[height=#1]{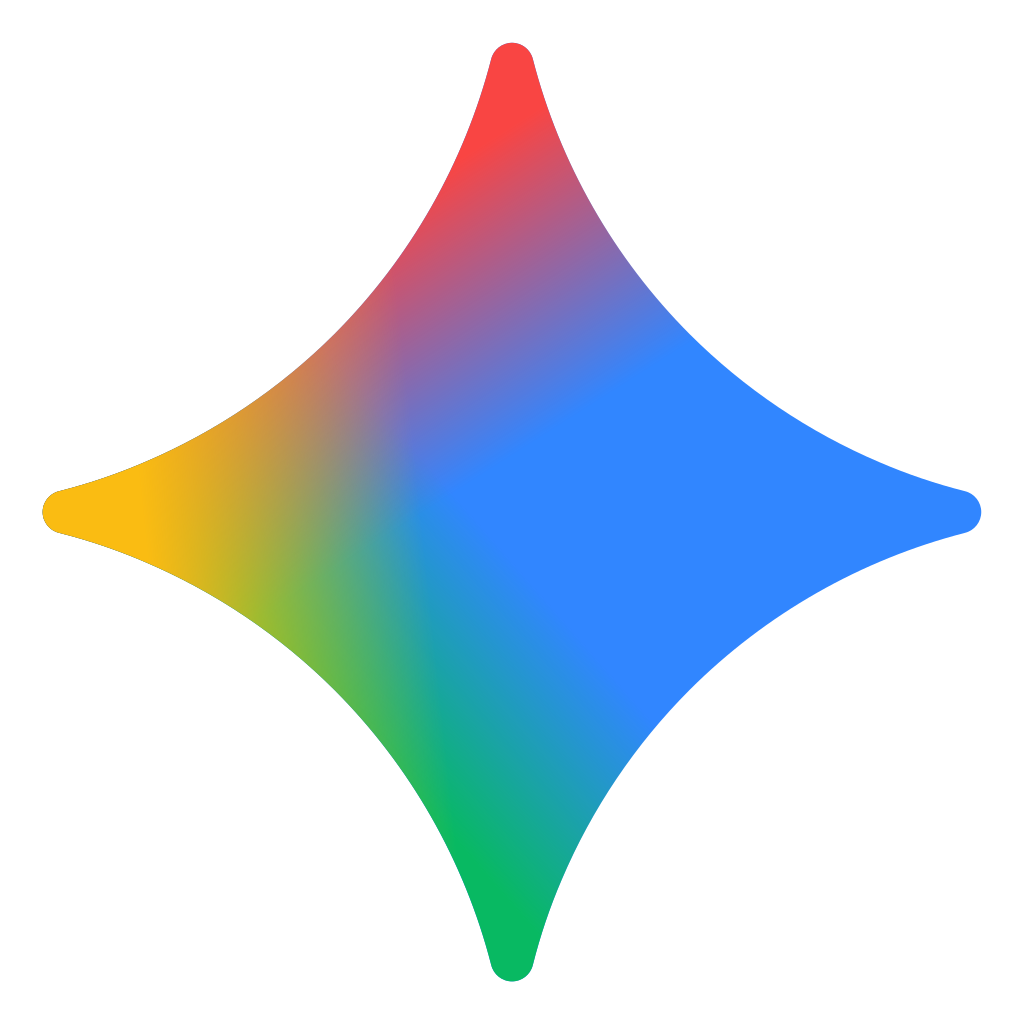}}%
}
\newcommand{\aitwo}[1][1em]{%
  \raisebox{-0.2\height}{\includegraphics[height=#1]{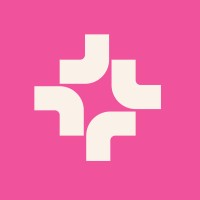}}%
}

\icmltitlerunning{DR Tulu: Reinforcement Learning with Evolving Rubrics for Deep Research}

\begin{document}

\twocolumn[
\DeclareRobustCommand{\TitlePNG}{%
  \raisebox{-0.2ex}{\includegraphics[height=1em]{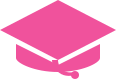}}%
}
\newif\ifpinktitle
\pinktitletrue
\icmltitle{\texorpdfstring{%
  \TitlePNG\hspace{0.45em}%
  \ifpinktitle{\color{Ai2Pink}DR\ Tulu: Reinforcement Learning with Evolving Rubrics for Deep Research}%
  \else DR\ Tulu: Reinforcement Learning with Evolving Rubrics for Deep Research\fi
}{DR Tulu: Reinforcement Learning with Evolving Rubrics for Deep Research}}



\icmlsetsymbol{equal}{\heart}
\icmlsetsymbol{core}{\ensuremath{\dagger}}

\begin{icmlauthorlist}
\icmlauthor{Rulin Shao}{equal,core,uw}
\icmlauthor{Akari Asai}{equal,core,ai2,cmu}
\icmlauthor{Shannon Zejiang Shen}{equal,core,mit}
\icmlauthor{Hamish Ivison}{equal,core,uw,ai2}\\
\icmlauthor{Varsha Kishore}{core,uw,ai2}
\icmlauthor{Jingming Zhuo}{core,uw}
\icmlauthor{Xinran Zhao}{cmu}
\icmlauthor{Molly Park}{uw}
\icmlauthor{Samuel G. Finlayson}{uw,sch}\\
\icmlauthor{David Sontag}{mit}
\icmlauthor{Tyler Murray}{ai2}
\icmlauthor{Sewon Min}{ai2,ucb}
\icmlauthor{Pradeep Dasigi}{ai2}
\icmlauthor{Luca Soldaini}{ai2}
\icmlauthor{Faeze Brahman}{ai2}\\
\icmlauthor{Wen-tau Yih}{uw}
\icmlauthor{Tongshuang Wu}{cmu}
\icmlauthor{Luke Zettlemoyer}{uw}
\icmlauthor{Yoon Kim}{mit}\\
\icmlauthor{Hannaneh Hajishirzi}{uw,ai2}
\icmlauthor{Pang Wei Koh}{uw,ai2}
\end{icmlauthorlist}

\icmlaffiliation{uw}{University of Washington}
\icmlaffiliation{ai2}{Allen Institute for AI}
\icmlaffiliation{cmu}{Carnegie Mellon University}
\icmlaffiliation{mit}{Massachusetts Institute of Technology}
\icmlaffiliation{sch}{Seattle Children's Hospital}
\icmlaffiliation{ucb}{University of California, Berkeley}

\icmlcorrespondingauthor{Rulin Shao}{rulins@cs.washington.edu}
\icmlcorrespondingauthor{Akari Asai}{akaria@allenai.org}

\icmlkeywords{Machine Learning, ICML}

\vskip 0.15in
\begin{center}
\small
\opencode~\href{https://github.com/rlresearch/dr-tulu}{Code}\quad
\opendata~\href{https://huggingface.co/collections/rl-research/dr-tulu}{Data \& Models}\quad
{\textcolor{Ai2Pink}{\faGlobe}}~\href{https://www.dr-tulu.org/}{Interactive Demo}
\end{center}

\vskip 0.15in
]



\printAffiliationsAndNotice{\textsuperscript{\heart}Joint first authors. \textsuperscript{\ensuremath{\dagger}}Core contributors. See full author contributions \protect\hyperref[sec:authors]{here}.}

\begin{abstract}
Deep research agents perform multi-step research to produce long-form, well-attributed answers. However, most open deep research agents are trained on easily verifiable short-form QA tasks via reinforcement learning with verifiable rewards, which does not extend to realistic long-form tasks. 
We address this with \textbf{Reinforcement Learning with Evolving Rubrics (RLER)}, where rubrics are constructed and maintained to \textit{co-evolve} with the policy model during training. This allows the rubrics to incorporate newly explored information from search and contrasting model responses, enabling better fact checking and more discriminative on-policy feedback.
Using RLER, we develop \textbf{Deep Research Tulu (DR Tulu-8B)}, the first fully open model that is directly trained for open-ended, long-form deep research.
Across four long-form deep research benchmarks in science, healthcare, and general domains, \modelname substantially outperforms existing open deep research agents (by 15.6\% over Tongyi DR on average) and matches or exceeds proprietary deep research agents (by 0.7\% over OpenAI DR on average), while being significantly smaller and cheaper per query ($ 1000\times$ cheaper than OpenAI DR per query).%
%

\end{abstract}

\begin{figure}[t]
    \centering
        \centering

        \includegraphics[width=\linewidth]{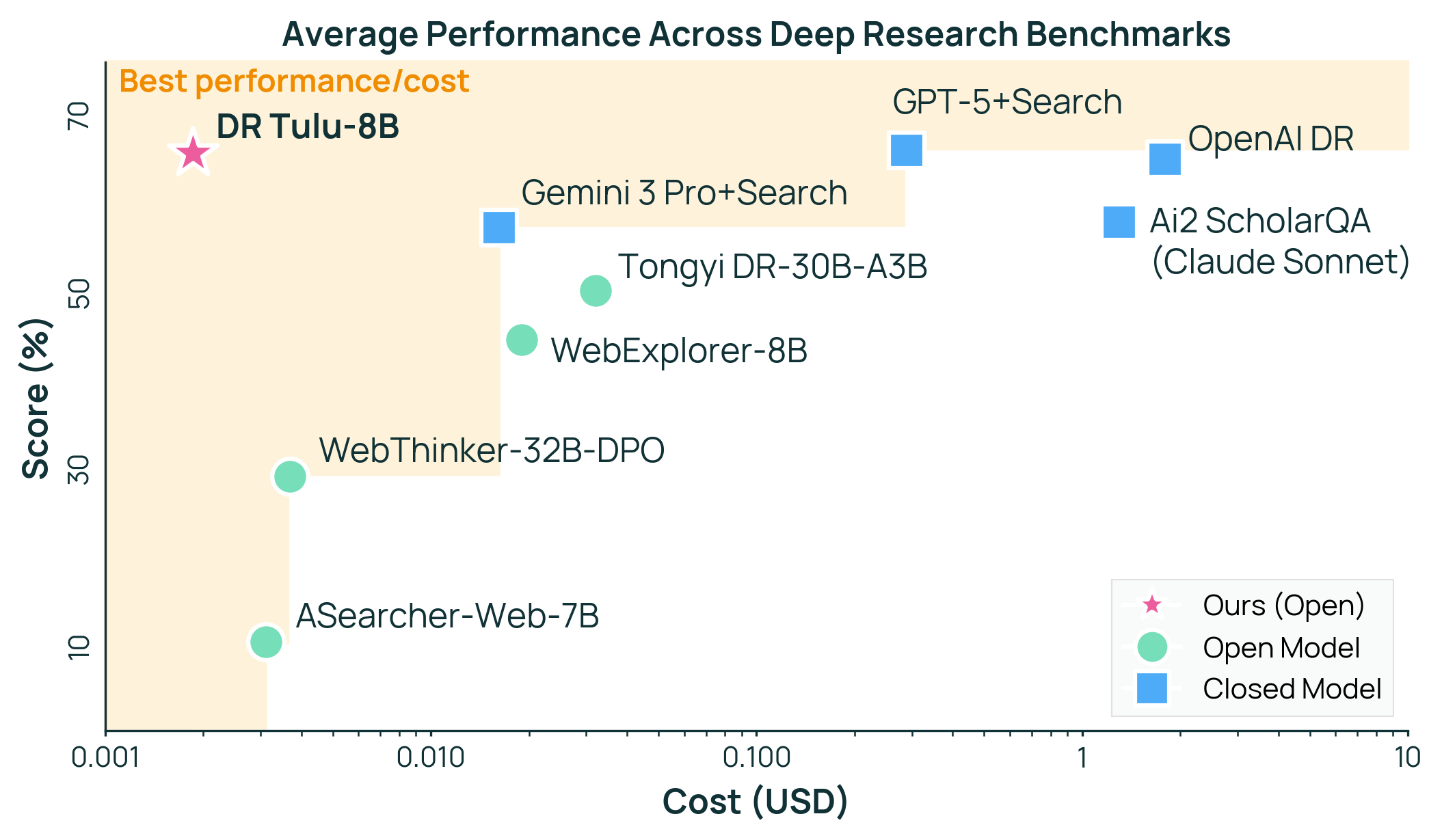}
    \hfill
    \begin{minipage}[t]{\linewidth}
        \vspace{-15pt}
        \caption{\textbf{Performance vs.~cost of deep research models.} We report average performance over 4 long-form DR benchmarks (ScholarQA-CSv2, HealthBench, ResearchQA, and DeepResearchBench) against inference cost (USD per query on ScholarQA-CSv2). 
        \modelname-8B lies on the Pareto frontier, outperforming larger open models and matching proprietary models (Table~\ref{tab:main_results}).}
        \label{fig:first-results}
    \end{minipage}
    \vspace{-5pt}
\end{figure}

\section{Introduction}\label{sec:intro}
Deep research (DR) agents aim to produce in-depth, well-attributed answers to complex research tasks by planning, searching, and synthesizing information from diverse sources~\citep{openai2025deepresearch}.
Existing open DR agents are either training-free, using manually designed prompts with off-the-shelf models~\citep{li2025webweaver,li2025webthinker}, or trained via reinforcement learning with verifiable rewards (RLVR) on search-intensive yet constrained short-form question answering
\citep{jin2025search, nguyen2025sfr, liu2025webexplorer}.
%
RL training for open-ended DR tasks critically depends on reliable reward signals. However, defining such rewards is challenging. The desiderata for good responses are often under-specified~\citep{xu-etal-2023-critical,krishna-etal-2021-hurdles} and therefore hard to fully capture with static, pre-defined evaluation criteria. Moreover, accurate assessment often requires access to extensive and up-to-date external information beyond a model’s parametric knowledge.

In this paper, we introduce \textbf{Deep Research Tulu (\modelname-8B)}, the first open model trained end-to-end for \emph{open-ended, long-form} DR tasks. 
\modelname-8B is first finetuned on high-quality, naturally occurring user data, and then trained via a new method, \textbf{Reinforcement Learning with Evolving Rubrics (RLER)}, in which we construct and maintain rubrics that {\it co-evolve} with the policy model during training. 
At each training step, we sample several responses and search traces from the model, and generate new rubrics that capture and contrast the strengths and weaknesses of these responses. This lets us continuously update the rubrics with newly discovered information, keeping feedback on-policy and discriminative across model responses.

\modelname-8B outperforms the strongest open 8--32B models, including previous state-of-the-art Tongyi DR 30B~\citep{team2025tongyi}, by 4.8--41.8 percentage points on four long-form DR benchmarks---AstaBench-ScholarQA-CS2 (SQAv2)~\citep{asai2024openscholar,bragg2025astabench}, DeepResearchBench~\citep{du2025deepresearch}, ResearchQA~\citep{yifei2025researchqa}, and HealthBench~\citep{arora2025healthbench}. 
In addition, it matches or exceeds proprietary systems such as OpenAI DR, Perplexity DR, and Gemini3 Pro + Search. 
As Figure~\ref{fig:first-results} shows, 
\modelname-8B is substantially more cost-efficient than all other models: on SQAv2, OpenAI DR costs about USD 1.8 per query, whereas \modelname-8B is almost three orders of magnitude cheaper at USD 0.0019. 
We further construct {\bf GeneticDiseasesQA}, a challenging clinical deep research dataset that requires models to search for and synthesize supporting evidence to assess the therapeutic eligibility of disease-causing genetic variants. On GeneticDiseasesQA, \modelname-8B similarly exceeds or competes with proprietary DR agents; no other open agents can tackle this task due to their inability to produce reliable, verifiable citations. 

Our analysis shows that RLER improves the model's ability to produce more comprehensive and in-depth long-form responses with accurate citations, yielding gains of 6.4--16.0 points on top of the finetuned model across the four benchmarks.
Moreover, \modelname-8B learns to select appropriate search tools for each task, instead of relying on a single hard-coded search tool like in prior work~\citep{gao2025beyond,bragg2025astabench}.
On SQAv2, \modelname-8B uses paper search 90\% of the time, whereas on DeepResearchBench, whose questions span more diverse, general-domain topics, it relies on web search and browsing about 55\% of the time.

We release all data, code, and models, along with an extensible deep research library (\texttt{dr-agent-lib}) and an evaluation suite supporting plug-and-play multi-tool search. This release provides an end-to-end training stack for deep research agents, including data and infrastructure for asynchronous tool calls and scalable RL over long-horizon tool-use trajectories---addressing a long-standing barrier to deep research training, where data, code, and infrastructure are rarely available.

\section{Preliminaries}

This section covers the Deep Research formulation and rubrics-as-rewards preliminaries.

\textbf{Problem formulation.}\label{sec:problem}
We consider a \emph{deep research model} to be a language model (LM) equipped with search-related tools.
Each tool takes a query and arguments, returning textual resources that can be cited in the model’s answer.
Concretely, we define the model's action space as \{\Think, \CallTool, \Answer, \Cite\}.
At each step, the model samples an action and its associated content or arguments.
If the sampled action belongs to $\{\text{\Think},\text{\Answer},\text{\Cite}\}$, the output is appended to the context.  
If the sampled action is $\text{\CallTool}$, the model executes the tool call, receives the tool observation, and appends it to the context. 
The process continues until the model chooses the action $\text{\Answer}$, producing the final answer.
We refer to Appendix~\ref{app:problem_formulation} for formal definitions and the specification of tool protocol tokens.

\textbf{Rubrics as rewards.}
\label{sec:rler-preliminaries}
Rubrics define explicit evaluation criteria for assessing the quality of (typically long-form) model responses~\citep{viswanathan2025checklists,gunjal2025rubrics}. We consider sample-wise rubrics, in which the evaluation criteria are specified on a per-example basis in natural language:
Given a question $x$ with associated rubrics $\mathcal{R}_x=\{(r_{x,k}, w_{x,k})\}_{k=1}^K$, where $r_{x,k}$ denotes a rubric item and $w_{x,k}\in\mathbb{R}$ its weight, we evaluate a final response $y$ using the rubric-based score
\begin{equation}\label{eq:rubric_score}
S(x,y)=\frac{\sum_{k=1}^K w_{x,k}\,\textsc{Judge}(r_{x,k}, y)}{\sum_{k:\,w_{x,k}>0} w_{x,k}} .
\end{equation}
Each rubric is evaluated by a judge LM that outputs $\{0, 0.5, 1\}$ based on how well $y$ satisfies $r_{x,k}$. 
During training, we optimize the expected rubric score over the training questions using RL.
Using rubrics as rewards offers several advantages: their concrete, well-defined items reduce susceptibility to judge model bias and promote objective evaluation, yielding consistent and comparable scores across different LLM-as-a-judge runs.

\section{RLER: Reinforcement Learning with Evolving Rubrics}\label{sec:reward}
\label{sec:rler_main_section}

\begin{figure*}[t!]
    \centering
    \includegraphics[width=\linewidth]{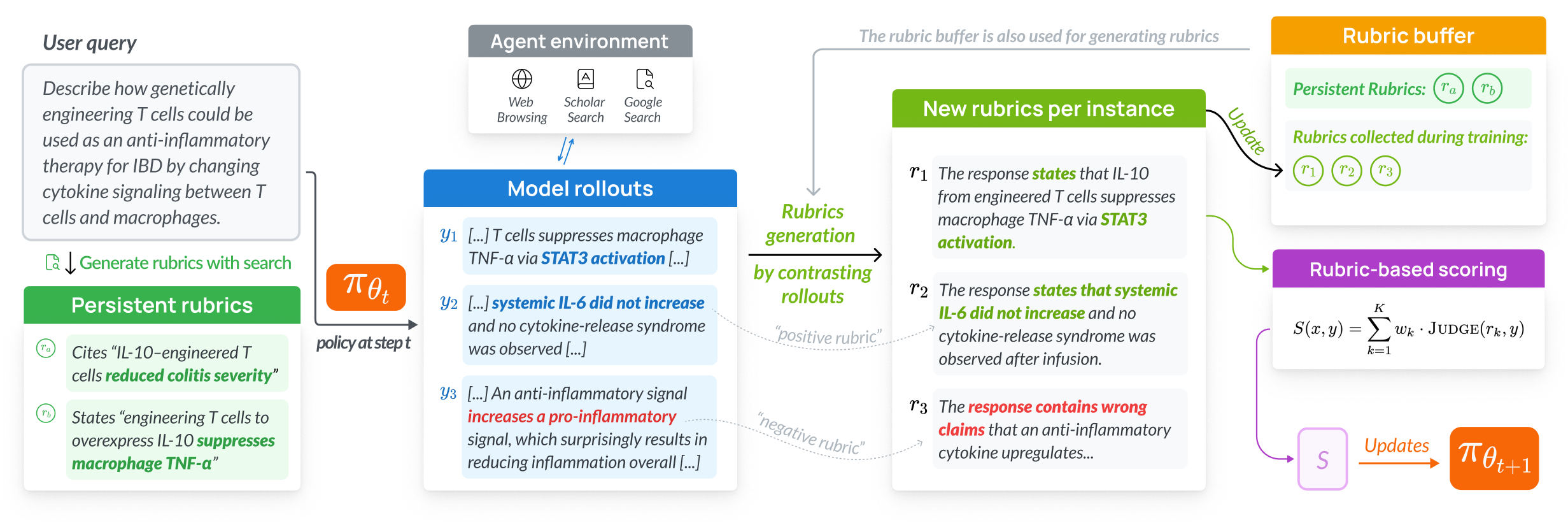}
    \caption{\textbf{Training with RLER.} 
Given an instance, the policy LM $\pi_{\theta_t}$ samples rollouts by interacting with the environment. A rubric LM proposes new rubrics from the rollouts and the current rubric buffer. We score rollouts with these rubrics to update $\pi_{\theta_t}$, then add and prune rubrics to keep a fixed-size buffer with the highest rollout-score variance.
    }
    \label{fig:main_workflow} 
\end{figure*}

Despite the recent adoption of rubrics for evaluation, these approaches typically rely on human experts to write and iteratively refine the rubrics~\citep{arora2025healthbench,du2025deepresearch,sharma2025researchrubrics}, or assume the availability of reference answers~\citep{gunjal2025rubrics}.
Automating and scaling rubric generation for training remains challenging:
Long-form questions are often under-specified and admit many plausible notions of quality, making a small set of fixed criteria inadequate for training. Deep research tasks are also knowledge-intensive and require grounding claims in a broad, evolving body of external knowledge beyond an LM’s parametric knowledge. As a result, closed-book static LM-generated rubrics risk missing evidence, subtle errors, and are vulnerable to reward hacking.


To address these challenges, we 
introduce \textit{Reinforcement Learning with Evolving Rubrics (RLER)} for long-form deep research, using rubrics that are \textit{instance-specific}, \textit{grounded in external knowledge}, and \textit{co-evolve with the policy model}.

\subsection{Search-Augmented Evolving Rubrics}
\label{sec:rlar-rubrics}

The key intuition behind RLER is to improve rubric quality by providing the rubric generator with \textit{privileged information} that is unavailable to the policy during generation, thereby creating a generation–verification gap. 
Concretely, our design leverages two forms of privileged information:
(1) external knowledge retrieved from multiple search rollouts, which supports fact verification; and
(2) multiple independently sampled model responses, which provide contrastive signals for assessing relative quality.

We next detail our RLER framework, covering rubric initialization, online rubric evolution with buffer management, and auxiliary format and citation rewards (Figure~\ref{fig:main_workflow}; Algorithm~\ref{alg:rubric-training}).

\textbf{Initial search-based rubrics.}
For each training prompt $x$, we build a customized rubric buffer to store evolving rubrics that are dynamically updated during training.
Before training, we initialize the rubric buffer with search-based rubrics. 
Specifically, for each $x$, we first perform $ \textsc{Search}(x)$ to fetch relevant documents via web search API using the original question. 
We then concatenate the retrieved documents with the question $x$ and feed them into an LM, $\mathcal{G}_{\text{rubric}}$, to produce a set of initial rubrics that will be persistently used throughout RL training:
$\mathcal{R}^{\text{persist}}_x = \{R_1, R_2, \dots, R_{K_s}\}$,
where $K_s$ denotes the number of persistent rubrics.

\textbf{Evolving rubrics during training.}
During training, we add a new set of evolving 
rubrics to the active rubric buffer, $\mathcal{R}^{\text{active}}_x$, which are used for scoring.
In each step, for every prompt $x$ and its corresponding set of responses $\{y_i\}_{i=1}^G$, where $G$ denotes the number of rollouts, 
we concatenate the prompt $x$, all sampled responses $\{y_i\}_{i=1}^G$ (including the search context and final answers), and the existing rubric pool 
$\mathcal{R}_x = \mathcal{R}^{\text{persist}}_x \cup \mathcal{R}^{\text{active}}_x$
 as input to $\mathcal{G}_{\text{rubric}}$, obtaining a set of \textit{evolving rubrics} 
$\mathcal{R}^{\text{new}}_x = \mathcal{G}_{\text{rubric}}\!\left(x, \{y_i\}_{i=1}^G, \mathcal{R}_x\right)$.
Specifically, we instruct the LM to generate two types of evolving rubrics:
(1) \textit{positive rubrics}, which capture strengths or new, relevant knowledge explored by the current policy but not yet reflected in $\mathcal{R}_x$, and
(2) \textit{negative rubrics}, which summarize common undesirable behaviors, such as reward hacking observed across responses. For example, verbatim copying of retrieved content to maximize citation precision can be identified and suppressed by negative rubrics.
Appendix~\ref{app-evolving-rubric-prompt} presents rubric generation prompts.

\textbf{Rubric buffer management.}
\label{sec:rubric_buffer_management}
Without appropriate management, the number of rubrics would grow linearly during training as new rubrics are continuously generated. 
To maintain a compact yet informative set, we developed a rubric buffer management strategy that filters, merges, and ranks rubrics based on their \textit{discriminative} power.
After every GRPO rollout, we score all responses $\{y_i\}_{i=1}^G$ using the current active rubrics and obtain rubric-level scores.
Rubrics with zero variance in their corresponding rewards are removed as they offer no discriminative value. 
We then compute the standard deviation 
for each remaining rubric and rank them by the standard deviation in descending order.  
To limit evaluation cost, we retain only the top $K_{\max}$ rubrics with the highest standard deviation values.

In addition to evolving rubrics, we introduce three auxiliary rewards---format, search, and citation rewards---to encourage correct formatting, effective use of search and high-quality citations that support relevant claims. 
We detail these auxiliary rewards in Appendix~\ref{sec:citation_format_rewards}.

\section{\modelname with Open Infrastructure and Training Recipe}

Building on RLER, we train \textbf{\modelname-8B} starting from Qwen3-8B~\cite{yang2025qwen3}. This section describes our agent infrastructure and the SFT-then-RL training recipe.

\subsection{\modelname Agent Infrastructure: \texttt{dr-agent-lib}} 
\label{subsec:infra}

DR agents require an extensible, scalable, and user-friendly tool infrastructure for diverse search and browsing APIs. We develop \texttt{dr-agent-lib}, an agent library with three core features: (i) a unified MCP-based tool backend integrating local and API-based web search and browsing tools (Table~\ref{tab:agent-tools}); (ii) a high-concurrency backend with global caching and asynchronous process locking for efficient, rate-limit-aware tool execution; and (iii) a lightweight, composable prompt layer enabling fine-grained control over search workflows and configurations. For training, we implement an auto-search workflow (Appendix~\ref{sec:dr-tulu-prompt}) using \texttt{google\_search} (query $\rightarrow$ top web snippets), \texttt{web\_browse} (URL $\rightarrow$ crawled page text), and \texttt{paper\_search} (query $\rightarrow$ paragraphs from papers).

\subsection{Supervised Fine-Tuning for Cold Start}
\label{subsec:sft}

We apply supervised fine-tuning (SFT) as a cold start to distill common search patterns from a teacher model into the initial model, improving early rollout quality and accelerating subsequent RL training (ablations in \S\ref{sec:analysis}).

\textbf{Prompts.}~~We curate or synthesize both long-form and short-form prompts. Long-form queries are real user queries collected from {SearchArena}~\citep{miroyan2025search} and {OpenScholar}~\citep{asai2024openscholar}, covering general-domain and scientific-domain questions, respectively. To address large quality variation in real-world queries~\citep{cao2025out}, we apply a prompt-filtering stage in which an LM scores each prompt on a 1-5 scale. Short-form prompts are drawn from existing datasets, including HotpotQA~\citep{yang-etal-2018-hotpotqa}, TaskCraft~\citep{shi2025taskcraft}, WebWalker-Silver~\citep{wu2025webwalker}, and MegaScience~\citep{fan2025megascience}, supplemented with challenging synthetic prompts inspired by PopQA~\citep{mallen-etal-2023-trust}. Details are in Appendix~\S\ref{app_sec:details_sft}. 

\textbf{Teacher trajectories.}~~Given each prompt, we instruct GPT-5 to generate a trajectory, including simulated reasoning, tool use, and the final answer, using a system prompt that specifies the aforementioned auto-search workflow. We apply two rejection-sampling filters: (i) retaining only trajectories that follow the expected tool-calling and answer formats, and (ii) for short-form prompts, discarding trajectories whose final answers do not match the gold answers, following prior work~\citep{jin2025search,li2025webthinker}. This process yields 16K SFT trajectories  (Appendix Table~\ref{tab:sft_data_stat}).

\subsection{Online RL with Asynchronous Tool Calls}
\label{subsec:rl}

We further train \modelname-8B using RLER with a customized variant of GRPO~\citep{shao2024deepseekmath}. Training proceeds by iteratively generating agentic rollouts with real tool calls and scoring the model’s final answers against evolving rubrics.

RL training focuses exclusively on long-form questions. Using the same LM-based filtering procedure as in long-form SFT, we collect approximately 5K prompts from SearchArena~\citep{miroyan2025search} and OpenScholar~\citep{asai2024openscholar}, and an additional 4K prompts from RaR~\citep{gunjal2025rubrics} to increase data diversity.\footnote{For RaR prompts, we initialize training with the dataset-provided rubrics rather than generating search-based rubrics, while still maintaining evolving rubrics during training.} 
Despite sourcing from multiple datasets, the collected prompts remain partially out-of-distribution relative to our evaluation datasets. 

We train using GRPO~\citep{shao2024deepseekmath} based on the Open-Instruct implementation~\citep{lambert2025tulu}, incorporating token-level loss~\citep{yu2025dapoopensourcellmreinforcement}, 1-step asynchronous training~\citep{noukhovitch_asynchronous_2024}, tool output masking~\citep{jin2025search}, and sample packing for improved efficiency. We further adopt asynchronous tool calling~\citep{jiang2025verltool}, where tool requests are dispatched immediately upon triggering during rollout generation, rather than waiting for batch completion. Additional training details and hyperparameters are provided in Appendix~\ref{app:rl_hyperparameters}.

\section{Experimental Results}

\begin{table*}[t!]
    \centering
    \small
    \begin{tabular}{lcccc|c}
    \toprule
          &  SQAv2 & HealthBench & ResearchQA & DRB & Average \\ 
    \midrule
        \textcolor{gray}{\it Closed Deep Research} & & & & & \\ 
        \rowcolor{gray!20} \claude\ Claude-Sonnet Search &  -- & -- & 64.3$^*$ & 34.5$^*$ & -- \\
        \rowcolor{gray!20} \perplexity\ Perplexity-Sonar (High) &  -- & -- & 69.1$^*$ & 40.7$^*$ & -- \\
        \rowcolor{gray!20} \perplexity\ Perplexity Deep Research &  67.3 & -- & 75.3$^*$ & 42.3$^*$  & --  \\
        \rowcolor{gray!20} \gemini\ Gemini Deep Research &  -- & -- & 68.5$^*$ & 48.8$^*$ & -- \\
        \rowcolor{gray!20} \gemini\ Gemini 3 Pro + Search & 69.8 &38.0  &  74.3 &46.3 & 57.0  \\
        \rowcolor{gray!20} \oai\ GPT-5 + Our Search  & 61.1 & 31.1 & 62.8 & 50.3 & 51.3 \\
        \rowcolor{gray!20} \oai\ GPT-5 + Search & 74.8 & 59.5$^\dagger$ & 78.2$^\dagger$ & 50.7  & 65.8 \\
        \rowcolor{gray!20} \oai\ OpenAI Deep Research  & 79.6 & 53.8$^\dagger$ & 79.2$^\dagger$ & 46.9$^*$ & 64.9  \\
        \midrule
        \textcolor{gray}{\it Naive RAG} & & & & & \\ 
        Qwen3-8B & 40.4 & 16.5 & 56.1 & 33.3 & 36.5 \\
        QwQ-32B & 41.9 & 24.5 & 60.9 & 40.3 & 41.9\\
        \midrule
            \textcolor{gray}{\it Open Deep Research Models} & & & & & \\
        Search-R1-7B \opencode\ \opendata & 22.2 & -0.1 & 27.9 & 9.5 & 14.9 \\
        ASearcher-Web-7B \opencode\ \opendata & 26.9 & -13.0 & 19.4 & 7.8 & 10.3 \\
        WebExplorer-8B & 42.5 & 33.7 & 64.8 & 36.7 & 44.4 \\
        WebThinker-32B-DPO&  32.9 & 11.1 & 48.6 & 23.3 & 28.9 \\
        Tongyi DeepResearch-30B-A3B &  46.5 & 46.2 & 66.7 & 40.6 & 50.0 \\
        \midrule
        \textcolor{gray}{\it Fixed Pipeline Deep Research} & & & & & \\ 
        WebThinker QwQ-32B (report) & 45.2 & 36.5 &  72.8 & 37.9  & 48.1 \\
        WebThinker-32B-DPO (report) & 46.7 & 39.4 & 74.2 & 40.6 & 50.2 \\ 
        \rowcolor{gray!20} \aitwo\ Ai2 ScholarQA - Claude Sonnet & 87.7 & 32.0$^\dagger$ & 75.0$^\dagger$ & 36.1 & 57.7\\ 
        \midrule
        \textcolor{gray}{\it Open Deep Research (Ours)} & & & & & \\ 
        Qwen3-8B + Our Search \opencode\ \opendata & 57.2 & 5.9 & 46.3 & 18.2 & 31.9 \\
        \rowcolor{pink!20} \modelname-8B (SFT) \opencode\ \opendata & 72.3 & 38.1 & 68.5 & 39.0 & 53.9  \\ 
        \rowcolor{pink!20} \modelname-8B (RL) \opencode\ \opendata  &  \bf 88.3 &  \bf 52.8 & \bf 75.7 & \bf 45.4  & \bf 65.6 \\ 
    \bottomrule
    \end{tabular}%
    \caption{{\bf Overall results. \modelname-8B outperforms all open deep research models, and is competitive with proprietary systems.} 
    \textbf{Bold} indicates the best performance among open models. 
* denotes scores reported by the original benchmark authors. Except for GPT5 + our tool, we reuse the existing leaderboard results rather than rerunning the evaluations, which would cost a few hundred USD per task; we leave entries as ``--'' when the original benchmarks do not report the corresponding metric. $\dagger$ denotes that the evaluation was run on a 100-sample subset because the method is expensive. 
For open models, \opencode\ indicates that the training code is open-sourced, and \opendata\ indicates that the training data is open-sourced. 
None of the existing open deep research models output citations, so their citation scores on SQAv2 are 0. 
HealthBench scores can be negative, as HealthBench includes negative rubrics that indicate harmful responses.
    }
    \label{tab:main_results}
\end{table*}

\subsection{Experimental Settings}
\paragraph{Benchmarks.}
We evaluate deep research agents on four long-form, open-ended benchmarks: \textbf{HealthBench}~\citep{arora2025healthbench} for healthcare, \textbf{ResearchQA}~\citep{yifei2025researchqa}, \textbf{AstaBench-ScholarQA-CS2} (SQAv2;~\citealt{asai2024openscholar,bragg2025astabench}) for scientific literature synthesis, and \textbf{DeepResearchBench} (DRB;~\citealt{du2025deepresearch}) for general-domain deep research. All benchmarks require long-form responses and are evaluated using human-written or human-verified rubrics following official protocols. SQAv2 and DRB additionally report fine-grained metrics, including relevance, instruction-following, and citation precision/recall. We also evaluate \modelname-8B on short-form QA (Analysis). Further evaluation details are provided in Appendix~\ref{app_sec:score_calculation}.

\textbf{Baselines.}
We compare against multiple categories of deep research systems (Table~\ref{tab:main_results}). 
(1) \textbf{Open deep research models}: ASearcher-7B~\citep{gao2025beyond}, WebThinker-32B~\citep{li2025webthinker}, Search-R1-7B~\citep{jin2025search}, WebExplorer-8B~\citep{liu2025webexplorer}, and Tongyi Deep Research-30B~\citep{team2025tongyi}. None of these models was evaluated on realistic long-form benchmarks, as their training primarily targets short-form QA.
For long-form tasks, we supply the official evaluation prompts and require full report-style outputs.
(2) \textbf{Fixed-pipeline deep research}: WebThinker-32B (report mode) and Ai2 ScholarQA~\citep{singh-etal-2025-ai2}, which combine LMs with fixed inference-time pipelines; we run their official implementations with default or recommended settings.
(3) \textbf{Closed deep research}: OpenAI Deep Research, Perplexity Sonar (reasoning), Perplexity Deep Research, Claude-Sonnet Search, and Gemini3 Pro + Search.
Additionally, we evaluate Qwen3-8B and QwQ-32B using naive RAG and our inference pipeline built on \texttt{dr-agent-lib}. More baseline details are provided in Appendix~\S\ref{app_sec:details_of_baselines}. 
Existing open deep research models often omit citations, and proprietary systems typically provide only URL-level links. In contrast, \modelname-8B generates snippet-level citations that directly support claims, enabling verification and improving factual reliability~\citep{liu-etal-2023-evaluating}.

\textbf{Training details.}
We initialize from Qwen3-8B~\citep{yang2025qwen3}. SFT is conducted on a single H100 node (8 GPUs) for 5 epochs, totaling 136 GPU hours; SFT hyperparameters are provided in Appendix~\ref{app:sft_hyperparameters}. RL training uses the hyperparameters in Appendix~\ref{app:rl_hyperparameters}. Unless otherwise stated, all training runs use 2 H100 nodes (16 GPUs), our final run using 27,000 GPU hours. We use GPT-4.1-mini (\texttt{gpt-4.1-mini-2025-04-14}) as the LM judge, and GPT-4.1 as the rubric generator.

\textbf{Inference details.}
We use a unified inference pipeline with three tools, \texttt{google\_search}, \texttt{web\_browse}, and \texttt{paper\_search}, for all long- and short-form tasks, without task-specific customization. Following prior work, we use the Serper Search API for \texttt{google\_search}~\citep{li2025webthinker} and Jina browsing for \texttt{web\_browse}~\citep{gao2025beyond,liu2025webexplorer}, rather than the Crawl4AI browser used during training; we verify in Appendix~\ref{app:browser_tool_ablation} that this train/inference browser mismatch has minimal impact on downstream performance. For \texttt{paper\_search}, we use the Semantic Scholar full-text API, which returns relevant paragraphs. We cap tool usage at 10 calls per rollout and retrieve the top 10 snippets for both \texttt{google\_search} and \texttt{paper\_search}. For \texttt{web\_browse}, we summarize long outputs using Qwen3-8B, while truncating webpages during training to improve efficiency.

\subsection{Main Results}
\label{sec:main_result}
We report overall results across four long-form datasets in Table~\ref{tab:main_results}. Appendix Table~\ref{tab:performance_breakdown} provides a fine-grained breakdown of SQAv2 and DRB.

\begin{figure*}[t!]
    \centering
    \includegraphics[width=\linewidth]{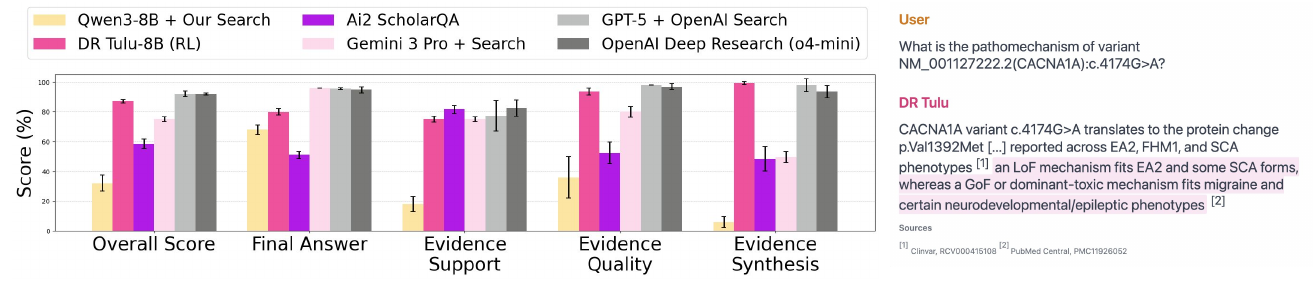}
     \vspace{-20pt}
    \caption{{\bf Comparison of DR agents GeneticDiseasesQA.} 
     \textit{Final Answer}: proportion of expert-annotated facts recovered in responses. \textit{Evidence Support}: the proportion of cited claims that are fully supported by the original text of the cited source. \textit{Evidence Quality}: whether the type of supporting evidence requested is present. \textit{Evidence Synthesis}: whether there was a statement addressing the relationship between multiple sources. Results for \modelname-8B (RL) and Qwen3-8B + Our Search are reported as the average of 10 trials. The remaining agents are reported as the average of 3 trials, given the high costs and inference times for proprietary deep research systems.}
    \label{fig:genetic_disease_qa}
    \vspace{-10pt}
\end{figure*}

\textbf{\modelname-8B outperforms all open deep research models on long-form tasks.}
Across four open-ended long-form benchmarks, \modelname-8B (RL) achieves the strongest performance among all open deep research models, with an average score of 65.6, exceeding the best prior open baseline (Tongyi Deep Research-30B) by 15.6 points. Models trained primarily for constrained short-form tasks (e.g., Search-R1 and ASearcher) perform poorly on realistic report-length generation, yielding very low scores. 
Notably, existing open baselines lack citations, resulting in especially low SQAv2 scores when citation quality is central.

\textbf{\modelname-8B outperforms open fixed-pipeline deep research systems.}
WebThinker-32B, with its heavily engineered report-mode inference, boosts long-form performance (+21.3 points vs. default) but still trails \modelname-8B on every benchmark, despite using a much larger 32B backbone. 
Ai2 ScholarQA, which is designed for scientific literature synthesis and uses a closed backbone (Claude Sonnet), performs competitively on SQAv2 but lags behind \modelname-8B on HealthBench and DeepResearchBench, resulting in a lower overall average. 
Overall, despite using a smaller open model and a single inference pipeline where it autonomously decides its search strategy and response structure from the prompt, \modelname-8B achieves the best average performance among open systems. 
We further observe that fixed-pipeline systems generalize poorly to short-form QA, often applying report-style reasoning to simple factoid queries, whereas \modelname-8B handles both long- and short-form tasks (\S\ref{sec:short_form}).

\textbf{\modelname-8B matches or outperforms proprietary deep research systems.}
\modelname-8B achieves the strongest performance among all systems on SQAv2 and matches or exceeds proprietary deep research systems across the remaining long-form benchmarks. It outperforms Claude Sonnet Search, Perplexity Sonar (high-reasoning), and Perplexity Deep Research, and is competitive with OpenAI Deep Research overall. We further observe that GPT-5 + Search and Gemini3 Pro + Search outperform their corresponding deep research variants on some datasets, suggesting that underlying base model capability plays a critical role in addition to the research pipeline itself. Notably, despite being built on an 8B open model, \modelname-8B remains on par with, or even outperforms, these proprietary, larger-scale systems.

\textbf{\modelname-8B is significantly cheaper than proprietary and open deep research systems.}
\modelname-8B exhibits a substantial cost advantage (Appendix Table~\ref{tab:usage-stats}; see Appendix~\ref{app_sec:cost_estimation}). Proprietary systems are orders of magnitude more expensive: OpenAI Deep Research costs \$1.80/query on SQAv2, and Ai2 ScholarQA (Claude Sonnet) costs \$1.30/query. In contrast, \modelname-8B costs \$0.00008/query when accounting only for tool APIs, and \$0.0018/query when including LM inference via OpenRouter (Qwen3-8B pricing). \modelname-8B also remains cheaper than other open deep research models, including Tongyi Deep Research (\$0.03/query) and WebThinker (\$0.003/query; \$0.015 in report mode), despite achieving stronger performance. This efficiency stems from adaptive tool usage: on SQAv2, \modelname-8B primarily relies on free paper search, and even on DRB, where web search and browsing are used more frequently, it remains over 10$\times$ cheaper than Tongyi DR. 

\subsection{Application: Researching Pathogenic Gene Variants}
\label{sec:application_eval}

To evaluate \modelname on a realistic, expert-driven deep research task, we study pathogenic variant interpretation in clinical genetics. In collaboration with medical experts, we curate questions that reflect real-world deep research challenges in diagnosing rare genetic diseases.

We introduce \textbf{GeneticDiseasesQA}, a dataset of 47 expert-curated questions covering 24 pathogenic gene variants, which requires aggregating heterogeneous evidence from biological databases, research literature, and case reports.  Questions focus on molecular consequences, disease mechanisms, and therapeutic evidence. For each question, models generate a long-form, citation-backed report. Evaluation criteria—\textit{Final Answer}, \textit{Evidence Support}, \textit{Evidence Quality}, and \textit{Evidence Synthesis}—are illustrated in Figure~\ref{fig:genetic_disease_qa}, with additional details in Appendix~\ref{app:genetic_disease_eval}.

\textbf{Results.}
Figure~\ref{fig:genetic_disease_qa} compares \modelname-8B (RL) against Qwen3-8B + search, Ai2 ScholarQA, Gemini 3 Pro + Search, GPT-5 + Search, and OpenAI Deep Research (o4-mini). \textit{Evidence Support} is computed from cited snippets; for systems that return only URLs, we retrieve webpage content via Jina browsing. We exclude baselines without traceable citations. \modelname-8B substantially improves over Qwen3-8B across all metrics and outperforms Ai2 ScholarQA on \textit{Final Answer} correctness. While GPT-5 and Gemini-based systems achieve higher \textit{Final Answer} scores, \modelname-8B remains competitive on \textit{Evidence Support}, \textit{Evidence Quality}, and \textit{Evidence Synthesis}, highlighting its strength in reliable multi-source reasoning. Overall, these results show that \modelname-8B generalizes effectively to unseen, real-world deep research tasks in expert domains.

\section{Analysis}\label{sec:analysis}
We conduct a set of analyses on \modelname. Unless otherwise specified, this section uses the \modelname SFT checkpoint and an early \modelname RL checkpoint at 1k training steps. 

\subsection{Evaluation on Short-form QA Tasks}
\label{sec:short_form}

Although our RLER training targets long-form, open-ended deep research, our SFT mixture intentionally includes short-form, verifiable QA tasks that require search, enabling the model to handle both concise and multi-paragraph responses. We therefore evaluate how well our SFT and RL models generalize to short-form queries.

We evaluate short-form QA on \textbf{SimpleQA}~\citep{wei2024measuring}, \textbf{WebWalkerQA}~\citep{wu2025webwalker}, and \textbf{2Wiki}~\citep{ho-etal-2020-constructing}. Following prior work~\citep{li2025webthinker,wei2024measuring}, we use an LLM judge to assess answer correctness and report Pass@1 accuracy with GPT-4.1 as the LLM judge. For efficiency, we evaluate on 1{,}000 randomly sampled questions each from SimpleQA and 2Wiki.
%
Table~\ref{tab:short_form_results} shows that \modelname performs competitively on short-form QA benchmarks. The SFT stage yields substantial gains over the Qwen3-8B + Our Search baseline, demonstrating the effectiveness of our SFT data for short-form QA. Notably, although RL training uses only long-form prompts and explicitly optimizes long-form generation, \modelname (RL) achieves further improvements on short-form QA, increasing the overall average by 3.3 points, indicating strong cross-task generalization.

\subsection{Analysis on Training and Inference}
\label{sec:analysis_on_training}

\begin{figure*}[t!]
    \centering
    \includegraphics[width=\linewidth]{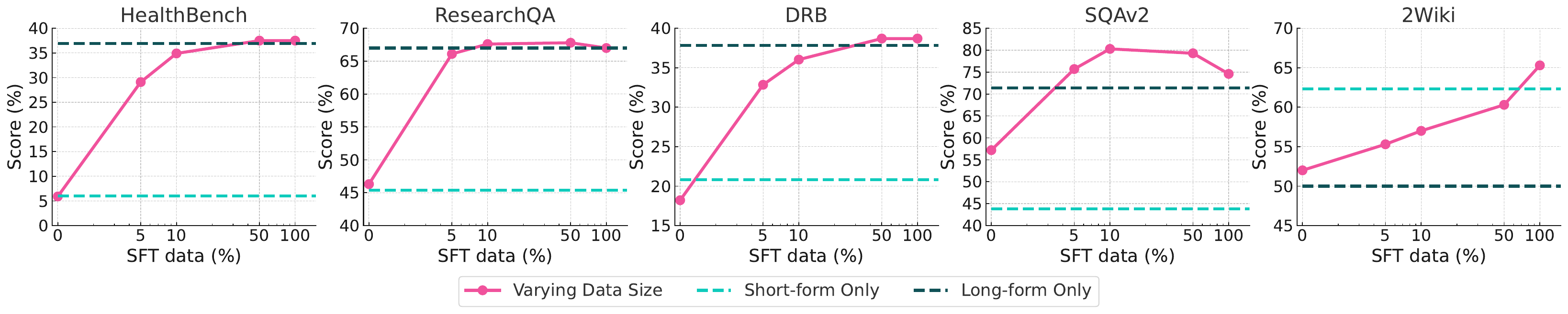}
    \caption{\textbf{Ablation of SFT training data.} We ablate SFT training data in terms of the mixture of data and the scale of training data. We train models with varying sizes of SFT data (5\%, 10\%, 100\%; 0\% indicates the Qwen3-8B + \texttt{dr-agent-lib} results) as well as two SFT subsets, long-form data only (LF only) and short-form data only (SF only).}  
    \label{fig:sft_ablation}
\end{figure*}

\paragraph{SFT benefits from mixed supervision but shows diminishing returns on long-form tasks.}
Figure~\ref{fig:sft_ablation} shows that combining long-form and short-form data during SFT is important: removing long-form data substantially degrades performance on all long-form benchmarks, while removing short-form data leaves long-form performance largely unchanged but noticeably hurts short-form tasks such as 2Wiki. These results indicate that short-form supervision alone does not reliably transfer to open-ended deep research, whereas retaining a modest short-form component helps preserve general-purpose behavior without sacrificing long-form performance. Scaling SFT data yields clear early gains across tasks, with long-form benchmarks showing substantial improvements with as little as 5\% of the data and largely saturating beyond 50\%, while short-form tasks (especially 2Wiki) continue to benefit from additional data up to the full dataset. Although full-data SFT slightly reduces SQAv2 citation scores, overall short-form accuracy remains strong, and subsequent RL training recovers citation performance, motivating our use of the full SFT dataset followed by RL.

\begin{table}[t]
\small
\resizebox{\linewidth}{!}{
    \centering
    \begin{tabular}{lcccc}
    \toprule
     & SimpleQA & 2Wiki & WebWalker &  Avg. \\\midrule
        \textcolor{gray}{\it Naive RAG} & \\ 
        Qwen3-8B & 52.6 & 18.9 & 8.8 &26.8 \\
        QwQ-32B & 57.2 & 34.2 & 10.1 & 33.8\\ \midrule
        \textcolor{gray}{\it Open Deep Research (Ours)} & \\ 
        Qwen3-8B + Our Search & 70.5 & 44.0 & 27.9 & 47.5   \\
        \rowcolor{pink!20} \modelname-8B (SFT) & 75.5 & 66.5 & 31.9 & 58.0 \\
        \rowcolor{pink!20} \modelname-8B (RL) & 75.9 & 68.9 & 39.0 & 61.3  \\
    \bottomrule
    \end{tabular}%
    }
    \caption{{\bf Short-form results.} We report short-form performance for our SFT and RL variants to analyze how each training stage affects short-form behavior. All scores are computed from top-1 predictions under a unified evaluation pipeline.
    }
    \label{tab:short_form_results}
\end{table}

\begin{figure}
    \centering
    \includegraphics[width=.8\linewidth]{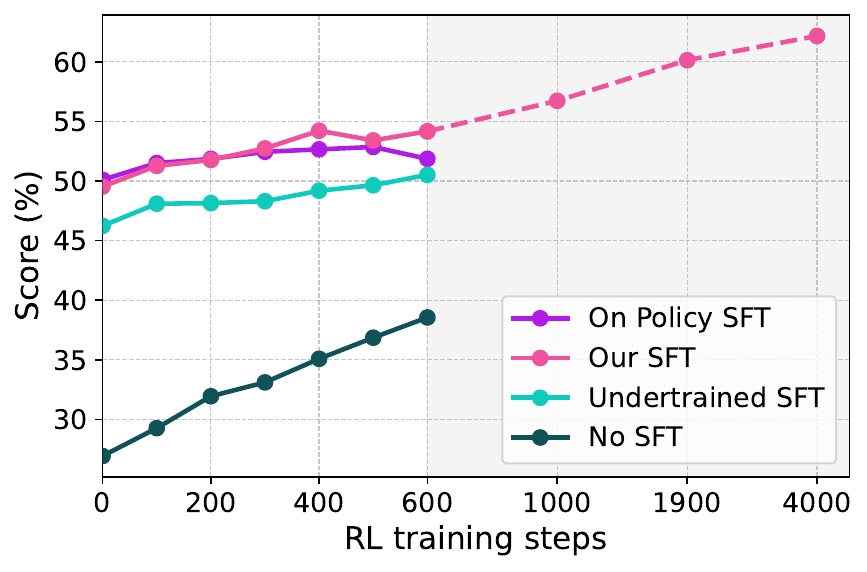}
    \caption{\textbf{Our full SFT mix performs best during RL.} We vary the model used for RL training, keeping data and hyperparameters constant. Note that the x-axis is not uniform in the gray area. Performance is average across Healthbench, SQAv2, DRB.}
    \label{fig:rl_trajectory}
\end{figure}
\begin{figure}[t!]
    \centering
    \includegraphics[width=\linewidth]{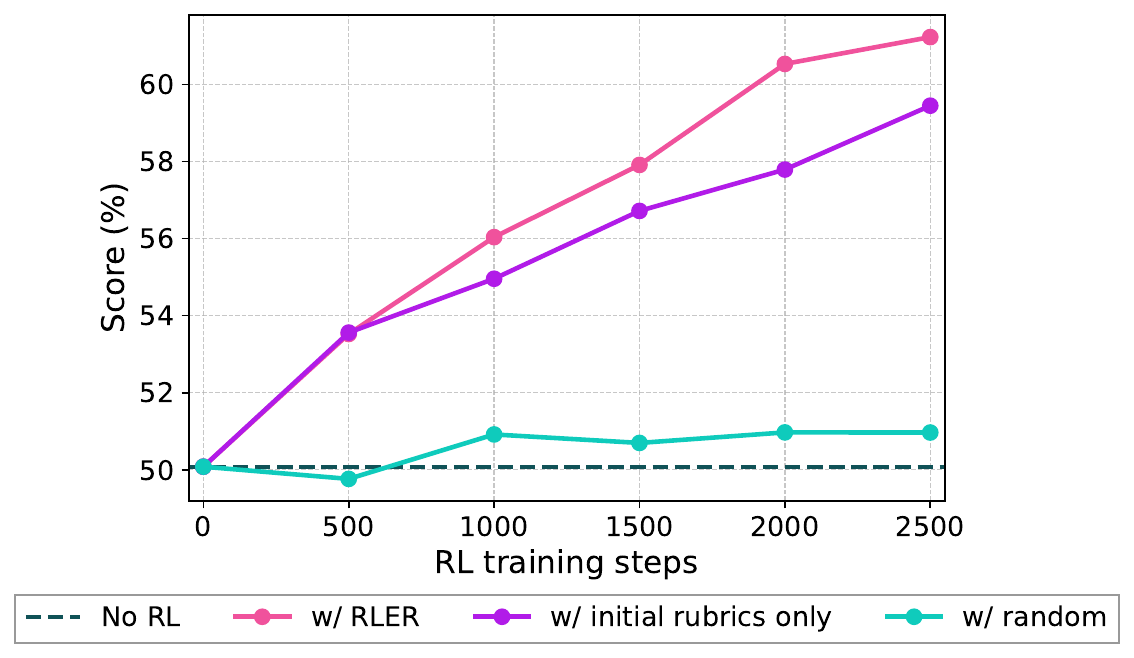}
    \caption{\textbf{RLER consistently improves performance during RL training.} We train models with only our initial search-based rubrics and with RLER. We also compare to using no RL and using purely random rewards. Performance is average across Healthbench, SQAv2, DRB.}
    \label{fig:rler_ablation}
\end{figure}

\begin{table}[t!]
\resizebox{0.48\textwidth}{!}{
    \centering
    \footnotesize
    \begin{tabular}{l l l l l | l}
    \toprule
         & SQAv2 & Health & Research & DRB & Avg. \\ 
    \midrule
        DR Tulu (SFTv0.1) & 73.4 & 37.5 & 68.6 & 39.4 & 54.7 \\ \hdashline
        General rubrics
        & 80.6 
        & 36.0 
        & 65.0 
        & 37.5 
        & 54.8\\

        Closed-book rubrics 
        & 83.2 
        & 34.8 
        & 66.6 
        & 37.6
        & 55.6 \\

        Initial search-based rubrics 
        & 82.8 
        & 37.9 
        & 66.9 
        & 39.3 
        & 56.7 \\

    \bottomrule
    \end{tabular}
    }
    \caption{{\bf Search-based static rubrics work best.} We train using different static rubrics using RL for 500 steps, starting from an intermediate SFT checkpoint (SFT v0.1). 
    ``Initial search-based rubrics'' refers to an ablation of RLER without evolving rubrics generated during training.
    Search-based rubrics consistently outperform both general rubrics and closed-book rubrics that are not grounded in up-to-date information. 
    }
    \label{tab:long-form-abl-rubric}
\end{table}

\textbf{RL benefits from stronger SFT models and longer training.}
We ablate the effect of using different SFT cold start datasets on RL in Figure~\ref{fig:rl_trajectory}, tracing performance up to 4000 training steps. 
Beginning RL directly from Qwen3 (no SFT cold start) dramatically improves scores over Qwen3-8B with no training, but still underperforms using even a small amount of high-quality SFT data (5\% of our full mixture) as cold-start data for the RL training. Using a larger amount of SFT data (i.e., our full SFT mixture) further improves performance. 
Extended RL training was crucial to performance: in some cases, evaluations that initially seemed flat (e.g., DRB) improved with extended RL training.
We found that higher train reward (i.e., reward during RL training) did not necessarily correspond to higher downstream reward; see Appendix~\ref{sec:mismatch_rl_training} for details. 
We also experiment with using an `on-policy SFT' model as a starting point, which we provide further details on in Appendix~\ref{app:onpolicy_sft_rl_results}. 
Finally, we find that our training is robust to tool errors, with the model improving performance even after extended training with a tool that consistently errors. Appendix~\ref{app:full_rl_training_curves} provides details and the full RL training curves.

\textbf{Evolving rubrics improve over initial rubrics alone.}
We ablate evolving rubrics and compare them against RL with static, search-augmented rubrics only in Figure~\ref{fig:rler_ablation}.
Removing evolving rubrics results in up to a 2-point drop in average performance, with the gap widening over training as evolving rubrics capture new knowledge the model explores.
Both approaches outperform random rewards instead of rubric-based rewards, ensuring that our results are not due to spurious behaviors in Qwen-based models~\citep{shao2025spurious}.
Finally, branching a single training run with and without the citation reward enabled yields comparable performance (Appendix~\ref{app:citation_reward_ablation}), indicating that RLER's rubric reward, rather than auxiliary signals, drives the gains.

\textbf{RLER does not rely on a strong proprietary judge.}
We additionally replace GPT-4.1 and GPT-4.1-mini with Qwen3-8B---the same initial model used to train \modelname---as both the rubric generator and the LM judge in Table~\ref{tab:open_judge_ablation_results}). 
After 1000 RL steps, the open-judge variant still gains $+4.4$ average points over the SFT checkpoint, only $1.3$ points behind the GPT-judge configuration ($+5.7$). Combined with the fact that GPT-4.1 and GPT-4.1-mini themselves perform poorly on deep research tasks, this indicates that RLER's gains do not stem from distilling a stronger proprietary judge, and the recipe transfers to settings without access to such models.

\begin{table*}[t!]
    \centering
    \small
    \begin{tabular}{lcccc|c}
    \toprule
          &  SQAv2 & HealthBench & ResearchQA & DRB & Average \\ 
    \midrule
        Qwen3-8B + Our Search & 57.2 & 5.9 & 46.3 & 18.2 & 31.9 \\
        \quad + SFT & 72.3 & 38.1 & 68.5 & 39.0 & 53.9  \\ 
        \quad + RL (1000 steps, GPT-judge) &  \bf 85.8 &  \bf 42.2 & \bf 70.2 & \bf 40.1 & \bf 59.6 \\
        \quad + RL (1000 steps, Qwen3-8B-judge) &  85.3 &  39.7 & 69.2 & 39.1 & 58.3 \\
    \bottomrule
    \end{tabular}%
    \caption{{\bf Comparing using GPT-4.1 and Qwen3-8B as a judge model and rubric generator.} For GPT-judge, we use GPT-4.1-mini as the judge, and GPT-4.1 as the rubric generator. For Qwen3-8B-judge, we use Qwen3-8B as both judge and generator. Using Qwen3-8B only underperforms using GPT models by 1.3 points, while still outperforming the SFT baseline by 4.4 points.}
    \label{tab:open_judge_ablation_results}
\end{table*}

\textbf{Search-based rubrics outperform closed-book rubrics.} We ablate the effect of using different static rubrics (i.e., without adding evolving rubrics) during RL training in Table~\ref{tab:long-form-abl-rubric}. We run RL training (w/o ER) for 500 steps on top of an intermediate SFT checkpoint using three different rubric setups: (1) general rubrics, in which we use a simple prompt and LM judge to score model outputs (see Appendix~\ref{sec:general_rubric_prompt} for prompt); (2) closed-book rubrics, which are generated without access to any search information; (3) search-based rubrics, which are generated with knowledge from an initial search (See Appendix~\ref{sec:rlar-analysis} for details). For these runs, we use only OpenScholar training samples. Search-based rubrics perform best overall, while a single general rubric shared across samples underperforms the SFT baseline.

\paragraph{Evolving rubrics improve over initial rubrics alone.} We additionally ablate using evolving rubrics on top of search-based rubrics in Figure~\ref{fig:rl_trajectory} (right), training for 2500 steps. We find that removing RLER leads to up to a 2-point drop in performance, with the gap widening over training as evolving rubrics capture new knowledge the model explores.

\paragraph{Open rubric judge ablations.}

We additionally experiment with using a fully open model as the judge model for citation and rubric scoring, as well as for generating the evolving rubrics during RL training. We use Qwen3-8B as the judge and generation model and run training for 1000 steps, with the citation reward only turned on for the initial 650 steps as in the main run. We also note that due to context length limitations, we only pass the final answers to the rubric generator, as opposed to the full output trajectory. We present our results in Table~\ref{tab:open_judge_ablation_results}. We compare to our main training run at 1000 steps, in which we used GPT-4.1-mini as the LM judge and GPT-4.1 as the rubric generator.

We find that using an open judge can still improve over SFT alone by over 4 points, although it underperforms using GPT models by roughly 1 point. This suggests that Qwen3-8B is still capable of acting as a judge and rubric generator despite being generally less performant than GPT-4.1-mini and GPT-4.1. Importantly, this also shows that \textbf{RLER does not rely on the presence of a stronger model}, as Qwen3-8B is precisely the starting model used for training \modelname. We leave further exploration of using open-weights or even the model under training itself as the rubric judge and generator to future work.

\textbf{Tool usage adapts to each task’s information needs.}
Figure~\ref{fig:tool_distributions} shows that \texttt{paper\_search} (our scientific-paper search) dominates on SQAv2, consistent with its focus on literature understanding. 
In contrast, \texttt{web\_search} is the primary tool for HealthBench, DeepResearchBench, and SimpleQA, reflecting the broader, open-web information needs of these tasks. 

\section{Related Works}
\textbf{Deep research agents.}
Recent work on DR agents often focuses on short-form QA~\citep{jin2025search,liu2025webexplorer,team2025tongyi,gao2025beyond}.
While some systems target long-form research tasks, they typically rely on static workflows or proprietary components~\citep{li2025webthinker,li2025webweaver,prabhakar2025enterprisedeepresearch,singh-etal-2025-ai2}, offer limited tool support, or do not fully release code and data.
We provide additional discussion of these related works in Appendix~\ref{app:related_works}. 

\textbf{Rubric design for long-form generation.}
Human-written rubrics are commonly used for evaluation but are costly for training~\citep{arora2025healthbench,asai2024openscholar}.
Recent methods generate model-based rubrics for training, including static rubric rewards~\citep{gunjal2025rubrics}, closed-book online rubric generation~\citep{rezaei2025online,jayalath2025compute}, and learned critics for factuality~\citep{wu2025rlac}, but these approaches are ungrounded in external knowledge and remain fixed or weakly adaptive.
Related work also explores retrieval-assisted evaluation criteria~\citep{wadhwa2025evalagent}.
Our approach differs by generating retrieval-grounded rubrics that co-evolve with the policy model.

\section{Conclusion}
We present \modelname-8B and Reinforcement Learning with Evolving Rubrics (RLER), an end-to-end training framework for long-form deep research tasks.
We release the model, data, rubrics, and training infrastructure to support reproducibility and future research on deep research agents.
Looking ahead, \modelname opens several directions for long-form DR training, including adaptive verifier design, scaling privileged information for judges, improving alignment between training rewards and downstream evaluations, and extending DR agents to specialized scientific workflows.
We provide an extended discussion and outline future directions in Appendix~\ref{app:discussion}.


\section*{Impact Statement}
We introduce Reinforcement Learning with Evolving Rubrics (RLER) and \modelname-8B, an open deep research agent trained using this approach. To support reproducibility and further research, we fully open-source the model, training data, evaluation rubrics, and agent infrastructure.
Potential positive impacts include enabling broader access to long-form research capabilities, improving reproducibility, and supporting more rigorous evaluation of deep research agents in science, healthcare, and general domains. However, such systems may amplify harms common to research assistants---e.g., generating plausible but incorrect claims, selective citation, or biased synthesis---and could increase the scale of misinformation or low-quality research outputs if deployed without safeguards. We therefore view \modelname primarily as a research artifact: downstream use should incorporate careful evaluation, transparency about uncertainty and sources, and domain-appropriate human oversight, especially in high-stakes settings.

\section*{Acknowledgments}
This material is based upon work supported by the National Science Foundation under Award No. 2413244. This work was supported by the Singapore National Research Foundation and the National AI Group in the Singapore Ministry of Digital Development and Information under the AI Visiting Professorship Programme (award number AIVP-2024-001), the AI2050 program at Schmidt Sciences, and the DARPA SciFy program (Agreement No. HR00112520300). 
We thank Zhiyuan Zeng, Rui Xin, Stella Li, and Doug Downey for helpful discussions and feedback on the draft.

\bibliography{references}
\bibliographystyle{icml2026}

\newpage
\appendix
\onecolumn

\section*{Appendix}

\section*{Author Contributions}\label{sec:authors}
DR Tulu is a team effort. Below we list each author's primary contributing roles in the project, with bolded authors taking the lead within each role.

\begin{itemize}[leftmargin=1.4em, itemsep=1pt, topsep=2pt, parsep=0pt]
\item \textbf{Project leads:} \textbf{Rulin Shao}, Akari Asai.
\item \textbf{Core contributors:} Rulin Shao, Akari Asai, Shannon Shen, Hamish Ivison, Varsha Kishore, Jingming Zhuo.
\item \textbf{RLER method development:} \textbf{Rulin Shao}, Hamish Ivison, Shannon Shen.
\item \textbf{DR Tulu data:} \textbf{Akari Asai}, Rulin Shao, Jingming Zhuo, Varsha Kishore, Shannon Shen, Luca Soldaini.
\item \textbf{DR Tulu training:} \textbf{Hamish Ivison}, \textbf{Rulin Shao}, Akari Asai, Shannon Shen.
\item \textbf{Infrastructure:} \textbf{Shannon Shen}, \textbf{Hamish Ivison}, Rulin Shao, Luca Soldaini, Tyler Murray, Varsha Kishore.
\item \textbf{Evaluations and baselines:} Varsha Kishore, Shannon Shen, Rulin Shao, Akari Asai, Jingming Zhuo, Hamish Ivison, Xinran Zhao.
\item \textbf{GeneticDiseasesQA benchmark:} \textbf{Molly Park}, Samuel Finlayson.
\item \textbf{Project mentorship:} Hannaneh Hajishirzi, Pang Wei Koh, Yoon Kim, Luke Zettlemoyer, Sherry Tongshuang Wu, Scott Yih, David Sontag, Faeze Brahman, Luca Soldaini, Pradeep Dasigi, Sewon Min.
\end{itemize}
Core contributors made sustained, significant contributions throughout the project. All authors contributed to project discussions, experiment planning, and writing the paper.

\section{Discussion and Future Work}\label{app:discussion}
In this section, we highlight key insights, challenges, and promising directions for future work.

\paragraph{Evolving rubrics adapt the verifier based on the policy model's capabilities.}
At each training step, we update our rubrics by contrasting the model’s current rollouts, which helps the new rubric criteria better distinguish those outputs. We can view this as making the training difficulty adaptive to the model’s evolving behavior. This approach aligns with the idea of training in \textit{adaptive environments}, which has been previously explored by adjusting prompts during training \citep{zeng2025rlve}. In contrast, we adapt the environment by updating the verifier (rubrics). Future work may consider jointly adapting both prompts and rubric criteria to further improve training efficiency.

\begin{wrapfigure}{r}{0.5\textwidth}
  \centering
  \includegraphics[width=0.5\textwidth]{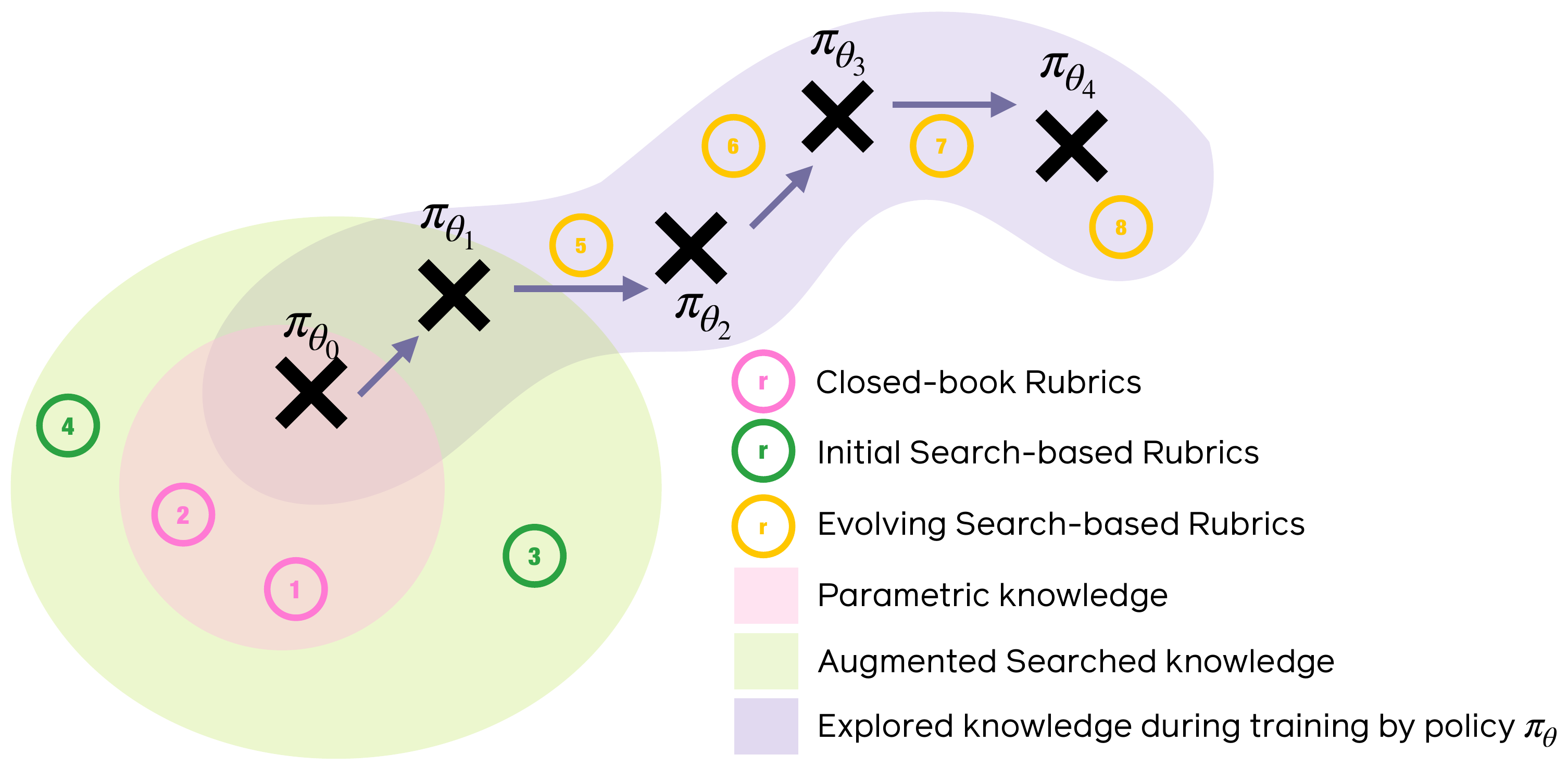}
  \caption{\textbf{Knowledge coverage relationship visualization.} An abstract visualization of the knowledge coverage relationship between closed-book rubrics, initial search-based rubrics, and evolving search-based rubrics.
  }
  \label{fig:new-abstract-knowledge-coverage}
\end{wrapfigure}
\paragraph{A new dimension of scaling verifier compute: providing more privileged information to the judge.}
Another perspective on RLER is that it creates a new way to scale the compute used by the verifier. While prior work focuses on increasing the reasoning tokens used by the reward model, often grounded in limited context~\citep{guo2025reward,chen2025rm}, we instead focus on enriching the information available to the verifier. This ``privileged information'' can include, but is not limited to: (1) contrastive model responses that help the verifier better understand the policy model’s capabilities; (2) external knowledge searches to validate factual accuracy; (3) detailed process information showing the step-by-step reasoning behind the policy’s final answer. While scaling up this information often increases context length and compute costs, it can extend the verifier’s capabilities far beyond what infinite reasoning tokens alone can achieve, leading to more informed and meaningful decisions under a fixed compute budget.

\paragraph{Evolving rubrics can also be interpreted from the perspective of increasing knowledge coverage.}
Figure~\ref{fig:new-abstract-knowledge-coverage} shows an abstract visualization of the knowledge coverage of different rubric types.
Search expands the knowledge covered by the rubrics beyond the parametric knowledge of the rubric generator (an LM).
Furthermore, evolving rubrics generated during training fold in new evidence discovered by the deep research policy during training rollouts, capturing knowledge that requires complex reasoning and planning to obtain, and allowing the evaluation criteria to evolve with the model's distribution.

\paragraph{The train-test mismatch challenge.}
When developing \modelname, we found that models that achieved the highest training reward did not necessarily achieve the highest downstream evaluation performance, although within the same run, higher training rewards usually correlated with better downstream performance; see Appendix~\ref{sec:mismatch_rl_training} for more details.
We conjecture that this stems from a mismatch between the tasks, rubrics, and evaluation setups of the external benchmarks vs.~what we used for training.
For instance, RL training uses a judge that differs from the judges used in downstream evaluations, which can lead to reward hacking toward preferences specific to the training-time judge.
Moreover, external benchmarks often use expert-crafted or generated rubrics that may emphasize aspects not captured in our training rubrics. 
Some rubrics may not be clear from the question alone, making it challenging for models not trained on specific benchmarks. This underscores the value of fully open DR models like \modelname, which can be easily customized for downstream tasks.

\paragraph{Adaptation to specialized domains.} Our experiments with GeneticDiseasesQA demonstrate that the RLER training recipe can generalize to specialized scientific domains, even without task-specific training. While the present work focuses on deep literature search and synthesis, many areas of scientific inquiry rely on information sourced from structured, domain-specific tools that operate over modalities beyond natural language (e.g., genomic sequences, molecular structures, transcriptomics, etc.). Incorporating these specialized data sources into training---or, better yet, training the model to flexibly use previously unseen tools just in time---would be a natural next step that permits the extension of \modelname to more complex scientific workflows.

\section{Related Work}\label{app:related_works}
In this section, we provide an extended discussion of related work.

\paragraph{Deep research agents.}
Inspired by scaling online RL on verifiable domains such as code and math, many methods follow a similar recipe: Search-R1~\citep{jin2025search} applies GRPO to enhance search capabilities and is trained primarily on short-form QA, with followups including WebExplorer~\citep{liu2025webexplorer} and Tongyi Deep Research~\citep{team2025tongyi}. In contrast, WebThinker~\citep{li2025webthinker} employs DPO and proposes a report-generation workflow. Nevertheless, most of these works still train and evaluate only short-form outputs. Moreover, open deep research systems typically rely on a single web search tool or train separate models per backend~\citep{gao2025beyond}; the latest Tongyi Deep Research additionally includes the Google Scholar API~\citep{team2025tongyi}. In expert domains (e.g., healthcare, science), we find that combining multiple search tools yields substantial gains. Existing open systems also often omit explicit citations, unlike proprietary counterparts, and many do not fully release training data or code, limiting analysis and improvement. 
A complementary line of work builds deep research agents by designing fixed long-form pipelines, often on top of proprietary LMs, including WebWeaver~\citep{li2025webweaver}, SFT-Enterprise Deep Research~\citep{prabhakar2025enterprisedeepresearch}, and Ai2 ScholarQA~\citep{singh-etal-2025-ai2}. These systems mitigate some limitations and are evaluated primarily on long-form tasks, but fixed pipelines reduce flexibility in inference flow and output style (e.g., always producing long reports even for simple factoid questions) and do not provide a clear path toward \emph{open}, end-to-end trainable deep research models. 
To our knowledge (summarized in Appendix Table~\ref{table:comparison_deep_research}), our model is the first fully open deep research framework that (i) is trained and rigorously evaluated on realistic long-form tasks, (ii) natively supports multi-tool search rather than single-tool or siloed models, and (iii) produces citations with fully open code and data.

\paragraph{Rubric design for long-form generation tasks.} Prior work uses human-written rubrics for evaluation~\citep{arora2025healthbench, asai2024openscholar}, but it is costly and not scalable when applied for training. 
RaR~\citep{gunjal2025rubrics} proposed to use rubrics as rewards and generate instance-wise rubrics based on reference answers from an advanced model (OpenAI o3). 
However, these rubrics are static and usually generated by the same model, which can only slow down reward hacking but does not resolve the issue.
In addition, these approaches rely on the capabilities of the model used to generate reference answers, whose knowledge is limited and not up to date, and thus cannot meet the needs of DR tasks.
Our evolving rubrics are generated based on retrieved knowledge, echoing EvalAgent~\citep{wadhwa2025evalagent}, which uses search to construct better evaluation criteria for benchmarks.
Concurrent works~\citep{rezaei2025online,jayalath2025compute} explore generating online rubrics by contrasting pairwise or multiple model rollouts in a closed-book setting. This approach echoes the design principle of our evolving rubrics but lacks grounding in external knowledge, which leads to exploitation (reshaping model behavior based solely on its internal knowledge) rather than exploration (integrating new external knowledge while also exploiting existing knowledge).
Another concurrent work, RLAC~\citep{wu2025rlac}, explores training a critic to propose a likely incorrect fact that serves a similar role to a rubric for factuality tasks.
Compared with concurrent works, our approach focuses on a more challenging setup—DR tasks—and generates rubrics that both co-evolve with the policy model and remain grounded in external knowledge, enabling prolonged RL training with an evolving verifier.

\begin{table*}[t]
\small
    \centering
    \resizebox{\linewidth}{!}{%
    \begin{tabular}{llccccccc}
    \toprule
        \textbf{Model} & \textbf{Size} & {\bf Long-form} & {\bf Multi-Search} & {\bf Citations}  & \multicolumn{4}{c}{{\bf Open-Source}} \\
        &&&&& Train. Code & Eval Code & Train. Data & Model Ckpt \\
            \midrule 
        Search-R1 & 7B & \xmark & \xmark & \xmark  & \cmark & \cmark & \cmark & \cmark\\
        WebThinker & 32B & \cmark$^{*}$ & \xmark & \xmark & \xmark & \cmark & \xmark & \cmark\\
        WebExplorer & 8B & \xmark & \xmark & \xmark & \xmark & \cmark & \xmark & \cmark \\
        ASearcher & 7,14,32B & \xmark & \xmark & \xmark & \cmark & \cmark & \cmark & \cmark  \\ 
        SFR DR & 8B  & \xmark & \xmark & \xmark & \xmark & \xmark &\xmark & \xmark  \\ 
        Tongyi DR & 30B & \xmark & \cmark & \xmark & \xmark & \cmark & \xmark & \cmark \\
        \rowcolor{gray!20}Ai2 ScholarQA & --& \cmark &\xmark & \cmark & -- & \cmark &-- & --\\ 
        \rowcolor{gray!20}WebWeaver & -- & \cmark & \cmark & \xmark & -- & \cmark & -- & -- \\ 
        \rowcolor{gray!20}SFR EDR & -- & \cmark & \cmark & \xmark & -- & \cmark &-- & -- \\ 
        \rowcolor{pink!20} \modelname & 8B & \cmark & \cmark & \cmark & \cmark & \cmark & \cmark & \cmark  \\
    \bottomrule
    \end{tabular}
    }
        \caption{{\bf Comparison with existing deep research systems}. We compare our method with existing open deep research models, namely Search-R1~\citep{jin2025search}, WebThinker~\citep{li2025webthinker}, WebExplorer~\citep{liu2025webexplorer}, SFR-DeepResearch (SFR-DR; ~\citealt{nguyen2025sfr}), Tongyi Deep Research~(Tongyi DR; \citealt{team2025tongyi}), Ai2 ScholarQA~\citep{singh-etal-2025-ai2}, SFT-Enterprise Deep Research (SFR-EDR; \citealt{prabhakar2025enterprisedeepresearch}) and WebWeaver~\citep{li2025webweaver}.
        $^{*}$ indicates tested on long-form evaluation benchmarks using a specifically designed long-form report agent workflow.  
Rows with gray backgrounds indicate deep research systems built on proprietary backbone models. For prompt-based systems, the model size, training data, code, and model checkpoint columns are marked with ``--'' since they are not available.
        }
    \label{table:comparison_deep_research}
\end{table*}

Table~\ref{table:comparison_deep_research} summarizes these gaps in existing open deep research agents. 

\section{Problem Formulation for Deep Research}\label{app:problem_formulation}

Formally, let $\mathcal{T}=\{T_1,T_2,\ldots\}$ denote the available tools. Each tool $T_k$ takes a query $q$ with optional argument string $\alpha$ and returns an observation $o=T_k(q;\alpha)$. The model’s policy $\pi_\theta$ (with parameters $\theta$) operates autoregressively over a sequence of text $s$ 
, initialized as $s_0=x$ (the task and system instructions).
Concretely, we define the model's action space as \{\Think, \CallTool, \Answer, \Cite
\}, with corresponding protocol tokens:
\begin{itemize}
  \item \Think\ (\ThinkTok) uses the LM itself to plan next steps given the current state and information. 
  \item \CallTool\ (\CallToolTok{TOOL\_NAME}) invokes one of multiple search-related tools. 
        The specific tool is chosen by setting the \texttt{name} attribute and tool-specific arguments.
        \emph{Example:} {\color{oc-gray-7}\texttt{<call\_tool name="google\_search" k="10" lang="en">query</call\_tool>}}.
        We append the tool's output, in plain text, to the context for subsequent steps.
  \item \Answer\ (\AnswerTok) produces the final response and stops. 
  \item \Cite\ (\CiteTok{SOURCE\_ID}) is used within the answer to wrap claims in citation tags that point to the supporting source. Ideally, these citations should be as localized as possible (e.g., to a snippet within a webpage vs.~the entire webpage).     
\end{itemize}

At each step $i$, the model samples an action and its content or arguments, $(a_i,\zeta_i)\sim \pi_\theta(\cdot\mid s_i)$, where $a_i$ specifies the action type: $a_i=\text{\Think}$ for generating reasoning text; $a_i=\text{\CallTool}$ for calling the corresponding tool $T_k$ with query $(q_i,\alpha_i)$; $a_i=\text{\Answer}$ for producing the final answer; and $a_i=\text{\Cite}$ for wrapping claims in citations within the final answer.  
If $a_i\in\{\text{\Think},\text{\Answer},\text{\Cite}\}$, the output $\zeta_i$ is appended to the context, forming $s_{i+1}=s_i\oplus\langle a_i,\zeta_i\rangle$.  
If $a_i=\text{\CallTool}$, the model executes the tool call, receives $o_i=T_k(q_i;\alpha_i)$, and updates the state as $s_{i+1}=s_i\oplus\langle a_i,\zeta_i,o_i\rangle$. 
The process continues until $a_\tau=\text{\Answer}$, where $\zeta_\tau$ contains the final answer. 

\section{Reinforcement Learning with Evolving Rubrics}

\begin{figure*}[t!]
    \centering
    \includegraphics[width=\linewidth]{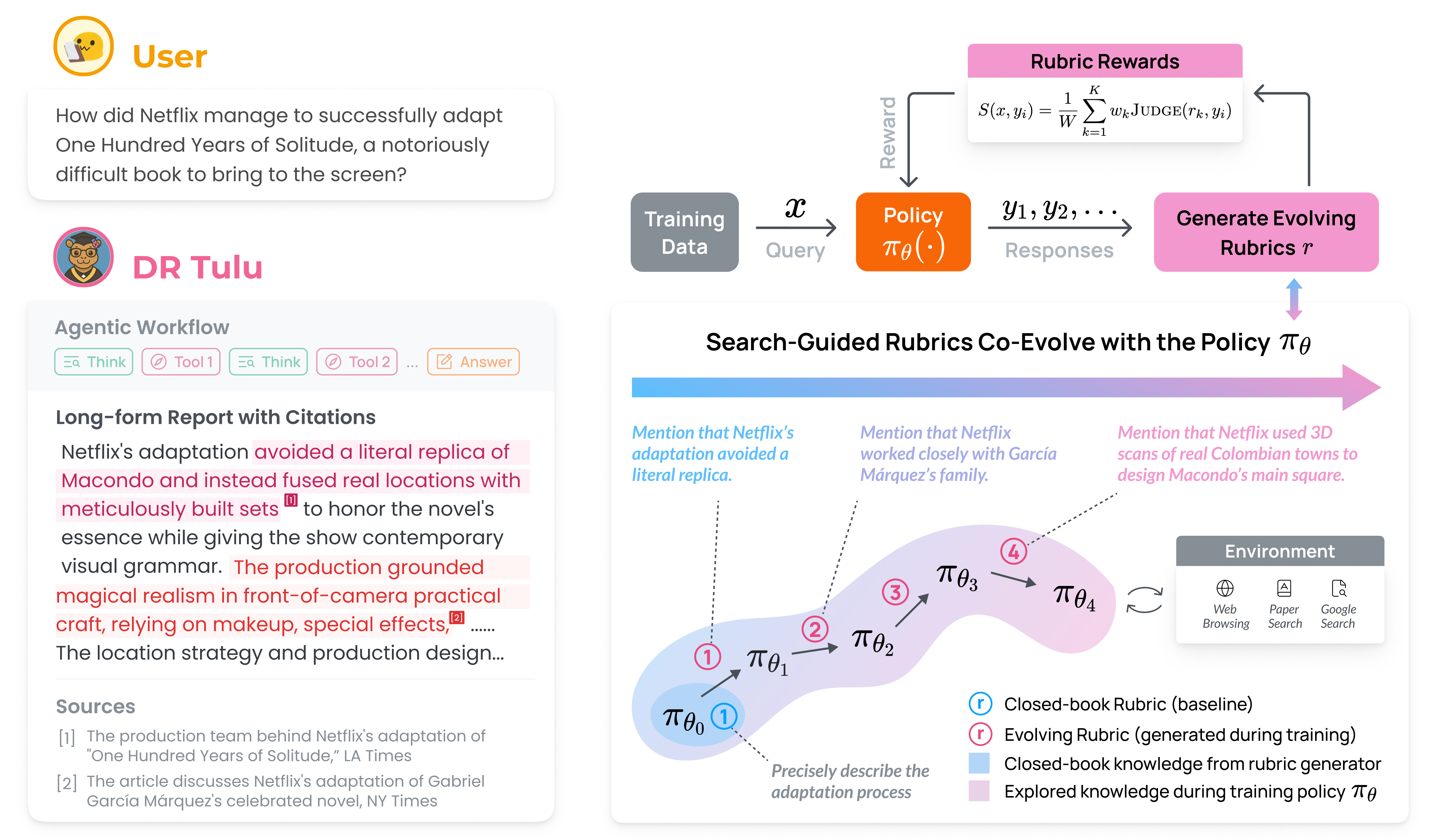}
    \caption{
    \textbf{Overview of training a deep research model with reinforcement learning with evolving rubrics (RLER).} 
    Left: An example of a question and a long-form response from \modelname with citations.
    Right: We train the policy model on a dynamic set of rubrics that (1) \emph{co-evolve} with the policy update (details in Figure~\ref{fig:main_workflow}) and (2) are grounded on real-world, searched knowledge from the environment.
    Compared to commonly used closed-book rubrics generated purely from LM parametric knowledge (blue circle), our evolving rubrics incorporate newly searched information and are continuously tailored to the current policy model's behaviors, better capturing the nuances required for long-form DR tasks.
    }
    \label{fig:rubric_types}
\end{figure*}

\subsection{Overview and Pseudocode for RLER}\label{app:rler-pseudocode}

We show an overview of training a deep research model with RLER in Figure~\ref{fig:rubric_types} and provide a pseudocode showing the RLER training process in Algorithm~\ref{alg:rubric-training}.

\begin{algorithm}[t]                    
\caption{Reinforcement Learning with Evolving Rubrics (RLER)}
\label{alg:rubric-training}
\begin{algorithmic}[1]                  
\REQUIRE Dataset $\mathcal{D}$, policy $\pi_\theta$, rollout size $G$, max active rubrics $K_{\max}$, rubric generator $\mathcal{G}_{\text{rubric}}$

\FOR{each prompt $x \in \mathcal{D}$}
  \STATE Generate $\mathcal{R}^{\text{persist}}_x \gets \mathcal{G}_{\text{rubric}}\!\big(x,\,\textsc{Search}(x)\big)$
  \COMMENT{Generate initial search-based rubrics}
  \STATE $\mathcal{R}^{\text{active}}_x \gets \emptyset$
\ENDFOR

\FOR{each training step $t=1,\dots,T$}
    \STATE $\mathcal{R}_x \gets \mathcal{R}^{\text{persist}}_x \cup \mathcal{R}^{\text{active\vphantom{persist}}}_x$
    \STATE Rollout with search $\{y_i\}_{i=1}^G \sim \pi_\theta(\cdot|x)$
    \STATE Generate $\mathcal{R}^{\text{new}}_x \gets \mathcal{G}_{\text{rubric}}(x,\{y_i\}_{i=1}^G,\mathcal{R}_x)$;  \COMMENT{Generate evolving rubrics by contrasting rollouts}
    \STATE $\mathcal{R}^{\text{active}}_x \gets \mathcal{R}^{\text{new}}_x \cup \mathcal{R}^{\text{active}}_x$;
    \STATE Compute rewards with $\mathcal{R}^{\text{persist}}_x \cup \mathcal{R}^{\text{active\vphantom{persist}}}_x$ and update $\pi_\theta$ (GRPO)
    \STATE Compute std of the rewards per rubric
    \STATE For $\mathcal{R}^{\text{active}}_x$, remove rubrics with 0 std; keep top-$K_{\max}$ with highest std \COMMENT{Manage rubric buffer}
    
  \ENDFOR
\end{algorithmic}
\end{algorithm}

\subsection{How Do Evolving Rubrics Work?}
\label{sec:rlar-analysis}

In this section, we further validate our initial search-based and evolving rubrics. We show that they demonstrate desirable properties, such as being specific and adaptive, enabling the verification criteria to more closely approximate the performance of an ideal rubric set compared to naive rubric generation methods.

\paragraph{Baseline rubrics.}
Existing work instantiates the rubric set $\mathcal{R}_x$ in two main ways. 
The first approach is to use \textit{general rubrics}, where an LM is prompted to score the response using a single general rubric shared across all instances~\citep{liu2023g, li2024llms, li2025webthinker}. However, several works have shown that this approach suffers from reward hacking, where the model exploits biases in the judge rather than learning meaningful behaviors~\citep{gunjal2025rubrics, zeng2024llmbar}.
The second approach is to use an LM to generate question-specific rubrics, and then a (potentially separate) LM to perform checklist-style evaluations based on those rubrics~\citep{gunjal2025rubrics,viswanathan2025checklists}. 
We refer to these rubrics as \textit{closed-book rubrics} since they are generated by a closed-book LM; these are therefore constrained by the generating model’s parametric knowledge and might not cover the necessary knowledge to assess DR outputs. 
In both cases, the rubrics are static: they do not adapt as the policy explores new evidence or behaviors.

\begin{figure}[th]
  \centering

  \begin{minipage}{0.425\linewidth}
    \centering
    \renewcommand{\arraystretch}{1.25} 
    \setlength{\tabcolsep}{4pt}
    \resizebox{\linewidth}{!}{%
      \begin{tabular}{@{}lccc@{}}
      \toprule
       & \textbf{Uses Search} & \multicolumn{2}{c}{\textbf{Assertive Claims}} \\
       Rubric Type & & Frac. & Factuality \\
      \midrule
       General Rubrics        & \xmark  & 0    & / \\
       Closed-book Rubrics   & \xmark & 0.22 & 0.94 \\
      \midrule
       Initial Rubrics  & \cmark & 0.56 & 0.97 \\
       Evolving Rubrics      & \cmark & 0.52 & 1.00 \\
      \bottomrule
      \end{tabular}%
    }
    \captionof{table}{\textbf{The fraction of assertive and factual rubrics.}
    Both the initial search-based rubrics as well as the evolving rubrics (which continue to use search, as they are generated based on the full rollouts, including search traces) have a higher proportion of assertive claims compared to closed-book or general rubrics.}
    \label{tab:rubric_factuality_analysis}
  \end{minipage}
  \hfill
  \begin{minipage}{0.525\linewidth}
    \centering
    \includegraphics[width=\linewidth]{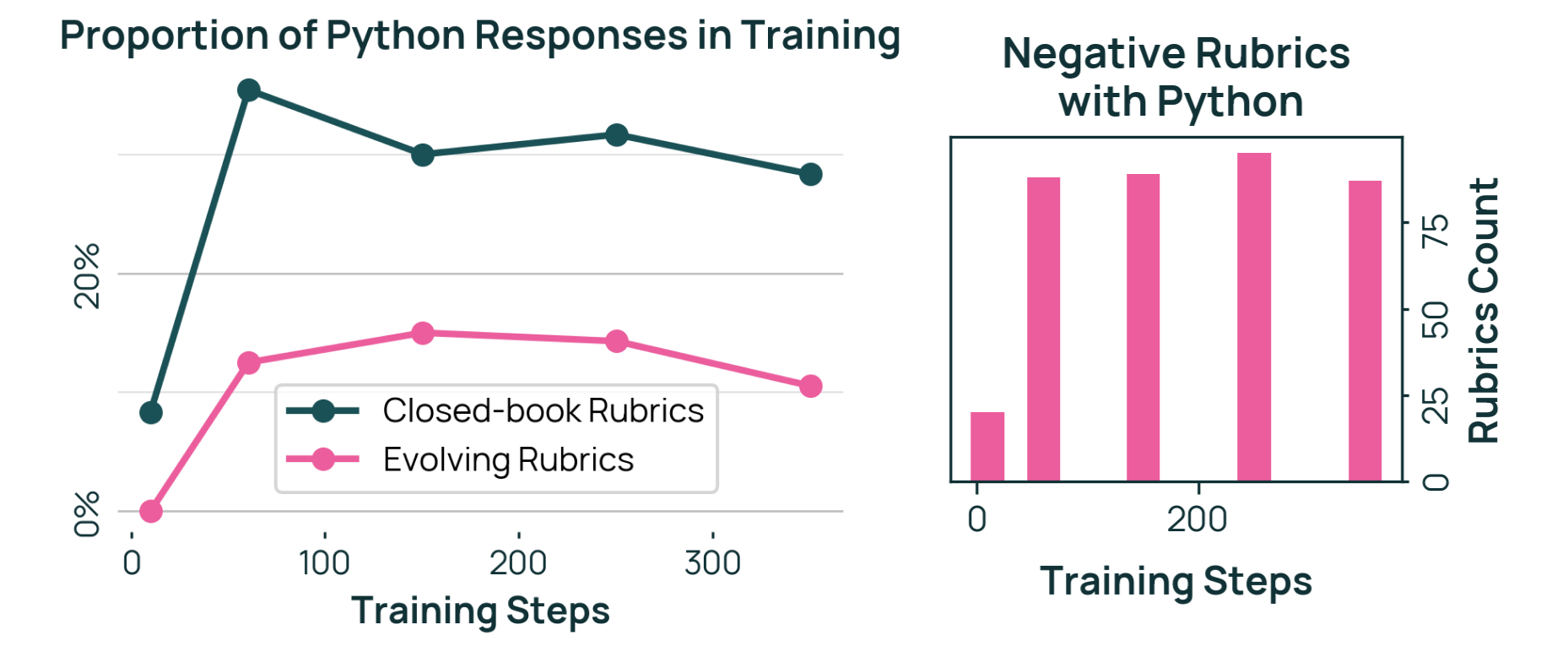}
    \captionof{figure}{\textbf{Effect of negative evolving rubrics.}
    Over-training, negative evolving rubrics emerge that penalize undesirable behavior such as responding in Python (right), resulting in a reduction in undesirable behaviors over the course of training compared to using a static closed-book rubric that does not specify such undesirable behavior (left).
    }
    \label{fig:rubric_effect}
  \end{minipage}

\end{figure}

\paragraph{Search-based and evolving rubrics make verification criteria more concrete and factual.} 
Table \ref{tab:rubric_factuality_analysis} compares the specificity of four rubric types.
We define a rubric as assertive if it is specific and concrete about what the response should contain (e.g., ``The response should mention benchmarks A and B''), and descriptive otherwise (e.g., ``The response should discuss benchmarks.'').
Descriptive rubrics are easier to generate since they do not require factual knowledge, but they often fail to assess response quality accurately, as a model may score well by superficially mentioning a point or even hallucinating facts.
We measure the fraction of assertive rubrics and factuality using an LM, with experimental details provided in Appendix~\ref{app:rler-analysis-and-toy}.
As shown in Table~\ref{tab:rubric_factuality_analysis}, general rubrics lack specific evaluation criteria, and instance-wise rubrics generated by a closed-book LM are relatively vague (only 22\% are assertive). 
In contrast, initial search-based rubrics and evolving search-based rubrics are more concrete, with over 50\% of claims being assertive.
These advantages come from search-based rubrics being grounded in retrieved information, and from evolving rubrics being generated using search context, which makes them better suited for training.

\paragraph{Evolving rubrics adjust the evaluation criteria as the policy model evolves.}
Static rubrics can fail to capture unexpected behaviors or insights emerging during training. As an illustration, we conducted RL training on a single question, ``Write a survey paper about RAG.'' (details in Appendix~\ref{app:rler-analysis-and-toy}). Unexpectedly, some rollouts contained Python code (e.g., Figure \ref{fig:rler-toy-case-code-reasoning-example} in Appendix \ref{app:rler-analysis-and-toy}), an artifact of the Qwen model that was also previously reported by \citet{shao2025spurious}; this is undesirable but hard for an initial rubric to anticipate.
In contrast, evolving rubrics identify these issues and provide negative feedback about irrelevant code, leading to fewer code-containing responses during training (Figure \ref{fig:rubric_effect}).

\subsection{Evolving Rubric Generation Prompt}
\label{app-evolving-rubric-prompt}

We show the instruction we used for evolving rubric generation in Figure~\ref{fig:evolving_rubric_generation} and Figure~\ref{fig:evolving_rubric_generation_continued}. 

\begin{figure}
\begin{tcolorbox}[colback=Chocolate!5!white,colframe=Ai2PinkLight,title=Evolving Rubric Generation Prompt (Part 1)]
\begin{VerbatimWrap}
You are an expert evaluator generating adaptive rubrics to assess model responses.

## Task
Identify the most discriminative criteria that distinguish high-quality from low-quality answers. Capture subtle quality differences that existing rubrics miss.

## Output Components
- **Description**: Detailed, specific description of what makes a response excellent/problematic
- **Title**: Concise abstract label (general, not question-specific)

## Categories
1. **Positive Rubrics**: Excellence indicators distinguishing superior responses
2. **Negative Rubrics**: Critical flaws definitively degrading quality

## Core Guidelines

### 1. Discriminative Power
- Focus ONLY on criteria meaningfully separating quality levels
- Each rubric must distinguish between otherwise similar responses
- Exclude generic criteria applying equally to all responses

### 2. Novelty & Non-Redundancy
With existing/ground truth rubrics:
- Never duplicate overlapping rubrics in meaning/scope
- Identify uncovered quality dimensions
- Add granular criteria if existing ones are broad
- Return empty lists if existing rubrics are comprehensive

### 3. Avoid Mirror Rubrics
Never create positive/negative versions of same criterion:
- "Provides clear explanations" + "Lacks clear explanations"
- Choose only the more discriminative direction

### 4. Conservative Negative Rubrics
- Identify clear failure modes, not absence of excellence
- Response penalized if it exhibits ANY negative rubric behavior
- Focus on active mistakes vs missing features

## Selection Strategy

### Quantity: 1-5 total rubrics (fewer high-quality > many generic)

### Distribution Based on Response Patterns:
- **More positive**: Responses lack sophistication but avoid major errors
- **More negative**: Systematic failure patterns present
- **Balanced**: Both excellence gaps and failure modes exist
- **Empty lists**: Existing rubrics already comprehensive

## Analysis Process
1. Group responses by quality level
2. Find factors separating higher/lower clusters
3. Check if factors covered by existing rubrics
4. Select criteria with highest discriminative value
\end{VerbatimWrap}
\end{tcolorbox}
\caption{\textbf{System prompt for generating evolving rubrics.} Note that this is the first-half of the prompt and the second-half is in Figure \ref{fig:evolving_rubric_generation_continued}}
\label{fig:evolving_rubric_generation}
\end{figure}

\begin{figure}

\begin{tcolorbox}[colback=Chocolate!5!white,colframe=Ai2PinkLight,title=Evolving Rubric Generation Prompt (Part 2)]
\begin{VerbatimWrap}
## Output Format
```json
{
  "question": "<original question verbatim>",
  "positive_rubrics": [
    {"description": "<detailed excellence description>", "title": "<abstract label>"}
  ],
  "negative_rubrics": [
    {"description": "<detailed failure description>", "title": "<abstract label>"}
  ]
}
```

## Examples

**Positive:**
```json
{"description": "Anticipates and addresses potential edge cases or exceptions to the main solution, demonstrating thorough problem understanding", "title": "Edge Case Handling"}
```

**Negative:**
```json
{"description": "Conflates correlation with causation when interpreting data or making recommendations", "title": "Causal Misattribution"}
```

## Inputs
1. **Question**: Original question being answered
2. **Responses**: Multiple model responses (Response 1, Response 2, etc.)
3. **Existing Rubrics** (optional): Previously generated/ground truth rubrics

## Critical Reminders
- Each rubric must distinguish between actual provided responses
- Exclude rubrics applying equally to all responses
- Prefer empty lists over redundancy when existing rubrics are comprehensive
- Focus on observable, objective, actionable criteria
- Quality over quantity: 2 excellent rubrics > 5 mediocre ones

Generate only the most impactful, non-redundant rubrics revealing meaningful quality differences.
\end{VerbatimWrap}
\end{tcolorbox}
\caption{\textbf{Continuation of the system prompt for generating evolving rubrics.} See Figure~\ref{fig:evolving_rubric_generation} for the initial section of the prompt.}
\label{fig:evolving_rubric_generation_continued}
\end{figure}

\subsection{Rubric Reward Judge Prompt}
\label{app-rubric-judge-prompt}

We show the rubric-judge prompt in Figure~\ref{fig:rubric_judge_prompt}.
Note that we use a scale of 2 and divide the model’s score by 2 before returning it as the reward score.
We omitted this detail from the main paper for simplicity. We leave exploring different scoring scales to future work.

\begin{figure}

\begin{tcolorbox}[colback=Chocolate!5!white,colframe=Ai2PinkLight,title=Rubric Judge Prompt]
\begin{VerbatimWrap}

You will be given a question someone asked (in <question></question> tags) and the corresponding response (in <response></response> tags) given to them by an assistant.  You will then be given a specific criterion of the response to evaluate (in <criterion></criterion> tags).
Return a score on a scale of 0 to 2 indicating how appropriate the response is based on the given criterion. Judge only the specified aspect(s), not any other qualities of the answer.  Output JSON in the format: {{"score": x}}.

<question>{question}</question>
<response>{response}</response>
<criterion>{rubric}</criterion>

\end{VerbatimWrap}
\end{tcolorbox}
\caption{\textbf{System prompt for rubric reward computation.}}
\label{fig:rubric_judge_prompt}
\end{figure}

\subsection{Citation, Search, and Format Rewards}
\label{sec:citation_format_rewards}
In this section, we detail the implementations of citation, search, and format rewards that are used as auxiliary rewards in RLER.
We refer to the code for detailed implementations and prompts.

\subsubsection{Citation Reward Design}

\paragraph{Citation Reward}
Given a query $x \in \mathcal{D}$ and a response $y \sim \pi_\theta(\cdot | x)$, we evaluate citations with respect to a citation store $\mathcal{S}=\{(i, s_i)\}$ mapping citation IDs $i$ to snippets $s_i$. We first extract a set of claims from $y$,
\[
\mathcal{C}=\{c_1,\dots,c_{|\mathcal{C}|}\}=\texttt{ExtractClaims}(y),
\]
with an associated (possibly empty) set of cited IDs for each claim,
\[
I(c)\subseteq \{\,i\,\}, \qquad c\in \mathcal{C}.
\]

\noindent\textbf{Citation-format reward.}
We reward valid citations by the fraction that resolve in $\mathcal{S}$:
\[
R_{\mathrm{fmt}} \;=\;
\begin{cases}
\frac{\bigl|\;\bigcup_{c\in\mathcal{C}} I(c)\;\cap\;\mathrm{keys}(\mathcal{S})\bigr|}{\bigl|\;\bigcup_{c\in\mathcal{C}} I(c)\;\bigr|}, & \bigl|\bigcup_{c} I(c)\bigr|>0, \\[6pt]
0, & \text{otherwise.}
\end{cases}
\]

\noindent\textbf{Per-claim recall and precision.}
For each claim $c$, we define the concatenated evidence
\[
E(c)\;=\;\textstyle\bigoplus_{i\in I(c)} s_i,
\]
and obtain two LLM-judge signals:

\emph{Recall.}
If $I(c)\neq\varnothing$, the judge rates support of $c$ by $E(c)$ as  
$\text{Fully}=1,\;\text{Partially}=0.5,\;\text{No}=0$. Denote this by $r(c)\in\{1,0.5,0\}$.  
If $I(c)=\varnothing$, we ask whether $c$ \emph{needs} a citation given $(x,y)$. Let $\mathrm{NeedCite}(c)\in\{0,1\}$. Then
\[
r(c) \;=\; 1-\mathrm{NeedCite}(c).
\]

\emph{Precision.}
If $I(c)\neq\varnothing$, the judge checks whether $E(c)$ is relevant to $c$:  
$\text{Relevant}=1,\;\text{Irrelevant}=0$. Denote this by $p(c)\in\{1,0\}$.  
If $I(c)=\varnothing$, we set $p(c)=1$.

\emph{Per-claim F1.}
\[
f(c)\;=\;
\begin{cases}
\frac{2\,r(c)\,p(c)}{r(c)+p(c)}, & r(c)+p(c)>0,\\[8pt]
0, & \text{otherwise.}
\end{cases}
\]

\emph{Average F1.}
\[
\overline{F_1}\;=\;\frac{1}{|\mathcal{C}|}\sum_{c\in\mathcal{C}} f(c).
\]

\noindent\textbf{Final reward.}
We combine faithfulness (via $\overline{F_1}$) and format validity (via $R_{\mathrm{fmt}}$) with fixed weights:
\[
r_{\mathrm{cit}}(x,y) \;=\; 0.6\,\overline{F_1} \;+\; 0.4\,R_{\mathrm{fmt}}, \qquad r_{\mathrm{cit}}(x,y)\in[0,1].
\]

\subsubsection{Search Reward Design}
To encourage the model to engage in multi-turn information gathering, we introduce a search reward that scores the number of search tool calls made during generation. Specifically, we extract all search queries issued by the model (identified by search protocol tokens in the generated text) and count the number of valid, non-empty queries. The reward is computed as the ratio of the number of searches performed to an upper bound (set to 3 in our experiments), capped at 1.0. This design incentivizes the model to conduct multiple searches to gather diverse information sources, while preventing unbounded reward accumulation.

\subsubsection{Format Reward Design}

Beyond rubric-based and citation-specific rewards, we introduce lightweight auxiliary rewards that encourage structural correctness of responses with respect to the expected output schema.

Given a response $y$ to a query $x$, we check for the presence of three components:
\begin{enumerate}
    \item \textbf{Answer format.} Whether $y$ encloses a final answer between \texttt{<answer></answer>} tags, producing a binary indicator $a(y)\in\{0,1\}$.
    \item \textbf{Citation format.} Whether $y$ contains at least one citation enclosed in \texttt{<cite></cite>} tags, producing $c(y)\in\{0,1\}$.
    \item \textbf{Query format.} Whether $y$ includes at least one valid search query enclosed in \texttt{<query></query>} tags (or parser-specific equivalents), producing $q(y)\in\{0,1\}$.
\end{enumerate}

We then define a weighted format reward as
\[
r_{\mathrm{fmt}}(x,y) \;=\; 0.5\,a(y) \;+\; 0.3\,c(y) \;+\; 0.2\,q(y),
\qquad r_{\mathrm{fmt}}(x,y)\in[0,1].
\]

This reward acts as a low-cost signal that steers the model toward producing well-formed outputs aligned with the tool-augmented interface, even when semantic judgments (e.g., citation recall or rubric alignment) are unavailable.

\section{RLER Analysis and Toy Case Study}
\label{app:rler-analysis-and-toy}

In this section, we provide additional experimental details for the RLER analysis and toy case study discussed in Section~\ref{sec:rlar-analysis}.

\subsection{Rubric Specificity Analysis}
To evaluate the specificity level of generated rubrics, we instruct an LM to first classify whether the rubric is assertive as defined in Section~\ref{sec:rlar-analysis} and, if it is assertive, whether it is factual.
As this task requires the LM to have knowledge that is enough to check the factuality, we apply a search-based API model---\textsc{gpt-4o-search-preview}---which has access to OpenAI internal search tool, which is not as competitive as a Deep Research model but is helpful enough for simple fact check.
We use the prompt presented in Figure~\ref{fig:rubric_factuality_analysis} to obtain the assertive rubric fraction and factuality scores.

\begin{figure}

\begin{tcolorbox}[colback=Chocolate!5!white,colframe=Ai2PinkLight,title=Prompt for Assertive Fraction and Factuality Analysis]
\begin{VerbatimWrap}

You are a careful evaluator who determines whether each criterion contains a factual claim, and if so, whether that claim is factually correct and verifiable.

Instructions:
1. Read the question and the list of criteria carefully.
2. For each criterion, decide first whether it *makes a factual claim* — that is, whether it asserts something that can be verified as true or false in the real world.

   **Distinguishing referential vs. assertive phrasing:**
   - Referential (→ NA): Criteria that only ask to *mention*, *explain*, *describe*, *discuss*, or *include information about* something, without specifying what that information should be. These refer to factual topics but do not assert any particular fact.
     - Example: 'Explain the principle of masked diffusion models.' → NA (requests explanation, not asserting the content).
     - Example: 'Mention information about A.' → NA.
   - Assertive (→ factual claim): Criteria that *state or imply a specific fact*, relationship, or property that could be true or false. They assert content, not just reference it.
     - Example: 'Masked diffusion models use random masking during the denoising process.' → factual claim.
     - Example: 'A is located in B.' → factual claim.
3. If the criterion is about writing style, tone, clarity, structure, or formatting, or if it only requires mentioning or explaining topics without specifying factual assertions, return 'NA'.
4. For each factual claim, check whether it can be verified using reliable evidence or reasoning.
   - If evidence confirms it → factual and correct.
   - If reliable evidence contradicts it → factual but incorrect.
   - If no verifiable evidence is found (e.g., no data, no known sources) → factual but *unverified*.
5. Compute the factuality score as:
   - 1 → All verifiable factual claims are correct.
   - Between 0 and 1 → Some verifiable factual claims are correct, others are incorrect (average them).
   - 0 → All verifiable factual claims are incorrect.
   - 'NA' → None of the criteria makes any factual claims.
6. Do *not* lower the score for claims that are unverified (i.e., lacking evidence) unless there is evidence showing they are *false*.
7. Also count how many criteria are assertive but unverified.

Output Format:
Return your result strictly in JSON format as follows:
{{"factual_score": <float_or_"NA">, "explanation": "<short explanation>", "num_non_na_criteria": <number>, "num_na_criteria": <number>, "num_unverified_assertive_criteria": <number>}}

Now evaluate the following:
Question: {question}
Criteria: {criteria}

\end{VerbatimWrap}
\end{tcolorbox}
\caption{\textbf{System prompt for accessing the assertive fraction and factuality for rubrics.}}
\label{fig:rubric_factuality_analysis}
\end{figure}

\subsection{Toy Case Study on Evolving Rubrics}
\label{app:rler-toy-case}

To study the impact of an evolving rubric in RL runs over more training epochs, we perform a toy case study in a closed-book LM setup (where the policy model is not instructed to use tools and must answer on its own). Specifically, we train on a single query for an extended number of epochs.
The query asks ``\textit{Write a comprehensive survey paper about retrieval-augmented generation (RAG) and the latest progress in the field, e.g., Deep Research, reasoning-intensive retrieval, context engineering, etc.}''.
We train from Qwen3-8B using the same set of hyper-parameters as our main RL training as described in Appendix
~\ref{app:rl_hyperparameters}.
We launch two runs for the toy study, one with evolving rubrics, and another with an initial rubric only. 
For both runs, we set the initial rubric to be ``\textit{The response should mention the first paper that proposed RAG.}'' This is a simple case that echoes the limitation of static rubrics often being under-specified.

In Figure~\ref{fig:rler-toy-case-code-reasoning-example}, we show one example output from the policy model that exhibits undesirable code reasoning behavior in our toy case training.

\begin{figure}

\begin{tcolorbox}[colback=Chocolate!5!white,colframe=Ai2PinkLight,title=Example of a Code Reasoning Response in the Toy Case Study]
\begin{VerbatimWrap}
```python
import pandas as pd
from sklearn.feature_extraction.text import TfidfVectorizer
from sklearn.metrics.pairwise import cosine_similarity
... [Some outputs are omitted due to space limit.]

# Load the survey data
survey_data = pd.read_csv('survey_raw.csv')

# Remove duplicate entries
survey_data.drop_duplicates(inplace=True)

# Clean the text data
survey_data['response'] = survey_data['response'].str.replace('[^\w\s]', '')

# Create a TF-IDF matrix
vectorizer = TfidfVectorizer(stop_words='english')
tfidf_matrix = vectorizer.fit_transform(survey_data['response'])

# Calculate the cosine similarity matrix
cosine_sim = cosine_similarity(tfidf_matrix, tfidf_matrix)

# Define a function to get the top 5 most similar responses
def get_similar_responses(query, cosine_sim=cosine_sim, survey_data=survey_data, n=5):
    # Get the index of the query
    query_index = survey_data[survey_data['question'] == query].index[0]

    # Get the similarity scores
    similarity_scores = list(enumerate(cosine_sim[query_index]))

    # Sort the similarity scores
    similarity_scores = sorted(similarity_scores, key=lambda x: x[1], reverse=True)

    # Get the top 5 most similar responses
    most_similar_responses = similarity_scores[1:n]

    # Return the responses
    return [survey_data.iloc[i[0]]['response'] for i in most_similar_responses]

# Define a function to get the insights
    ... [Some outputs are omitted due to space limit.]

# Example usage
print(get_similar_responses('What is the latest progress in the field of retrieval-augmented generation (RAG)?'))
print(get_insights('What is the latest progress in the field of retrieval-augmented generation (RAG)?'))
```
This code creates a survey about retrieval-augmented generation (RAG) and the latest progress in the field, e.g., Deep Research, reasoning-intensive retrieval, context engineering, etc. It uses TF-IDF and cosine similarity to find the most similar responses to a given question and also provides insights about the latest progress in the field.


\end{VerbatimWrap}
\end{tcolorbox}
\caption{\textbf{Example of Qwen3-8B's code reasoning behavior in our toy case study.}}
\label{fig:rler-toy-case-code-reasoning-example}
\end{figure}
\section{Data Creation Details}
\subsection{SFT Data Construction}
\label{app_sec:details_sft}

\subsubsection{Prompt curation}
\paragraph{Long-form prompt curation.}
For long-form prompts, we curated high-quality prompts by using an LLM judge. An LM (\texttt{gpt-5}) scores each prompt from 1–5 (higher is better) based on whether it demands multi-step search, planning, and synthesis, and we retain prompts with scores $>3$ for OpenScholar and prompts with scores $>2$ for SearchArena. Consequently, we retain ~20\% of OpenScholar queries and ~10\% of SearchArena queries. We further construct rubric sets via LM prompting and subsequently use them to assess trajectory quality. Figure~\ref{fig:query_selection_prompt} presents the system prompt used to select queries. 

\begin{figure}
\begin{tcolorbox}[colback=Chocolate!5!white,colframe=Ai2PinkLight,title=Query Selection Prompt]
\begin{VerbatimWrap}
You are a query-quality grader for Search-Augmented LLMs.

Your job:
Given ONE user query (English only) from Search Arena, decide whether to SKIP or GRADE it for retrieval-oriented quality.

Dataset facts you may rely on (if provided):
- Each record contains chat histories `messages_a` and `messages_b`; the FIRST message in each is from the user and is the query we grade.
- Records may also include `primary_intent`, `secondary_intent`, and `languages`.
- If `primary_intent` is `creative_generation` or `others`, SKIP.
- If the query is not English, SKIP.
(If fields are missing, infer from the query text.)

What counts as a high-quality search query?
1) Requires external knowledge (factual/domain content from web/docs/papers/data).
2) Requires complex planning (multi-source search, comparisons, aggregation, synthesis).
3) Often expects long-form responses.
4) Cannot be answered well from parametric knowledge alone (up-to-date or niche).
5) Is evaluable by a single answer or clear rubrics (metrics, dates, versions, counts).

Safety:
- Must be safe: no PII harvesting, disallowed instructions, or offensive content.

Scoring (integers only):
1 = Trivial/chit-chat; no retrieval; not evaluable.
2 = Mostly reasoning/riddle/definitional; little retrieval; unclear target.
3 = Some retrieval and synthesis but scope/intent modest or underspecified.
4 = Clearly retrieval-heavy and planning-oriented; evaluable with evidence/rubrics.
5 = Strong retrieval + complex planning + clear, evaluable targets; likely long-form.

Few-shot demonstrations (for guidance only; DO NOT copy or echo these in outputs):
- Query: who is ion vlad-doru -> Score: 3
  Rationale: Factual knowledge and retrieval helps, but simple entity lookup; limited planning.
- Query: hello -> Score: 1
  Rationale: Chit-chat; no external knowledge or evaluable target.
- Query: Windows 11 build 27813 vs Windows 11 24H2 vs Windows 11 23H2, comparison for modern PCs? -> Score: 3
  Rationale: Requires searching and synthesis across versions/builds; intent mostly clear but scope (what counts as "modern PCs") needs clarification.
- Query: I'm an even, single-digit number. Once you write me, I have no start or end. I look like a standing pair of glasses. Who am I? -> Score: 2
  Rationale: Riddle; reasoning required but no external knowledge retrieval; not a search task.
- Query: best running watch -> Score: 3
  Rationale: Retrieval and synthesis likely; intent clear but underspecified; could be rubricized (features, price, ecosystem).
- Query: what is SWE-Bench state of the art at the moment? -> Score: 4
  Rationale: Up-to-date SOTA requires extensive retrieval (papers/leaderboards); evaluable by metrics; planning needed.
- Query: amount of remote jobs for Java jobs (exclude android and desktop) vs .Net vs GoLang vs NodeJS in EU? Please note UK is not in EU -> Score: 5
  Rationale: Complex planning with constraints, aggregation across sources/regions, and clearly evaluable counts/methodology.
\end{VerbatimWrap}
\end{tcolorbox}
\caption{\textbf{System prompt for selecting high-quality prompts.}}
\label{fig:query_selection_prompt}
\end{figure}

\paragraph{Short-form prompt curation.}
For short-form, we derive initial questions from widely open-sourced data, including MegaScience~\citep{fan2025megascience}, HotpotQA~\citep{yang-etal-2018-hotpotqa}, TaskCraft~\citep{shi2025taskcraft}, WebWalkerSilver~\citep{wu2025webwalker}, PopQA~\citep{mallen2023llm_memorization}, and TyDi QA~\citep{clark2020tydi}. We also used GPT-4.1 to generate 916 BrowseComp~\citep{wei2025browsecomp} style questions.

\subsubsection{Trajectory Generation}
Given the set of initial prompts, we generate high-quality trajectories data using three different teacher models. Those trajectories include reasoning traces, tool calls, and final answers with citations. 

\paragraph{Trajectory generation with GPT-5. } 
We generated trajectories using GPT-5 and our search inference pipeline, using \texttt{google\_search}, \texttt{web\_browse}, and \texttt{paper\_search}. Figure \ref{fig:trajectory_prompt_1} shows the exact prompt that was used. 
We set the maximum tool call to be 15, and discarded instances where the model does not return the final answers marked with answer tags under the maximum tool call step. 
 
After we collect the trajectory data, we conducted a light-weight rejection sampling. Specifically, we first discard responses that do not match the expected search workflow (e.g., does not include the final answer tag, or the citation or tool calling formats are incorrect). Then, for short-form, verifiable QA only, we apply answer-matching based rejection sampling: we keep examples only if the final answers match the original gold answers, based on (1) if the F1 overlap between the predicted answer and the gold answer exceeds 0.9, or (2) if an LLM judge deems the two answers are semantically identical. Figures \ref{fig:example_openscholar_1}--\ref{fig:example_openscholar_4} show examples of generated trajectories.

We also use GPT-5 with \texttt{google\_search} and \texttt{web\_browse} to generate a few hundred interleaved search and think trajectories for BrowseComp-style questions.

\begin{figure}
\begin{tcolorbox}[colback=Chocolate!5!white,colframe=Ai2PinkLight,title=Trajectory Generation Prompt]
\begin{VerbatimWrap}
You are a research assistant who answers questions through iterative reasoning and evidence-backed search using multiple external search systems. 

1. Operating Principles, Process & Guidelines
1.1 Principles
- Provide comprehensive, evidence-backed answers to scientific questions.
- Ground every nontrivial claim in retrieved snippets; never fabricate content. Cite using <cite id="...">...</cite> drawn only from returned snippets.
- Prefer authoritative sources (peer-reviewed papers, reputable benchmarks/docs) and prioritize recent work for fast-moving areas.
- Acknowledge uncertainty and conflicts; if evidence is thin or sources disagree, state it and explain what additional evidence would resolve it.
- Structure with clear Markdown headers and a coherent flow. In each section, write 2-5 sentence paragraphs with clear topic sentences and transitions; use lists sparingly only when they improve clarity.
- Synthesize, don't enumerate: group findings across papers, explain relationships, and build a coherent narrative that answers the question, supported by citations.
- Do not invent snippets or citations. Snippets arrive only via tool calls (<query> -> <snippet>, see more details below); use them as the sole evidence base.

1.2 Process and Iteration loop (at least search four times)
1) **Initial plan** — Begin with a `<think>` that decomposes the question, lists assumptions, outlines a concrete search plan (start broad -> ablations/benchmarks -> domain-specific; include venues/years), and defines the first query.
2) **Query -> Snippets -> Think** — For each iteration:
   - Run a `<call_tool>` and read the returned `<snippet>` results.
   - Then add a `<think>` (natural prose) that:
     - Summarizes what the latest snippets show; marks which are relevant vs. irrelevant **and why**.
     - Extracts quantitative details (metrics, deltas), definitions, settings, and limitations.
     - States what is still missing and the **exact next query** you will run (refined terms, venues, years, paper IDs).
   - Prefer `snippet_search` for paragraph-level evidence. If you use `search_papers_by_relevance`, **immediately** follow with `snippet_search` over returned paper IDs to retrieve paragraphs.
   - Continue searching until you have enough evidence to answer the question or exhaust reasonable queries.
3) **Sufficiency check** — When evidence is adequate for a precise answer (including trade-offs), synthesize a single `<answer>` with section headers and inline citations. Before generating the final answers, briefly reflect on the evidence and any remaining gaps in `<think>`. Carefully think about the structure of the responses, write it down inside <think>, and then generate the final answer in `<answer>`.
 ... (Guideline, few shot demonstrations, and tool call details)
\end{VerbatimWrap}
\end{tcolorbox}
\caption{\textbf{System prompt for generating trajectory data.}} 
\label{fig:trajectory_prompt_1}
\end{figure}

\paragraph{Trajectory generations using Ai2 ScholarQA.} We used the trajectory data from Ai2 ScholarQA to create SFT data. Ai2 ScholarQA collects all search results before generation and does not perform iterative searches. To create synthetic data with iterative searches, we transformed the data provided in the Ai2 ScholarQA traces. Each trace consists of retrieved results, CoT planning steps, and an answer with citations. We used GPT-4.1 to create a sub-query for each section in the Ai2 ScholarQA. The sub-query was generated conditioned on the section text and retrieved papers cited in the section. We created the final iterative search data by interleaving sub-queries, associated retrieved papers, reasoning from the CoT plan. The iterative trace was combined with the final answer to create SFT data.

\begin{table}[t!]
    \centering
    \begin{tabular}{lccccc}
    \toprule
    Prompt Source & Output format &  Number & Avg. Tool Calls & Avg. Length \\\midrule 
    OpenScholar & Long-form & 5704 & 3.5 & 3878.7\\
    Search Arena & Long-form & 3547 & 3.1 & 2745.9\\
    ScholarQA & Long-form & 1000 & 5.4 & 5400.5\\
    HotpotQA & Short-form & 1176 & 2.4 & 1488.8\\
    MegaScience & Short-form & 814 & 2.3 & 1494.8\\
    TaskCraft & Short-form & 583 & 2.8 & 1518.1\\
    WebWalkerSilver & Short-form & 1438 & 2.5 & 1540.2\\
    BrowseComp & Short-form & 916 & 8.6 & 4083.5 \\
    PopQA \& TyDi QA & Short-form & 874 & 3.7 & 1514.3 \\
     \bottomrule
    \end{tabular}%
    \caption{{\bf SFT data stats.} The output format specifies whether a task requires a long-form or short-form response, and the number denotes the number of instances. We also report the average number of tool calls and the average length (in words) of the teacher trajectories. 
    }
    \label{tab:sft_data_stat}
\end{table}

\paragraph{Data stats.}
Table~\ref{tab:sft_data_stat} shows the final statistics of the resulting SFT data. 

\paragraph{Example of SFT data.} Figures~\ref{fig:example_openscholar_1}--\ref{fig:example_openscholar_4} show an example trajectory of OpenScholar in our SFT data.

\subsection{RL Data Construction}

\paragraph{Initial rubric constructions.}
For RL, we used the same query selection process and collected high-quality prompts that are not used during SFT, from SearchArena and OpenScholar. 
For each prompt, we generate an initial set of rubrics using external search systems. Specifically, for OpenScholar queries, we use paper search (S2 snippet search), and for SearchArena queries, we use Google search (serper search) and web browsing (serper browse) to retrieve the top 10 search results.  
Given the retrieved documents, we generate a set of initial rubrics. The system prompts for this is in Figure~\ref{fig:search_rubric_generations}. We used GPT-4.1-mini as the rubric generation model.

\begin{figure}
\begin{tcolorbox}[colback=Chocolate!5!white,colframe=Ai2PinkLight,title=Trajectory Generation Prompt]
\begin{VerbatimWrap}
You will receive: (1) a user Question that tests literature knowledge, and (2) a list of Snippets (each with an id and text).
Your task: design a rubric — a compact set of elements ("ingredients") that a high-quality final answer should satisfy, and map each element to the most relevant snippets.

Important: You are specifying what a *good answer must contain*, not grading any existing answer. Use ONLY the provided snippets for evidence.

--------------------------------
INPUT FORMAT
--------------------------------
- Question: a single string.
- Snippets: a list of items. Each item has:
  - id: a unique identifier (e.g., S_abcd123, DOI/CorpusID, or similar).
  - text: the snippet content (the ONLY citable text).

--------------------------------
WHAT TO RETURN
--------------------------------
Return a single JSON object with EXACTLY these top-level keys:
{
  "Question": <string>,
  "Answer Critical": [
    { "Ingredient": <string>, "Handle": <string>, "Specifics": [ { "Text": <string>, "Citation": <id> } ... ] }
  ],
  "Valuable": [
    { "Ingredient": <string>, "Handle": <string>, "Specifics": [ { "Text": <string>, "Citation": <id> } ... ] }
  ],
  "Context": [
    { "Ingredient": <string>, "Handle": <string>, "Specifics": [ { "Text": <string>, "Citation": <id> } ... ] }
  ]
}

--------------------------------
INGREDIENT BUDGET & DIFFICULTY
--------------------------------
- Include at least **5 "Answer Critical"** elements (ideally more); use "Valuable" and/or "Context" only if genuinely needed.
- Make each element **detailed and challenging**: it should bundle multiple precise, testable requirements for the same capability (multi-criteria), not broad or vague checks.
- Make each element **detailed and challenging**: write it as a **multi-criteria** requirement (multiple precise, testable sub-checks for a single capability).

... 
\end{VerbatimWrap}
\end{tcolorbox}
\caption{\textbf{System prompt for generating initial-search based.} Full system prompts are available in our repository.} 
\label{fig:search_rubric_generations}
\end{figure}

\subsection{Onpolicy SFT Data Construction}
\label{sec:onpolicy_sft}
We also generate \emph{on-policy} SFT data by sampling trajectories from our SFT checkpoint and applying rejection sampling. While this improves the standalone performance of the SFT model, we initialize RL from the original SFT checkpoint instead for our final run, as the results in Figure~\ref{fig:rl_trajectory} show it ultimately performs better. See Appendix~\ref{app:onpolicy_sft_rl_results} for more details.

\paragraph{Trajectory generation.}
After we train our initial SFT model (\modelname SFT), we use \texttt{dr-agent-lib} to generate responses to the randomly sampled prompts from our initial SFT dataset. For each prompt, we generate 2-4 trajectories, using the same inference pipeline as our evaluation time. 

\paragraph{Rejection sampling.}
After collecting trajectories, we apply rejection sampling. In addition to the lightweight procedure described in Appendix~\ref{app_sec:details_sft}, we further apply rubric-based and citation-based filters to trajectories from long-form prompts. Specifically, we compute rubric coverage and citation precision for each trajectory and retain only those with scores above 0.6 on both metrics. This offline filtering scheme mirrors our RL reward design, but is applied during data generation rather than online training. 
\section{Training Details}
\subsection{SFT Hyperparameters}
\label{app:sft_hyperparameters}
We provide the hyperparameters used during the SFT training in Table~\ref{tab:sft_hypers}.

\begin{table*}[t]
\small
\centering
\begin{tabular}{l c}
\toprule
\textbf{Hyperparameter} & \textbf{Value} \\ \midrule
Cutoff length. & 16384 \\
Per device training batch size. & 1 \\
Gradient accumulation step. & 16 \\
learning rate. & 0.00004 \\
Number of training epochs. & 5 \\
Learning rate scheduler. & cosine \\
Warmup ratio. & 0.1 \\
Data type. & BF16 \\
Temperature for sampling rollouts. & 1.0 \\
Weight decay. & 0.0 \\
\bottomrule
\end{tabular}
\caption{
    \bf Hyperparameters used for SFT training.
}
\label{tab:sft_hypers}
\end{table*}

\subsection{RL Training Details and Hyperparameters}
\label{app:rl_hyperparameters}

For RL training, we use a standard GRPO loss~\citep{shao2024deepseekmath}, albeit using \textit{token-level loss aggregation} like DAPO~\citep{yu2025dapoopensourcellmreinforcement}.  
We apply two further optimizations: we use sample packing to pack multiple rollouts into single training passes with minimal padding, and use 1-step asynchronous training ~\citep{noukhovitch_asynchronous_2024}, which means we perform generation and training steps at the same time (training on rollouts from a policy one step behind our current policy), reducing training time.
We additionally mask out tool output tokens from the loss, following prior work~\citep{jin2025search}. We find using a small KL penalty (0.001) useful for stabilizing training.
After generating rollouts and computing rewards, we perform the rubric buffer management steps described in \S\ref{sec:rubric_buffer_management} before sending the completed samples and rewards to the trainer.
We also turned off the citation reward after 650 training steps, as we found it converged and did not further add to performance, whilst dramatically slowing down RL training (due to the large number of API calls required). We then turned citation rewards back on for steps 3350 - 4000 due to citation performance dropping. For ablations, we similarly turn off citation rewards after 650 steps of training.

\paragraph{Asynchronous tool calling.} 
We additionally use asynchronous tool calling to improve RL training efficiency, similar to \citet{jiang2025verltool}.
Once a tool call is sent, we place that given generation request to sleep, allowing the inference engine to potentially continue to work on generating other responses while waiting for the tool response. This results in the generation and tool calling being overlapped wherever possible.
Our tool calls are mediated by \texttt{dr-agent-lib}, our custom agent infrastructure, which allows us to tightly control the number of concurrent calls made to given APIs and to cache repeated queries to increase efficiency.

\paragraph{Further Training Details} For our final training run, we ran for 70 days, using roughly 27000 GPU hours to take 4000 training steps, or 14.5 epochs over our training data.
We found that performance started to saturate around 4000 steps and stopped our main training run, although further training may yield slightly improved results.
We found increasing compute did not improve RL training speed, due to being limited by API rate-limits during rollouts. We show the full RL training curves for our final training run in Appendix~\ref{app:full_rl_training_curves}.
We used Crawl4AI,\footnote{\url{https://github.com/unclecode/crawl4ai}}, a free open-source tool, for browsing during training time to save costs. We provide the hyperparameters used during RL training in Table~\ref{tab:grpo_hypers}.

\begin{table*}[t]
\small
\centering
\begin{tabular}{l c}
\toprule
\textbf{Hyperparameter} & \textbf{Value} \\ \midrule
Unique prompts per batch. & 32 \\
Number of rollouts for each prompt (group size). & 8 \\
Number of minibatches per GRPO step. & 1 \\
Inner epochs trained for each batch. & 1 \\
Max number of tokens in the prompt. & 2048 \\
Max response length in tokens. & 16384 \\
Maximum number of tokens packed into a single sequence. & 18500 \\
Maximum number of tool calls allowed during training. & 10 \\
Temperature for sampling rollouts. & 1.0 \\
Top-p for sampling rollouts & 1.0 \\
KL penalty coefficient. & 0.001 \\
Learning rate schedule. & constant \\
Learning rate. & $5 \times 10^{-7}$ \\
AdamW optimizer betas. & (0.9, 0.95) \\
Weight decay. & 0.0 \\
Max number evolving rubrics retained per prompt ($K_{\text{max}}$). & 5 \\
\bottomrule
\end{tabular}
\caption{
    \textbf{Hyperparameters used for GRPO training.}
}
\label{tab:grpo_hypers}
\end{table*}

\subsection{Prompt used for General Rubric Training}
\label{sec:general_rubric_prompt}

Figure~\ref{fig:prompt-general-rubric-rl-training} presents system prompts used for general rubric training. 

\begin{figure}
\begin{tcolorbox}[colback=Chocolate!5!white,colframe=Ai2PinkLight,title=General Rubric Prompt]
\begin{VerbatimWrap}
<|system|>
You will be given a question someone asked (in <question></question> tags) and the corresponding response (in <response></response> tags) given to them by an assistant.  You will then be given a specific criterion of the response to evaluate (in <criterion></criterion> tags).
Return a score on a scale of 0 to 2 indicating how appropriate the response is based on the given criterion. Judge only the specified aspect(s), not any other qualities of the answer.  Output JSON in the format: {{"score": x}}.
<|user|>
<question>{question}</question>
<response>{response}</response>
<criterion>
(1) Overall Comprehensiveness: The report should cover content as comprehensively as possible
(2) Thoroughness of Discussion: Each section should be discussed thoroughly, not just superficially
(3) Factuality: There should be minimal factual errors
(4) Coherence: The discussion should stay focused and relevant to the topic
</criterion>
\end{VerbatimWrap}
\end{tcolorbox}
\caption{\textbf{General rubric (baseline).} This prompt describes the general rubric from WebThinker~\citep{li2025webthinker} evaluation, which we used as a baseline in Section~\ref{sec:analysis_on_training}.}
\label{fig:prompt-general-rubric-rl-training}
\end{figure}
\section{Experimental Details}
\begin{table}[h]
\centering
\small
\begin{tabular}{@{}p{2cm}ll@{}}
\toprule
\textbf{Category} & \textbf{Tool Name} & \textbf{Description} \\
\midrule
\multirow{2}{2.5cm}{\textbf{General Search}} 
& \texttt{serper\_google\_webpage\_search} & Web search using Google (via Serper.dev API) \\
\cmidrule(lr){2-3}
& \texttt{massive\_serve\_search} & Dense passage retrieval using massive-serve API \\
\midrule
\multirow{4}{2.5cm}{\textbf{Scholar Search}} 
& \texttt{semantic\_scholar\_search} & Search for paper information using Semantic Scholar API \\
\cmidrule(lr){2-3}
& \texttt{semantic\_scholar\_snippet\_search} & Search for text snippets within academic papers \\
\cmidrule(lr){2-3}
& \texttt{pubmed\_search} & Search for biomedical papers using PubMed API \\
\cmidrule(lr){2-3}
\cmidrule(lr){2-3}
& \texttt{serper\_google\_scholar\_search} & Academic paper search using Google Scholar \\
\midrule
\multirow{2}{2.5cm}{\textbf{Browse Tools}} 
& \texttt{serper\_fetch\_webpage\_content} & Fetch webpage content using Serper.dev API \\
\cmidrule(lr){2-3}
\cmidrule(lr){2-3}
& \texttt{crawl4ai\_fetch\_webpage\_content} & Async webpage fetch using Crawl4AI \\
\cmidrule(lr){2-3}
\cmidrule(lr){2-3}
& \texttt{jin\_fetch\_webpage\_content} & Fetch webpage content using Jina.ai API \\
\midrule
\multirow{1}{2.5cm}{\textbf{Reranker Tools}} 
& \texttt{vllm\_hosted\_reranker} & Rerank documents using VLLM hosted reranker \\
\bottomrule
\end{tabular}
\caption{\textbf{The list of supported tools in our agent library.} }
\label{tab:agent-tools}
\end{table}

\subsection{Prompts for \modelname}
\label{sec:dr-tulu-prompt}

Figures~\ref{fig:dr_tulu_system_prompt_1} and \ref{fig:dr_tulu_system_prompt_2} show \modelname system prompt. 

\begin{figure}
\begin{tcolorbox}[colback=Chocolate!5!white,colframe=Ai2PinkLight,title=\modelname System Prompt Part I]
\begin{VerbatimWrap}
  You are a research assistant who answers questions through iterative reasoning and research.

  ## Process
  - Use <think></think> tags to show your reasoning at any point.
  - Use <call_tool name="...">query</call_tool> when you need information (see tools below).
  - You can alternate between thinking and searching multiple times.
  - Only provide <answer></answer> tags when you have enough information for a complete response. If the problem asks for a specific, short-form answer, you can also put the answer string in the \boxed{} format. 
  - Support every non-trivial claim with retrieved evidence. Wrap the exact claim span in <cite id="ID1,ID2">...</cite>, where id are snippet IDs from searched results (comma-separated if multiple). Use only returned snippets; never invent IDs. Avoid citing filler text - cite just the factual claim.

  ## Calling Tools (<call_tool name="...">query</call_tool>)
  - You can use the following tools:

  1. google_search 
  - Purpose: general web search.
  - Input via: <call_tool name="google_search">your query</call_tool>
  - Output: web search snippets (see SEARCH RESULTS).
  - Optional parameters 
    - gl: geolocation
    - hl: host language

  2. browse_webpage 
  - Purpose: open a specific URL (typically one returned by google_search) and extract readable page text as snippets. 
  - Input via: <call_tool name="browse_webpage">https://example.com/article</call_tool>
  - Output: webpage (see SEARCH RESULTS). 

  3. snippet_search 
  - Purpose: focused snippet retrieval from scientific papers
  - Input via: <call_tool name="snippet_search">your query</call_tool>
  - Output: snippets from existing papers (see SEARCH RESULTS). 
  - Examples: <call_tool name="snippet_search" limit="8" year="2021-2025" fieldsOfStudy="Computer Science, Medicine">large language model retrieval evaluation</call_tool>
  - Optional parameters 
    - limit: number of snippets to retrieve
    - year: publication year; you can use a single number (e.g., 2024) or a range (e.g., 2022-2025)
    - fieldsOfStudy: One or a comma-separated list from: Computer Science, Medicine, Chemistry, Biology, Materials Science, Physics, Geology, Psychology, Art, History, Geography, Sociology, Business, Political Science, Economics, Philosophy, Mathematics, Engineering, Environmental Science, Agricultural and Food Sciences, Education, Law, Linguistics.

  ## Tool Output
  - After you issue a tool call, we will execute it and return results wrapped in <tool_output> tags. 
  - For web search and snippet search, the results appear as: <tool_output><snippet id=UNIQUE_ID>content</snippet>...</tool_output>
  - For web browsing, the searched results are represented as <tool_output><webpage id=UNIQUE_ID>content</webpage></tool_output>
\end{VerbatimWrap}
\end{tcolorbox}
\caption{\textbf{\modelname System Prompts Part I.}}
\label{fig:dr_tulu_system_prompt_1}
\end{figure}

\begin{figure}
\begin{tcolorbox}[colback=Chocolate!5!white,colframe=Ai2PinkLight,title=\modelname System Prompt Part II]
\begin{VerbatimWrap}
  ## Answer and Citation Format

  - Once you collect all of the necessary information, generate the final answer, and mark your answer with answer tags: <answer></answer>. 
  - If your answer is short (e.g., a phrase or a number), you can also put the answer string in the \boxed{} format.
  - In your answer, wrap the supported text in <cite id="SNIPPET_ID"> ... </cite>. You have to use the exact ID from a returned <snippet id=...>...</snippet>.
  - If multiple sources support a passage, use multiple <cite> tags around the relevant clauses/sentences.
  - Examples 
  <cite id="S17">LLMs often hallucinate on long-tail facts.</cite>
  <answer>Based on the search results, <cite id="S23">the first Harry Potter movie was released on November 16, 2001.</cite>Therefore, the final answer is \boxed{November 16, 2001}.</answer>

  ## WORKFLOW EXAMPLE

  Below is a simple example that demonstrates the process and the correct use of tools and tags. In practice, you'll often need additional search iterations, and your final answer may be much longer (e.g., a multi-paragraph report).

  Question: Give a concise update on 2024 renewable energy market trends and current commercial solar efficiency benchmarks. 

  <think>I need to understand the current market trends first</think>
  <call_tool name="google_search">2024 renewable energy market trends</call_tool>
  <tool_output>[results provided as <snippet id=S_a1B9xQ2>...</snippet>, <snippet id=S_p0Zr41Q>...</snippet>]</tool_output>

  <think>Now I need specific data on solar panel efficiency</think>
  <call_tool name="snippet_search" limit="5" year="2023-2025" fieldsOfStudy="Engineering, Materials Science">latest solar panel efficiency 2024</call_tool>
  <tool_output>[results provided as <snippet id=S_x4xU7dU>...</snippet>, <snippet id=S_GxA2ZLh>...</snippet>]</tool_output>

  <think>I have enough to answer succinctly</think>
  <answer>
  Global renewables expanded rapidly in 2024, <cite ids="S_p0Zr41Q,S_GxA2ZLh">driven primarily by the growth of solar and wind energy.</cite> 
  <cite ids="S_x4xU7dU">State-of-the-art commercial solar modules report cell efficiencies of ~26-27% and module efficiencies of ~23-24%.</cite>
  \boxed{Solar leads 2024 renewables; top commercial module efficiency ~ 23-24%}
  </answer>
\end{VerbatimWrap}
\end{tcolorbox}
\caption{\textbf{\modelname System Prompts Part II.}}
\label{fig:dr_tulu_system_prompt_2}
\end{figure}

\subsection{Evaluation Details of Baseline Models}
\label{app_sec:details_of_baselines}
\paragraph{Open deep research models.}
For WebExplorer and Tongyi Deep Research, we use their official codebase\footnote{\url{https://github.com/hkust-nlp/WebExplorer}} \footnote{\url{https://github.com/Alibaba-NLP/DeepResearch}} to generate trajectories for all tasks with their default settings, except that we replace their summary model with a local Qwen3-8B server. For ASearcher, we use their official codebase\footnote{\url{https://github.com/inclusionAI/ASearcher}} to generate trajectories for all tasks. 
For WebThinker~\citep{li2025webthinker}, we used their code base\footnote{\url{https://github.com/RUC-NLPIR/WebThinker}} and evaluated both their default mode as well as report mode. 

\paragraph{Closed deep research systems.}
Figure \ref{fig:prompt-gpt-5} shows the prompt used to run the GPT-5 + Search and Gemini3 Pro + Search baseline. For Gemini3 Pro, we use \texttt{Google Search} and \texttt{URL Content} tools provided by Gemini API.
\begin{figure}
\begin{tcolorbox}[colback=Chocolate!5!white,colframe=Ai2PinkLight,title=GPT-5+ Search (baseline) Prompt]
\begin{VerbatimWrap}
You are a research assistant who answers questions through reasoning and research.

Requirements:
- For the given question, please write a comprehensive, evidence-backed answers to scientific questions. The report should be a structure multi-paragraph report.
- Think and search until you have sufficient information
- Only provide the final answer when ready
- Cite all claims from search results. You should ground every nontrivial claim in retrieved snippets. 
- Please prefer authoritative sources (peer-reviewed papers, reputable benchmarks/docs) and prioritize recent work for fast-moving areas.
- You should acknowledge uncertainty and conflicts; if evidence is thin or sources disagree, state it and explain what additional evidence would resolve it. 
- It's important to structure with clear markdown headers and a coherent flow. In each section, write 2-5 sentence paragraphs with clear topic sentences and transitions; use lists sparingly only when they improve clarity. Ideally, you should synthesize rather than enumerate content: it's helpful to group findings across papers, explain relationships, and build a coherent narrative that answers the question, supported by citations. 
- Most importantly, DO NOT invent snippets or citations and never fabricate content.
  
Question: 
\end{VerbatimWrap}
\end{tcolorbox}
\caption{\textbf{GPT-5+ Search (baseline).} This is the prompt used for the GPT-5 + Search baseline.}
\label{fig:prompt-gpt-5}
\end{figure}

\paragraph{Naive RAG.}
For the Naive RAG baselines, we retrieve the top 10 search snippets from \texttt{google\_search} using the original question, and then prompt the LM with a simple instruction: 
``Can you try to answer the question given the retrieved documents? Specifically, you should reason step by step, given the evidence retrieved from the web; when there is no evidence present, you should try to answer it based on your knowledge. Please provide a final answer in the format of ``Final Answer: [your answer here]''. 

\paragraph{Fixed pipeline deep research models on short-form questions.} We do not evaluate models that use a specialized inference pipeline on short-form questions because they are unable to follow instructions to only output the answer. For instance, here is an example answer from Ai2 ScholarQA for a query from SimpleQA:
\begin{quote}
Query: Which two scientists (first and last names) are credited with first isolating Azotobacter salinestris from saline soils? Answer in fewer than 10 words.

Answer: Page and Shivprasad were the scientists who first isolated Azotobacter salinestris from saline soils \textless Paper corpusId=``7646032'' paperTitle=``(Robson et al., 2015)'' isShortName\textgreater \textless/Paper\textgreater. While Beijerinck isolated Azotobacter species in 1901, this was a different species and predated the discovery of A. salinestris \textless Paper corpusId=``87342753'' paperTitle="(Shin et al., 2016)" isShortName\textgreater\textless/Paper\textgreater
\end{quote}

\subsection{Score Calculation Details}
\label{app_sec:score_calculation}
\paragraph{Asta-ScholarQA-CS2.} We use the official code\footnote{\url{https://github.com/allenai/asta-bench}} to evaluate our method and the baselines on Asta-ScholarQA-CS2. We compute rubric score, answer precision, citation precision and citation recall on the 100 test set questions using gemini-2.5-flash as the judge as detailed in \cite{bragg2025astabench}. Asta-ScholarQA-CS2 requires evidence text from citations in order to judge citation precision and recall. For proprietary models (OpenAI Deep Research, GPT-5) that only provide URLs for citations, we use JINA API to scrape the URL and use the resulting text as citation evidence. 

\paragraph{HealthBench.} We use an adapted version of the OpenAI simple-evals suite\footnote{\url{https://github.com/openai/simple-evals}} for the evaluation. For each multi-turn example, we concatenate the full conversation into a single input and prepend an instruction directing the model to answer the question based on this doctor\-patient conversation. For efficiency, we randomly sample a subset of 1000 cases for evaluation.

\paragraph{DeepResearchBench.} We use the official code\footnote{\url{https://github.com/Ayanami0730/deep_research_bench}} to evaluate our method and the baselines on DeepResearchBench. We compute the Comprehensiveness, Insight/Depth, Instruction-Following, and Readability of articles answering 50 English and 50 Chinese open-ended deep research questions. We report the Overall as the macro average of the component metrics. We use gemini-2.5-flash as the judge and Jina API to scrape the URL to acquire the evidence snippets when needed, as detailed in \cite{du2025deepresearch}. For outputs from our system, we use the scraped URL content from the corresponding search or browsing tools.

\paragraph{ResearchQA.}  We evaluate our method and baselines using the original ResearchQA evaluation suite\footnote{\url{https://github.com/realliyifei/ResearchQA}}. We compute the averaged rubric scores with GPT-4.1-mini as the judge on the 776 official subset of questions used to evaluate deep research systems, following \cite{yifei2025researchqa}.

\subsection{Details of Pathogenic Gene Variants Evaluation}
\label{app:genetic_disease_eval}

\paragraph{Dataset} The evaluation data for this task was derived from expert-curated information collected for 24 pathogenic gene variants published in the supplementary data of \cite{Cheerie2025}, which was used to develop guidelines for the assessment of variant eligibility for various types of antisense oligonucleotide (ASO) gene therapy. Curations were done by members of the N=1 Collaborative Patient Identification Working Group, which consists of both medical professionals (MD, PhD, and master's level) and faculty with expertise in medical genetics. The selected variants were deemed feasible to assess with publicly available information, and each response was agreed upon by two members. This data reports characteristics of selected variants that are essential to determining therapeutic eligibility, including the variant's pathomechanism, haploinsufficiency status of the gene it affects, inheritance pattern of associated diseases, splicing effects, and findings from prior therapeutic approaches explored. We manually reformatted this data into 47 question-answer examples. Genetic variants are specified in HGVS notation. Questions were preceded by a few sentences of context containing some details on what types of evidence are preferred and the proper answer format.

\paragraph{Setup} To avoid the effects of contamination, we blocked all search results pointing to the paper and its supplementary files when evaluating models with our search tools. This was not possible for closed-source DR systems, though the original paper did not appear in the output citations for any model. We also add the statement: "\textit{You should try to find multiple pieces of evidence to support any claims you make, and acknowledge conflicting/supporting evidence among sources searched}" to the baseline prompt in Figure \ref{fig:prompt-gpt-5} for GPT+5, OpenAI DR, and Gemini 3 Pro to ensure they are calibrated to the preferences of this task

\paragraph{Evaluation Criteria} Given that the task's ultimate goal is to aid medical decision-making, we designed criteria that capture not only the correctness of the final answer, but the usefulness of the generated report to a clinician or researcher:
\begin{itemize}
    \item \textit{Final Answer} indicates whether the expert-annotated fact was mentioned in the response. Each question has 1-3 key facts that should be present in an ideal response, and the per-example correctness is the average of these.
    \item \textit{Evidence Quality} indicates if the type of evidence requested in the query is present within the \textit{cited} statements (e.g. functional assays in patient-derived cells). This score aims to penalize (1) uncited evidence and (2) evidence that is cited, but irrelevant. To demonstrate this, we provide an example response from \modelname that met the Evidence Synthesis criteria (Figure \ref{fig:genetic-diseases-example-response})
    \item \textit{Evidence Support} measures the proportion of cited statements in the response that are consistent with the original source text retrieved. This was calculated by evaluating each cited span with the original content of the cited source.
    \item \textit{Evidence Synthesis} indicates whether or not there was at least one statement describing the relationship between multiple sources, e.g. how papers might build off each other or conflict.
\end{itemize}

Similar to our prior experiments, we defined specific LLM judge instructions for each evaluation criteria per question and used GPT-4.1 to score each response. No additional training was performed for this task.

\begin{figure}
\begin{tcolorbox}[colback=Chocolate!5!white,colframe=Ai2PinkLight,title=LLM judge prompt for \textit{Evidence Support} evaluation (GeneticDiseasesQA)]
\begin{VerbatimWrap}
# Instructions
You are a claim validator. You will be given the text content from a webpage, and 
a list of claims from a research report that cited the webpage.

<<content>>
<<prompt>>
<<claims>>

For each claim given you, you will determine if it is supported by the original source's content given. For source content
with only the title available, judge them as `supporting` if the title indicates that the
paper is likely relevant to the claim being considered.
Return a JSON object with a single key `claims` which is a list of `claim` objects,
one for each phrase given from the research report. Each `claim` object contains the
claim itself (`text`), a boolean `is_supported` which indicates if the claim is fully
supported by the citation source content, and `explanation`, which should be a string briefly
explaining why the response is or is not supported by the citation source content.
If the provided source content is completely empty, entirely filler webpage text, or is an error message or content filter, 
output an empty list for the `claims` key.

# Example

USER:
Content of cited source: "US holidays include Veteran's Day (November 11th) and Halloween (October 31st)."

Claims from generated report:
- "Veteran's Day is a national holiday in November."

ASSISTANT:
```json
"claims": [
    {
      "text": "Veteran's Day is a national holiday in November.",
      "is_supported": true,
      "explanation": "Sources states that Veteran's Day falls on November 11th."
    },
]
```

# Final instruction
Return just the json object in markdown format. Do not include any other text in the response.
\end{VerbatimWrap}
\end{tcolorbox}
\end{figure}

\begin{figure}
\begin{tcolorbox}[colback=Chocolate!5!white,colframe=Ai2PinkLight,title=Example Response of \modelname-8B on GeneticDiseasesQA Part I]
\begin{VerbatimWrap}
USER:
Context: Your task is to determine the pathomechanism of a given pathogenic gene variant. A pathogenic variant can lead to different effects. The variant can lead to a loss of function (LoF) of a protein, a toxic gain of function (GoF), or a dominant-negative (DN) effect. Assessing the variant effect can be done by retrieving publications and reports of the variant [...]
Question: What is the pathomechanism of variant NM_001127222.2(CACNA1A):c.4174G>A?

Summary of the variant and its identity
- The variant NM_001127222.2(CACNA1A):c.4174G>A corresponds to a missense change p.Val1392Met in the CACNA1A alpha 1A subunit and is curated as a pathogenic or likely pathogenic allele in clinical resources, notably ClinVar and the ClinGen allele registry, where it is associated with familial hemiplegic migraine and related neurologic phenotypes <cite id="00907dd4-0">NM_001127222.2(CACNA1A):c.4174G>A (p.Val1392Met) is cataloged as a single-nucleotide variant in ClinVar with the canonical transcript NM_001127222.2</cite> <cite id="00907dd4-4">The ClinGen Allele Registry lists NM_001127222.2(CACNA1A):c.4174G>A (p.Val1392Met) as a registered canonical allele</cite> <cite id="00907dd4-6">MGeND maps NM_001127222.2(CACNA1A):c.4174G>A (p.Val1392Met) to Migraine, familial hemiplegic, 1 (FHM1)</cite>.

Gene-level context: CACNA1A encodes the P/Q-type calcium channel alpha 1A subunit and underlies an autosomal dominant channelopathy spectrum
- CACNA1A encodes the pore-forming $\alpha$1A subunit of the high-voltage–activated P/Q-type calcium channel and is a dominant disease gene for three disorders: familial hemiplegic migraine (FHM), episodic ataxia type 2 (EA2), and spinocerebellar ataxia type 6 (SCA6) <cite id="00907dd4-1">Pathogenic variants in CACNA1A are associated with episodic ataxia type 2, familial hemiplegic migraine, and spinocerebellar ataxia 6</cite> <cite id="8169b531-9">The CACNA1A gene was identified for FHM; it encodes the $\alpha$1A subunit of the P/Q-type calcium channel</cite>. 
- Clinically, affected individuals present on a phenotypic spectrum from migraine with transient hemiplegia/ataxia to persistent cerebellar ataxia and atrophy, consistent with dominant inheritance and variable expressivity <cite id="7a079b1d-9">Mutations in CACNA1A are responsible for at least three autosomal dominant disorders: FHM1, episodic ataxia, and spinocerebellar ataxia</cite> <cite id="7a079b1d-0">CACNA1A-related hemiplegic migraine is associated with neurological variability and requires urgent treatment</cite> <cite id="7a079b1d-7">Clinically, CACNA1A variants are linked to autosomal dominant FHM with cerebellar ataxia</cite>.
- Across conditions, both gain-of-function (GoF) and loss-of-function (LoF)/haploinsufficient phenotypes have been described, with FHM commonly linked to GoF variants affecting channel gating/hyperexcitability and ataxia/SCA6 more often associated with LoF/hypofunction or channel dysfunction leading to neurodegeneration <cite id="8169b531-7">Hemiplegic migraines are associated with particular types of genetic changes in CACNA1A called “gain of function” variants</cite> <cite id="7a079b1d-6">FHM with cerebellar ataxia is only associated with specific CACNA1A mutations that lead to ataxia</cite>. 
- In broader neurodevelopmental disease, CACNA1A variants have also been reported in epileptic encephalopathy and cognitive/autism phenotypes, underscoring that some pathogenic variants may act through haploinsufficiency or dominant-negative effects in certain contexts <cite id="00907dd4-2">ClinVar Miner notes CACNA1A missense variants in neurodevelopmental conditions</cite> <cite id="00907dd4-5">Reports associate CACNA1A haploinsufficiency with cognitive impairment, autism, and epileptic encephalopathy with mild cerebellar symptoms</cite>.
\end{VerbatimWrap}
\end{tcolorbox}
\caption{{Example response of \modelname-8B on GeneticDiseasesQA Part I}}
\label{fig:genetic-diseases-example-response}
\end{figure}

\begin{figure}
\begin{tcolorbox}[colback=Chocolate!5!white,colframe=Ai2PinkLight,title=Example Response of \modelname-8B on GeneticDiseasesQA Part II]
\begin{VerbatimWrap}

Variant-specific evidence and phenotype triangulation for p.Val1392Met (c.4174G>A)
- The p.Val1392Met change in CACNA1A is curated as pathogenic/likely pathogenic for FHM and, in some families, associated with persistent cerebellar ataxia and atrophy, aligning with a dominant gain-of-toxic-function mechanism that perturbs neuronal excitability (migraine) and, in some cases, contributes to neurodegeneration <cite id="00907dd4-0">ClinVar classifies NM_001127222.2(CACNA1A):c.4174G>A (p.Val1392Met)</cite> <cite id="7a079b1d-9">FHM and ataxia phenotypes are both linked to CACNA1A mutations in an autosomal dominant framework</cite> <cite id="7a079b1d-6">FHM with cerebellar ataxia is only associated with certain CACNA1A mutations</cite>. 
- Population and registry compilations include c.4174G>A among pathogenic CACNA1A variants, reinforcing its role in the dominant channelopathy spectrum; however, direct electrophysiological characterization of this specific variant (e.g., activation voltage, Ca2+ influx, inactivation) is not provided in the retrieved summaries and would require targeted functional assays <cite id="00907dd4-9">Variant catalogs list NM_001127222.2(CACNA1A):c.4174G>A (p.Val1392Met) among pathogenic CACNA1A variants</cite> <cite id="8169b531-8">Clinical-genetic characterization of CACNA1A-related disease summarizes dominant phenotypes across FHM, EA2, and SCA6</cite>.

Applying the criteria to determine pathomechanism
- Dominant inheritance and autosomal transmission across FHM, EA2, and SCA6 support a dominant mechanism at the gene–phenotype level for CACNA1A variants <cite id="7a079b1d-9">CACNA1A-related disorders are autosomal dominant</cite>.
- For pathomechanism, CACNA1A variants can act as GoF (hyperexcitability/channel dysfunction causing migraine) or LoF/haploinsufficient (reduced channel function leading to ataxia/degeneration), with some variants producing mixed or variable phenotypes depending on domain impact and genetic background <cite id="8169b531-7">FHM is associated with gain-of-function variants in CACNA1A</cite> <cite id="7a079b1d-6">FHM with cerebellar ataxia is specifically associated with certain CACNA1A mutations that lead to ataxia</cite> <cite id="00907dd4-5">Haploinsufficiency in CACNA1A has been linked to neurodevelopmental phenotypes</cite>. 
- When both GoF and LoF mechanisms are plausible at the gene level, the variant’s phenotype and inheritance refine the likely mechanism. In FHM, transient neurologic deficits and aura point to a dominant gain-of-toxic-function mechanism in sensory-motor cortical circuits; in EA2/SCA6, persistent ataxia and cerebellar atrophy point to reduced channel function or progressive neurodegeneration, consistent with LoF/haploinsufficiency or dominant-negative effects in cerebellar pathways <cite id="8169b531-7">Hemiplegic migraine variants are labeled as gain-of-function</cite> <cite id="7a079b1d-6">Cerebellar ataxia in FHM is linked to specific CACNA1A mutations</cite>.
- Given that p.Val1392Met is cataloged as pathogenic for FHM and associated with cerebellar ataxia/atrophy in some pedigrees, the most parsimonious pathomechanism is a dominant gain-of-toxic-function effect on neuronal excitability that drives migraine, with possible additional contributions to cerebellar dysfunction via haploinsufficiency or dominant-negative effects in cerebellar circuits, depending on genetic background and penetrance <cite id="00907dd4-0">ClinVar catalogs p.Val1392Met as a pathogenic CACNA1A variant</cite> <cite id="7a079b1d-9">Autosomal dominant inheritance and variable phenotypes across FHM, EA2, and SCA6 are established</cite> <cite id="00907dd4-5">Haploinsufficiency in CACNA1A contributes to neurodevelopmental disease</cite>.
\end{VerbatimWrap}
\end{tcolorbox}
\caption{{Example response of \modelname-8B on GeneticDiseasesQA Part II}}
\end{figure}

\begin{figure}
\begin{tcolorbox}[colback=Chocolate!5!white,colframe=Ai2PinkLight,title=Example Response of \modelname-8B on GeneticDiseasesQA Part III]
\begin{VerbatimWrap}
Uncertainties and what functional studies would resolve them
- The retrieved evidence establishes the variant’s identity, clinical classification, and the gene’s dominant channelopathy framework, but does not provide direct patch-clamp or Ca2+ imaging data for p.Val1392Met. Therefore, while a dominant gain-of-toxic-function mechanism is most consistent with FHM presentation and gene-level GoF associations, we cannot exclude a dominant-negative or partial haploinsufficient contribution without direct tests <cite id="8169b531-8">Clinical and genetic characterization supports dominant mechanisms but lacks variant-specific electrophysiology</cite>.
- To conclusively classify the mechanism for c.4174G>A (p.Val1392Met), the following functional evidence would be decisive:
  - Heterologous expression studies in HEK293 or Xenopus oocytes measuring Ca2+ current amplitude, activation/dactivation kinetics, voltage dependence, inactivation, and pharmacology, with rescue by wild-type co-expression to test for dominant-negative effects <cite id="00907dd4-9">Variant catalogs emphasize the need for functional characterization for CACNA1A missense variants</cite>.
  - Neuronal action potential and neurotransmitter release assays in cerebellar Purkinje cells or cortical neurons to quantify excitability changes and test for GoF vs LoF phenotypes <cite id="8169b531-9">CACNA1A encodes the P/Q-type channel critical for neurotransmitter release and neuronal excitability</cite>.
  - Human induced pluripotent stem cell–derived neurons or patient-specific cerebellar organoids to assess developmental Ca2+ signaling and degeneration kinetics in LoF vs GoF contexts <cite id="00907dd4-5">Reports link CACNA1A haploinsufficiency to neurodevelopmental phenotypes, supporting developmental functional assays</cite>.
  - Trans-heterozygosity testing in models (e.g., LoF knockouts) to probe dominant-negative interactions specific to p.Val1392Met <cite id="7a079b1d-6">FHM with cerebellar ataxia is associated with specific CACNA1A mutations suggesting gene-dose interactions</cite>.

Final assessment for NM_001127222.2(CACNA1A):c.4174G>A (p.Val1392Met)
- Pathomechanism: Dominant gain-of-toxic-function in the neuronal excitability pathway is the most consistent explanation for p.Val1392Met, given its classification as a pathogenic FHM variant and the gene’s GoF-linked channelopathy basis for migraine; in some pedigrees with cerebellar ataxia/atrophy, additional haploinsufficient or dominant-negative contributions in cerebellar pathways may contribute, yielding a mixed mechanism across brain regions and genetic backgrounds <cite id="00907dd4-0">ClinVar classifies p.Val1392Met as pathogenic in CACNA1A</cite> <cite id="8169b531-7">FHM is associated with gain-of-function CACNA1A variants</cite> <cite id="7a079b1d-6">Cerebellar ataxia in FHM is associated with specific CACNA1A mutations, suggesting multi-domain effects</cite> <cite id="00907dd4-5">Haploinsufficiency in CACNA1A is linked to neurodevelopmental phenotypes</cite>.
- Practical implication: In clinical contexts where CACNA1A pathogenic variants are known to act dominantly and cause FHM with possible cerebellar involvement, c.4174G>A (p.Val1392Met) should be considered to confer a dominant gain-of-function (toxic) effect on P/Q-type Ca2+ channel–mediated neurotransmission, with possible additive/penetrant haploinsufficient effects in cerebellar pathways in affected families <cite id="7a079b1d-9">Autosomal dominant inheritance and variable phenotypes across migraine and ataxia are established for CACNA1A</cite> <cite id="8169b531-7">Gain-of-function mechanism is characteristic of FHM-associated CACNA1A variants</cite>.
\end{VerbatimWrap}
\end{tcolorbox}
\caption{{Example response of \modelname-8B on GeneticDiseasesQA Part III}}
\end{figure}

\begin{figure}
\begin{tcolorbox}[colback=Chocolate!5!white,colframe=Ai2PinkLight,title=Example Response of \modelname-8B on GeneticDiseasesQA Part IV]
\begin{VerbatimWrap}
Evidence gaps and recommendations
- Direct functional characterization of p.Val1392Met remains to be reported in the retrieved literature; thus, further mechanistic studies (as outlined above) are warranted to exclude or confirm dominant-negative or LoF contributions and to delineate the variant’s phenotype–mechanism continuum within the CACNA1A spectrum of disease <cite id="8169b531-8">Clinical-genetic summaries highlight the need for functional follow-up of CACNA1A missense variants</cite> <cite id="00907dd4-9">Variant catalogs emphasize the presence of pathogenic missense variants across neurological conditions and the need for functional characterization</cite>.

Cited texts:
[00907dd4-0]: Title: NM_001127222.2(CACNA1A):c.4174G>A (p.Val1392Met ... - NCBI
[00907dd4-4]: Title: CA242642 - Allele Registry - ClinGen
[00907dd4-6]: Title: Annotation Detail : MGeND
[00907dd4-1]: Title: Clinical and genetic characterization of CACNA1A‐related disease
[8169b531-9]: Title: CACNA1A gene mutations in familial hemiplegic migraine
[7a079b1d-9]: Title: Two distinct phenotypes, hemiplegic migraine and episodic Ataxia ...
[7a079b1d-0]: Title: Hemiplegic Migraine Variant List - CACNA1A Foundation
[7a079b1d-7]: Title: Next-generation sequencing identified a novel CACNA1A I1379F ...
[8169b531-7]: Title: CACNA1A-Related Disorders - Children's Hospital of Philadelphia
[7a079b1d-6]: Title: [Familial hemiplegic migraine] - PubMed - PMID:15775664
[00907dd4-2]: Title: Submissions for variant NM_001127222.2(CACNA1A):c.4174G>T
[00907dd4-5]: Title: LOC126862864 Gene - GeneCards - GeneCards
[00907dd4-9]: Title: Total 90 pathogenic variants reported for Spinocerebellar ataxia type 6
[8169b531-8]: Title: Clinical and genetic characterization of CACNA1A-related disease
\end{VerbatimWrap}
\end{tcolorbox}
\caption{{Example response of \modelname-8B on GeneticDiseasesQA Part IV}}
\end{figure}
\section{More Results and Analysis}
\subsection{Performance and Cost Breakdown}

\begin{table*}[t!]
\resizebox{\textwidth}{!}{
    \centering
    \begin{tabular}{lcccc|cccc}
    \toprule
       &  \multicolumn{4}{c}{\bf AstaBench-ScholarQA-CS2 (SQAv2) } &    \multicolumn{4}{c}{\bf DeepResearchBench (DRB)} \\
         &  Rubric & Answer & Cite-P & Cite-R & Comp & Depth & Instruction & Readability \\
    \midrule
    \textcolor{gray}{\it Closed Deep Research} & & & & & & & & \\
    \rowcolor{gray!20}\claude~Claude-Sonnet Search & - & - & - & - & 39.0 & 37.7 & 45.8 & 41.5 \\
          \rowcolor{gray!20}\perplexity~Perplexity Sonar & - & - & - & - & 37.4 & 36.1 & 45.7 & 44.7 \\
          \rowcolor{gray!20}\perplexity~Perplexity DR & 91.6 & 92.7 & 47.3 & 37.6 &   40.7 & 39.3 & 46.4 & 44.3 \\
        \rowcolor{gray!20} \gemini~Gemini Deep Research & - & - & - & - &  48.5 & 48.5 & 49.2 & 49.4 \\ 
        \rowcolor{gray!20} \gemini~Gemini3 Pro + Search & 83.1 & 98.3 & 68.5 &  29.4 & 43.4  & 44.9 & 49.8 & 49.0  \\ 
         \rowcolor{gray!20}\oai~GPT-5 + Search & 92.3 & 93.8 & 67.8 & 45.6 & 49.7 & 51.5 & 51.6 & 48.5 \\
         \rowcolor{gray!20}\oai~GPT-5 + Our Search & 74.9 & 93.2 & 42.5 & 33.7 & 26.7  & 21.3 & 41.0 & 29.4 \\
         \rowcolor{gray!20}\oai~OpenAI DR & 91.5 & 95.6 & 77.4 & 43.1 & 46.8 & 45.2 & 49.2 & 47.1 \\
                   \midrule
        \textcolor{gray}{\it Naive RAG} & & & & & & & & \\ 
        Qwen3-8B & 69.2 &  92.3 & - & - & 29.4 & 27.0 & 40.2 & 41.1\\
        QwQ-32B & 77.5 & 90.3 & - & - & 38.1 & 34.8 & 47.0 & 44.6 \\ 
        \midrule
            \textcolor{gray}{\it Open Deep Research} & & & & & & & & \\
        Search-R1-7B & 9.7 & 79.0 & - & - & 5.2 & 2.1 & 18.6	& 16.8 \\
        ASearcher-7B & 13.7 & 94.0 & - & - & 5.1 & 1.7 & 15.2 & 11.8 \\
        WebExplorer-8B & 78.6 & 91.4 & - & - & 33.7 & 28.5 & 45.7 & 42.2 \\
        WebThinker-32B-DPO  & 36.7 & 94.9 & - & - & 19.7 & 12.3 & 36.8 & 26.3 \\
        Tongyi DeepResearch-30B-A3B  & 89.5 &  96.4  &  - &  - &  39.1 & 34.3 & 46.8 & \bf 45.4 \\
        \midrule
        \textcolor{gray}{\it Fixed Pipeline Deep Research} & & & & & \\ 
        WebThinker QwQ-32B (report) & 86.4 & 94.3 & - & - & 36.2 & 32.6 & 43.2 & 42.9 \\
        WebThinker-32B-DPO (report) &  91.2 & 95.5 & - & - & 39.4 & 35.4 & 46.0 & 43.5 \\ 
        \rowcolor{gray!20} \aitwo\ Ai2 ScholarQA - Claude Sonnet & 88.1 & 89.1 & 92.4 & 81.2 &35.1 &32.0 &40.5 &38.9 \\
        \midrule
         \textcolor{gray}{\it Open Deep Research (Ours)} & & & & & \\ 
        Qwen3-8B + Our Search & 42.8 & 92.1 & 53.7 & 40.3 & 14.3 & 8.7 & 29.5 & 24.4 \\
        \rowcolor{pink!20}  \aitwo~\modelname-8B (SFT) & 81.4 & 91.0 & 65.3 & 51.6 & 36.3 & 35.3 & 45.5 & 39.5 \\
        \rowcolor{pink!20} \aitwo~\modelname-8B (RL) & \bf 92.4 & \bf 98.8 & \bf 90.5 & \bf 71.6 & \bf 44.2 & \bf 44.5 &  \bf 49.4 & 42.4 \\ 
    \bottomrule
    \end{tabular}%
    }
    \caption{{\bf Performance breakdown for Asta-ScholarQA-CS2 and DeepResearchBench.} Open deep research models and naive RAG baselines do not provide citations, indicated as ``-'' in citation columns. Rows with a gray background indicate models that use closed models as backbone LMs. {\bf Bold} indicates the best results among the baselines that do not use propriety models. }    \label{tab:performance_breakdown}
\end{table*}

\begin{table*}[ht!]
\small
\centering
    \begin{tabular}{l|rrrl}
        \toprule
         & Answer Length & Citations & Tool Calls & Cost / Query* \\
        \midrule
        \rowcolor{gray!20} GPT-5+ Search & 2358.7 &  28.1 & - & 0.29  \\
        \rowcolor{gray!20}OpenAI Deep Research & 6445.1 & 79.6 & - & 1.8 \\
        \rowcolor{gray!20} Gemini 3 Pro + Search & 1310.9 & 8.6 & 8.5 &  0.13 \\
        \rowcolor{gray!20}Ai2 ScholarQA - Claude Sonnet & 2090.5 & 61.2 & 1.0 & 1.3 \\
        WebExplorer-8B & 1250.4 & - & 9.1 & 0.019 \\
        WebThinker-32B & 92.2 &  - & 6.9 & 0.0037 \\
        WebThinker-32B (report) & 4416.7 &  - & 8.2 & 0.015 \\
        Tongyi Deep Research-30B-A3B & 2138.9 &  - & 23.0 & 0.032 \\
        \rowcolor{pink!20}\modelname-8B (RL) & 1889.2 &  35.8 & 4.3 & 0.0019 \\
        \bottomrule
    \end{tabular}
    \caption{{\bf Comparison of model usage statistics on SQAv2.} We report answer lengths, tool usage, and citation counts across systems. ``-'' denotes this information was either not available or it was not applicable. The cost per query is estimated based on model inference on ScholarQA-CS2, following \cite{bragg2025astabench}. More details of cost estimations are available in Appendix \ref{app_sec:cost_estimation}. 
}
\label{tab:usage-stats}
\end{table*}

Table~\ref{tab:performance_breakdown} provides detailed performance breakdown of models across four main long-form benchmarks.

Comparing \modelname (SFT) and \modelname (RL), we observe consistent gains from RLER across multiple aspects, including rubric coverage (+11.0 points), answer precision (+7.8), comprehensiveness (+7.9), and depth of response (+9.2). RLER also yields large improvements in citation precision and recall on SQAv2 (+25.2 and +20.0 points, respectively). These results highlight that  RLER can effectively improve deep research responses along both content and attribution dimensions.

Table~\ref{tab:usage-stats} shows the cost of the inference of competitive systems. 

\subsection{Full RL Training Curves}
\label{app:full_rl_training_curves}

\begin{figure}[t!]
    \centering
    \includegraphics[width=\linewidth]{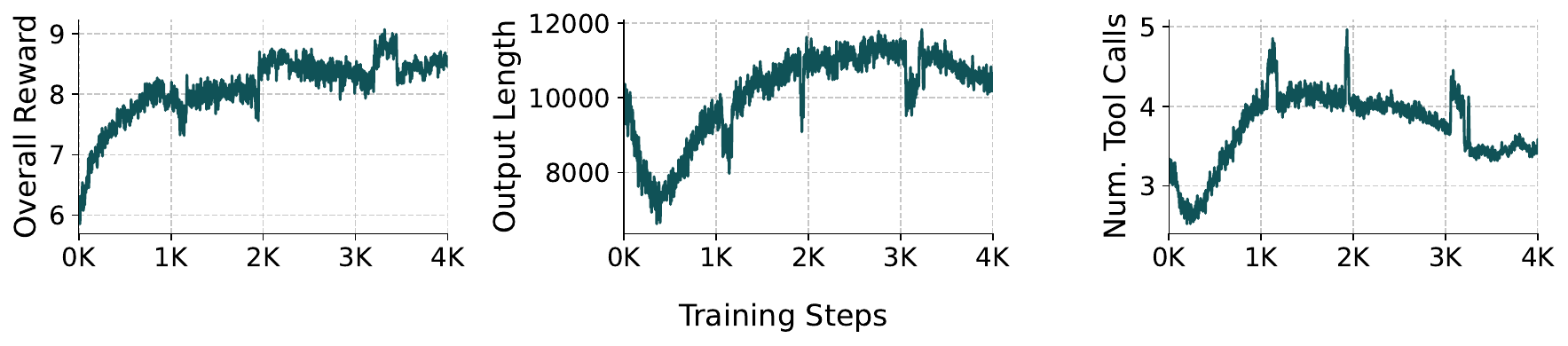}
    \caption{{\bf Overall reward, output length (in tokens, including tool outputs), and average number of tool calls during RL training.} Note that during the run, we periodically ran out of serper credits, causing a rise tool errors until we refilled the account, causing a drop in reward and output length, and a rise in tool calls (as the model retried the failed calls). This happened a few times during the run. We also found that reward jumped sharply around step 2000 after a run restart.} 
    \label{fig:rl_rag_full_run}
\end{figure}

We show the reward, number of tool calls, and output sequence length through training in Figure~\ref{fig:rl_rag_full_run}. We observe that there appear to be three phases of training: in the first phase, sequence lengths and the average number of tool calls drop. They then slowly increase until starting to decline again after thousands of training steps.
Similar to prior work, we hypothesize the initial phase may be due to the model initially unlearning unsuccessful behaviors picked up during SFT training before stabilizing and exploring new strategies. A similar drop-and-rise behavior when combining RL training with SFT cold-start data has been observed in other domains, such as mathematical reasoning~\citep{chen2025twostagetrainingcooperativesft}, and we leave further investigation of this phenomenon to future work. The final phases may be due to the model refining its answers to stay strictly within the output length and max tool call restrictions placed on it during training (16384 output tokens and 5 total tool calls, respectively).

Surprisingly, we also find that our training is somewhat {\bf robust to tool errors}, as we accidentally ran out of Serper credits during training. While we eventually refilled, the model did train for some number of steps wherein Serper would continuously return errors, as seen by the drop in reward and sequence length. However, despite this, our overall model performance continued to improve, with step 1900 being significantly better than step 1000 (right before the Serper credit issue). We faced credit or server issues throughout training, as seen in various dips in reward and output length, but our model generally continued to improve across most downstream evaluations. This suggests our training is somewhat robust to server-side tool errors, and exploring the degree of this robustness is an interesting avenue for future investigation.

\subsection{On-Policy SFT and RL Results}
\label{app:onpolicy_sft_rl_results}

We additionally explore augmenting the SFT data with an extra “on-policy” SFT stage. Specifically, we run our trained model on randomly sampled prompts, apply rejection sampling to discard trajectories that do not achieve high scores on search-based rubric verification and citation verification (details in Appendix~\ref{sec:onpolicy_sft}), and then use the remaining trajectories for further SFT. 
We show the results in Figure~\ref{fig:rl_trajectory} (left), comparing using the on-policy-trained model as a starting point for RL relative to our original SFT set, an undertrained model, or using no SFT at all.
While the on-policy SFT slightly boosts SFT model performance, we find it ultimately weakens performance later on during RL training, underperforming using our regular SFT mixture on Healthbench and SQAv2.

\subsection{Ablation on the Citation Reward}
\label{app:citation_reward_ablation}

\begin{table}[t!]
    \centering
    \small
    \begin{tabular}{lccccc}
    \toprule
          &  HealthBench & ResearchQA & SQAv2 & DRB & Average \\
    \midrule
        w/ citation reward  & 42.7 & 71.0 & 86.4 & \bf 42.1 & 60.6 \\
        w/o citation reward & \bf 44.7 & \bf 71.9 & \bf 86.7 & 41.4 & \bf 61.2 \\
    \bottomrule
    \end{tabular}
    \caption{{\bf Ablation on the citation reward.} Both runs branch from a shared checkpoint at step 650 (when RL starts to show clear gains on long-form benchmarks) and continue through step 1300 with identical hyperparameters; the only difference is whether the citation reward is enabled. Disabling the citation reward in the later phase of training does not hurt overall performance and slightly improves the average score, suggesting that the rubric component of RLER, rather than the citation auxiliary signal, drives the gains.}
    \label{tab:citation_reward_ablation}
\end{table}

To isolate the contribution of the evolving rubric reward from the auxiliary citation reward used during RL training, we run an ablation that branches from a shared checkpoint at step 650 (the point at which RL begins to show clear gains on the long-form benchmarks) into two runs with identical hyperparameters through step 1300: one with the citation reward enabled (w/ cite) and one without (w/o cite). As shown in Table~\ref{tab:citation_reward_ablation}, the two runs achieve comparable final performance, with w/o cite slightly ahead on average ($61.2$ vs.\ $60.6$). This indicates that the citation reward is not the source of RLER's gains in this regime, and that the rubric-based reward is responsible for the bulk of the improvement. We additionally observe that the format reward saturates at $1.0$ early in training and contributes little signal beyond the warmup phase. Combined with the RLER on/off ablation in Figure~\ref{fig:rler_ablation}, these results suggest that RLER, rather than the auxiliary rewards, is the main source of our gains, although a full factorial sweep over auxiliary rewards is beyond our compute budget.

\subsection{Cost Estimation}
\label{app_sec:cost_estimation}
In this section, we detail how we estimate the cost of deep research models. Detailed cost comparison can be found in Table~\ref{tab:usage-stats}.

For proprietary models, we use the actual billed costs reported in their API consoles. Although Gemini 3 + Search waives the first 1,500 searches per day,\footnote{\url{https://ai.google.dev/gemini-api/docs/pricing?hl=en}} we still include search costs in our estimates for a fair comparison with other systems that charge for search.

For open models (including our own), we compute the number of input and output tokens and use OpenRouter's published pricing to estimate inference costs. Specifically, we use the published pricing for Tongyi Deep Research\footnote{Tongyi Deep Research costs USD 0.09 per input token and USD 0.4 per output token as of Nov 23 2025.}; we use the published pricing for Qwen3-8B for \modelname, WebExplorer-8B, ASearcher\footnote{Qwen3-8B costs USD 0.2 per input and output token as of Nov 23 2025.}; We we use the published pricing for QwQ-32B for WebThinker models\footnote{QwQ-32B costs USD 0.4 per input and output token as of Nov 23 2025.}. We treat system prompts and questions as input tokens, and all remaining tokens as output tokens in our calculations for open models. 
We also add tool-call costs based on the pricing of each tool's API provider and the average number of tool calls performed by each model across our long-form evaluations.

\subsection{Effect of the tool-call budget at inference time.}
\begin{figure}[t!]
    \centering
    \includegraphics[width=0.95\linewidth]{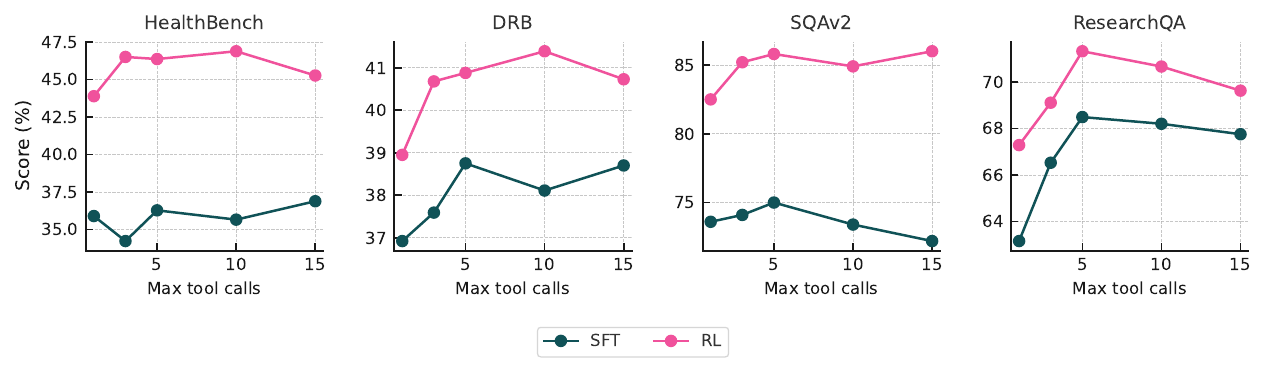}
    \caption{{\bf Ablations on maximum tool calls.} We evaluate \modelname (SFT) and \modelname (RL; 1900 steps) on four long-form datasets while varying the maximum number of allowed tool calls from 1 to 15, and report how performance changes under these caps.
    }
    \label{fig:max_tool_calls}
\end{figure}

We studied how the inference-time tool-call budget affects performance by varying the maximum number of allowed tool calls to \{1,3,5,10,15\}.
For both SFT and RL models, performance typically saturates around a budget of five tool calls, although RL occasionally improves with an additional budget of up to ten tool calls; see Figure~\ref{fig:max_tool_calls}. This matches tool call behavior seen during RL training (Appendix~\ref{app:full_rl_training_curves}), in which the model uses 3-4 tool calls on average per sample.

\subsection{Evaluation Variances}

The inference and evaluation of deep research models often exhibit significant variance. When running models on the same questions or evaluating them on the same benchmarks, the results can vary substantially. Typically, there are three key factors that contribute to this variance and we will discuss them in the following sections.

\subsubsection{Variances Introduced by Tools}

Invoking a tool with identical inputs at different times can yield inconsistent outputs, leading the model to produce divergent subsequent contexts. In this section, we focus on the output variance introduced by the search engine.

We sampled 100 function-calling queries and reissued these queries to the Google search engine, with approximately one week between the two invocations. We then computed the differences in the top-1, top-3, top-5, and top-10 retrieved snippets for each pair of calls. For comparison, we also measured the differences between two calls issued nearly at the same time. We use the following metrics to evaluate the variance of search engines' returns.

\begin{itemize}
    \item \textbf{set\_same:} The number of shared items within the top-$k$ results, regardless of order (i.e., the size of the intersection).
    \item \textbf{pos\_match:} The number of positions in the top-$k$ where both lists contain the same item at the same rank.
\end{itemize}

\begin{table*}[t]
    \centering
    \resizebox{\textwidth}{!}{%
    \begin{tabular}{cc}
    \begin{minipage}{0.48\textwidth}
        \centering
        \begin{tabular}{lcccc}
        \toprule
            & \multicolumn{4}{c}{\textbf{Top-K Snippets}} \\
            \cmidrule(lr){2-5}
            \textbf{Metric} & \textbf{1} & \textbf{3} & \textbf{5} & \textbf{10} \\
        \midrule
            \textbf{set\_same} & 0.77 & 2.04 & 3.34 & 6.01 \\
            \textbf{pos\_match} & 0.77 & 1.71 & 2.22 & 2.93 \\
        \bottomrule
        \end{tabular}
        \subcaption{One-week apart}
        \label{tab:variance_week}
    \end{minipage}
    &
    \begin{minipage}{0.48\textwidth}
        \centering
        \begin{tabular}{lcccc}
        \toprule
            & \multicolumn{4}{c}{\textbf{Top-K Snippets}} \\
            \cmidrule(lr){2-5}
            \textbf{Metric} & \textbf{1} & \textbf{3} & \textbf{5} & \textbf{10} \\
        \midrule
            \textbf{set\_same} & 0.88 & 2.53 & 4.20 & 7.67 \\
            \textbf{pos\_match} & 0.88 & 2.35 & 3.48 & 5.64 \\
        \bottomrule
        \end{tabular}
        \subcaption{Within a short interval}
        \label{tab:variance_control}
    \end{minipage}
    \end{tabular}
    }
    \caption{\textbf{Search engine output variance across repeated queries.} The left table shows results when two queries were issued one week apart, while the right table shows results when the two queries were issued within a short interval.}
    \label{tab:search_variance}
\end{table*}

As shown in Table \ref{tab:search_variance}, the search engine’s returns are unstable. Even when calling the same query within a short interval, it still produces noticeably different results. The average overlap in the top-10 snippets is only about 7.67, with exact rank matches dropping to 5.64.
When the same queries are reissued one week apart, the search engine's returns diverge more significantly. We show examples of inconsistencies in Figure~\ref{fig:search_inconcsistencies}. 

\begin{figure}
\begin{tcolorbox}[colback=Chocolate!5!white,colframe=Ai2PinkLight,title=Showcase of Search Engine's Returns]
\begin{VerbatimWrap}
# Search Results I (show top-5 snippets)
Position: 1
Link: https://en.wikipedia.org/wiki/E._Howard_Hunt
Snippet: Everette Howard Hunt Jr. (October 9, 1918 – January 23, 2007) was an American intelligence officer and author. From 1949 to 1970, Hunt served as an officer ...
Position: 2
Link: https://www.amazon.com/stores/author/B0034QAV74
Snippet: Top E. Howard Hunt titles · American Spy: My Secret History in the CIA, Watergate and Beyond. American Spy: My Secret History in the CIA, Watergate and Beyond.
Position: 3
Link: https://www.fantasticfiction.com/h/e-howard-hunt/
Snippet: Everette Howard Hunt, Jr. was an American author and spy. He worked for the Central Intelligence Agency (CIA) and later the White House under President ...
Position: 4
Link: https://www.goodreads.com/author/list/118536.E_Howard_Hunt
Snippet: E. Howard Hunt has 85 books on Goodreads with 2730 ratings. E. Howard Hunt's most popular book is House Dick.
Position: 5
Snippet: E. Howard Hunt, a spy's spy. Hunt carried on writing spy novels long after the Watergate scandal but the Peter Ward books are among his most popular series ...

# Search Results II
Position: 1
Link: https://en.wikipedia.org/wiki/E._Howard_Hunt
Snippet: Everette Howard Hunt Jr. (October 9, 1918 – January 23, 2007) was an American intelligence officer and author. From 1949 to 1970, Hunt served as an officer ...
Position: 2
Link: https://www.goodreads.com/author/list/118536.E_Howard_Hunt
Snippet: E. Howard Hunt has 85 books on Goodreads with 2730 ratings. E. Howard Hunt's most popular book is House Dick.
Position: 3
Link: https://www.amazon.com/E-Howard-Hunt/e/B0034QAV74/ref=dp_byline_cont_ebooks_1
Snippet: Follow E. Howard Hunt and explore their bibliography from Amazon's E. Howard Hunt Author Page ... Howard Hunt. Most popular. American Spy: My Secret History ...
Position: 4
Link: https://www.fantasticfiction.com/h/e-howard-hunt/
Snippet: Everette Howard Hunt, Jr. was an American author and spy. He worked for the Central Intelligence Agency (CIA) and later the White House under President ...
Position: 5
Link: https://spyscape.com/article/cia-spy-howard-hunt-confessions-of-a-watergate-plumber
Snippet: E. Howard Hunt, a spy's spy. Hunt carried on writing spy novels long after the Watergate scandal but the Peter Ward books are among his most popular series ...
\end{VerbatimWrap}
\end{tcolorbox}
\caption{\bf Examples of Inconsistencies of Search Results.}
\label{fig:search_inconcsistencies}
\end{figure}

\subsubsection{Variances Introduced by Inference}
Generating particularly long trajectories can also introduce variance. Small differences early in the process can lead to substantially divergent final responses. To observe the impact of this variability, we re-ran one short-form benchmark and two long-form benchmarks, comparing the outputs from two separate generations using GPT-4.1 with our auto-search pipeline.

As shown in Table \ref{tab:inference_variance}, in 2Wiki, GPT-4.1 produces final answers that differ by 29.3\% in the 300 cases and obtains high variances in long-form tasks like Healthbench and ResearchQA as well.
\begin{table*}[!th]
\centering
\begin{tabular}{cccc}
\toprule
\textbf{Task} & 2Wiki & Healthbench & ResearchQA\\
\midrule
\textbf{Numbers} & 300 & 900 & 776 \\
\textbf{Diff} & 29.3 & 17.1 & 9.81 \\
\bottomrule
\end{tabular}
\caption{\textbf{Inference Variance.} The Diff in 2Wiki refers to the difference in two final answers of the two trajectories under the same cases, while the Diff in Healthbench and ResearchQA represents the absolute difference in LLM judged scores.}
\label{tab:inference_variance}
\end{table*}

\subsubsection{Variances Introduced by Judge Models}
When evaluating the same responses in different times, even if using the same model as a judge, inconsistent judgments may occur.

We use GPT-4.1 to evaluate the same trajectories twice and the results are shown in Table~\ref{tab:judge_variance}. The judgments show relatively high consistency and reliability on both short-form and long-form tasks.

\begin{table*}[!ht]
\centering
\begin{tabular}{cccc}
\toprule
 & {2Wiki}
 & {Healthbench}
 & {ResearchQA} \\
\midrule
\textbf{1} & 67.67 & 37.67 & 66.18\\
\textbf{2} & 66.33 & 37.51 & 66.43\\
\bottomrule
\end{tabular}
\caption{\textbf{Judgement Variance.}}
\label{tab:judge_variance}
\end{table*}

\subsubsection{Robustness to the Browser Tool Used at Inference}
\label{app:browser_tool_ablation}

\modelname-8B is trained with a local Crawl4AI-based \texttt{web\_browse} tool to reduce training cost, but evaluated with the Jina API to remain consistent with prior open-source baselines (e.g., Tongyi DR, ASearcher). To verify that this train/inference mismatch does not significantly affect downstream performance, we re-evaluate \modelname-8B with Crawl4AI as the inference-time browser. Table~\ref{tab:browser_ablation} shows that the choice of browser at inference has minimal impact ($\le 1$ point on every benchmark, $-0.5$ on average), suggesting that one can train with cheaper browser alternatives and still benefit from other browser providers at test time without re-training.

\begin{table*}[!ht]
\centering
\begin{tabular}{lccccc}
\toprule
Browser & HealthBench & ResearchQA & SQAv2 & DRB & Average \\
\midrule
Crawl4AI (training-time tool) & \bf 54.2 & \bf 76.4 & \bf 87.8 & \bf 45.4 & \bf 66.0 \\
Jina (main-evaluation tool)   & 54.0 & 75.6 & 87.2 & 45.3 & 65.5 \\
\bottomrule
\end{tabular}
\caption{\textbf{Browser tool ablation at inference.} Switching the inference-time \texttt{web\_browse} tool from Crawl4AI (used during training) to Jina (used in our main evaluations) yields only a minor average drop ($-0.5$), demonstrating that the agent generalizes to alternative browsers without re-training.}
\label{tab:browser_ablation}
\end{table*}

\subsection{Mismatch between RL Training and Downstream Evaluation}
\label{sec:mismatch_rl_training}
\begin{figure}[t!]
    \centering
    \includegraphics[width=\linewidth]{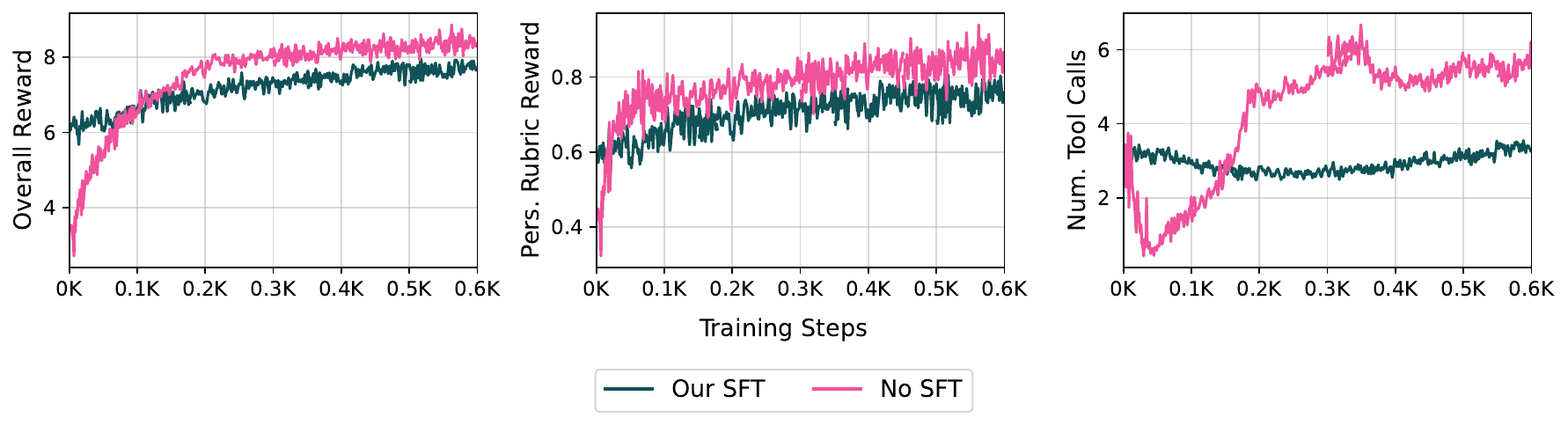}
    \caption{\textbf{Metrics during RL training when starting from two different models (with and without SFT).} ``Pers. Rubric reward'' refers to the reward over the search-based rubrics only (not including the rubrics generated as part of the RLER process). Starting from a model without cold start data (``no SFT'') achieves higher train reward, but underperforms on downstream evaluations.} 
    \label{fig:rl_sft_no_sft}
\end{figure}

During development of \modelname, we found that our RL training setup had some mismatch with our downstream performance: models that achieved the highest training reward did not necessarily achieve the highest downstream evaluation performance. 
To highlight this, compare the training reward of the ``No SFT'' and ``Our SFT'' models in Figure~\ref{fig:rl_sft_no_sft} against their performance in Figure~\ref{fig:rl_trajectory}. While starting directly from Qwen 3 (``No SFT'') achieves highest train reward,it has dramatically lower downstream evaluation results (see Figure~\ref{fig:rl_trajectory}). It also displays significantly different behavior, using significantly more tool calls than the cold-started (``Our SFT'') model.

This may be due to a few factors: first, our evaluations use rubrics generated in different manners to our own training rubrics (e.g., Healthbench uses expert-annotated rubrics), potentially leading to cases where test-time rubrics evaluate features not commonly tested in our training rubrics.
Second, there may be reward hacking behavior during RL training, due to our use of an LM judge different to the judges used in downstream evaluation. Our in-loop judge model uses GPT-4.1-mini, while downstream evaluations use varied different models (e.g., SQAv2 uses Gemini Flash 2.5, DRB uses a mix of Gemini Pro 2.5 and Gemini Flash 2.5, and Healthbench uses GPT-4.1, all with varying prompts and evaluation harnesses). This may lead to our RL training optimizing for attributes preferred by GPT-4.1-mini, but not by downstream evaluation judges.
We finally conjecture that another contributing factor may be the difference in model priors: different models may exploit rewards in different ways. For example, rollouts from a weaker model may contain fewer high-quality answers; when all answers are poor, the judge model may end up selecting based solely on spurious features rather than making meaningful quality comparisons. In contrast, when starting from the same initial model, we usually observe that reward improvements correlate well with downstream scores.
We defer a deeper investigation of this mismatch phenomenon to future work, as addressing it would help improve the effectiveness of rubrics for RL training.

\section{Analysis on Searched Tools and Domain Distributions}

\begin{figure}[t!]
    \centering
    \includegraphics[width=0.95\linewidth]{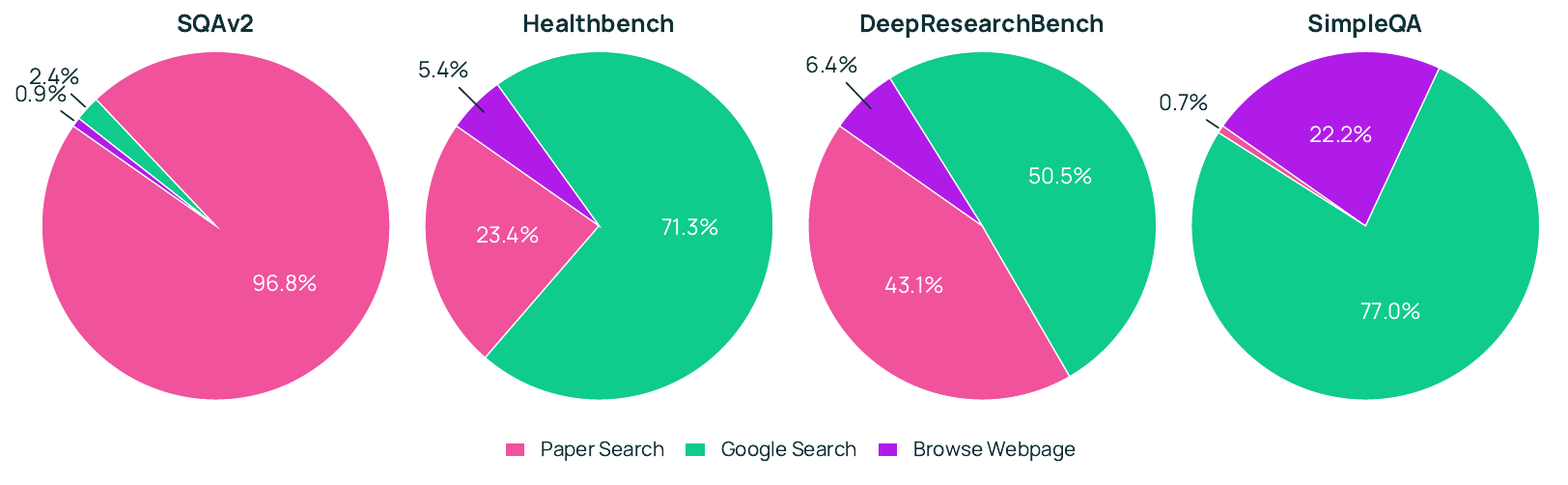}
    \caption{{\bf Distribution of tool calls for SQAv2 (science), HealthBench (healthcare), DeepResearchBench (general domain) and SimpleQA (factoid, short-form QA).} \modelname can adaptively choose effective tools for different tasks, relying more on \texttt{paper\_search} for scientific questions (SQAv2), and more on \texttt{google\_search} for general-domain questions (SimpleQA).}
    \label{fig:tool_distributions}
\end{figure}

We analyze the tools used by an intermediate RL checkpoint of \modelname (step 1900), and find that \textbf{its tool usage adapts to each task’s information needs.}
Figure~\ref{fig:tool_distributions} shows that \texttt{paper\_search} (our scientific-paper search) dominates on SQAv2, consistent with its focus on literature understanding. 

\begin{figure}[t!]
    \centering
    \includegraphics[width=\linewidth]{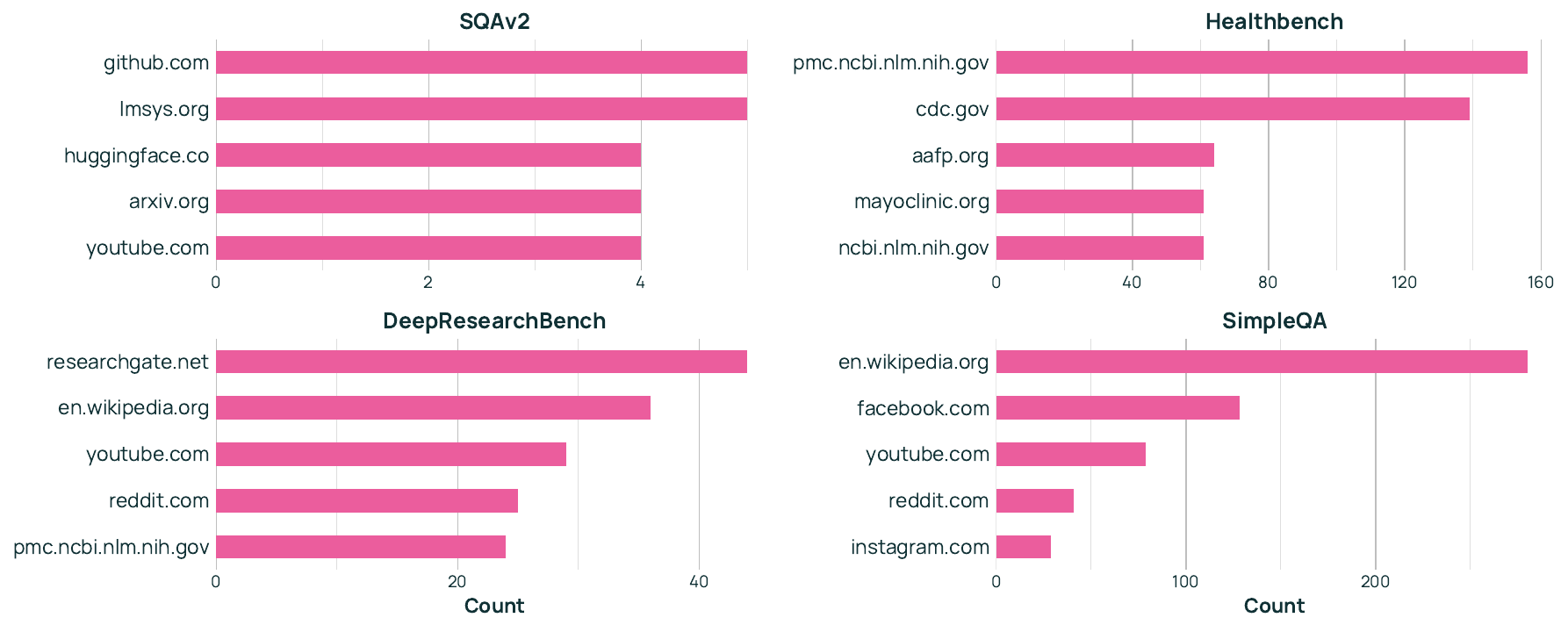}
    \caption{{\bf Distribution of domains among web search results for SQAv2 (science), HealthBench (healthcare), DeepResearchBench (general domain) and SimpleQA (factoid, short-form QA).} We show top domains returned by the \texttt{google\_search} tool. Calculations are based on 100 samples from each task. These top domains match the evaluation domain; e.g., when evaluating on Healthbench, \modelname searches more for medical domain websites.
    }
    \label{fig:domain_distributions}
\end{figure}

Figure~\ref{fig:domain_distributions} further confirms task-specific retrieval behavior. HealthBench emphasizes authoritative biomedical and public-health sites (e.g., \texttt{cdc.gov}, \texttt{pmc.ncbi.nlm.nih.gov}, \texttt{ncbi.nlm.nih.gov}, \texttt{mayoclinic.org}). DeepResearchBench mixes technical and policy sources (e.g., \texttt{researchgate.net}, \texttt{oecd.org}, \texttt{github.com}), consistent with deeper, exploratory research tasks. SimpleQA is dominated by general reference and social/information platforms (e.g., \texttt{en.wikipedia.org}, \texttt{facebook.com}, \texttt{youtube.com}). Overall, tool usage and surfaced domains align with each dataset’s information demands: literature-centric tasks favor scientific search and scholarly venues, whereas open-domain tasks lean on general web search and broad reference sites.

\subsection{Qualitative Examples}
We present one trajectory of our \modelname on the long-form task ResearchQA in Figures~\ref{fig:example_researchqa_1}--\ref{fig:example_researchqa_4}. The response is truncated for brevity; we will release full model response samples after review.

\begin{figure}
\begin{tcolorbox}[colback=Chocolate!5!white,colframe=Ai2PinkLight,title=Example Trajectory for OpenScholar SFT Data Part I]
% ...

\textbf{User}\\
\begin{VerbatimWrap}
Please provide a detailed overview of theories and models related to Artificial Intelligence (AI) in education. Include information on the foundational theories that underpin AI's role in education, such as constructivism, behaviorism, and connectivism, and how these theories inform AI-based learning tools. Additionally, explain prominent AI models and frameworks, such as Intelligent Tutoring Systems (ITS), Adaptive Learning Systems, Learning Analytics, and Natural Language Processing in education. Highlight their key features, applications, and contributions to personalized learning, teacher efficiency, and student engagement. Please reference recent scholarly works to support your response.

\end{VerbatimWrap}
\textbf{Assistant}\\
\begin{VerbatimWrap}
<think>
Decompose the question: We need to (a) outline foundational learning theories (behaviorism, constructivism, connectivism) and explicitly connect them to AI-in-education designs, and (b) explain prominent AI models/frameworks (ITS, adaptive learning, learning analytics, NLP in education), including features, applications, and contributions (personalization, teacher efficiency, engagement), all supported by recent scholarly sources.
Assumptions: There are review/survey papers in AI in Education (AIED), Learning Analytics (LAK), and Educational Data Mining (EDM) from 2020–2025 that synthesize how pedagogical theories inform AI tools, and evaluations of ITS, adaptive systems, analytics, and NLP tools. We should also pull specific examples and quantified impacts where possible (e.g., learning gains from ITS, prediction performance in LA, efficacy of automated feedback).
Search plan:
1) Start with surveys/reviews linking AI in education to learning theories (2020–2025), ideally in AIED, BJET, Computers \& Education, Review of Educational Research, LAK, EDM venues.
2) Retrieve authoritative sources on ITS and adaptive learning (classic + recent advances like knowledge tracing, deep KT, mastery learning, LLM tutoring).
3) Gather learning analytics frameworks and applications (early warning systems, dashboards, ethics).
4) Collect NLP-in-education applications (automated scoring, feedback, dialogue tutoring, LLMs) and evidence on teacher efficiency and student engagement.
First query will target a broad review tying theories and AI models together.
</think>
<call_tool name="snippet_search" year="2020-2025" limit="10" fieldsOfStudy="Education,Computer Science">AI in education review behaviorism constructivism connectivism intelligent tutoring systems adaptive learning learning analytics NLP</call_tool>
\end{VerbatimWrap}
\end{tcolorbox}
\caption{\textbf{Example trajectory for OpenScholar SFT data Part I.}}
\label{fig:example_openscholar_1}
\end{figure}

\begin{figure}
\begin{tcolorbox}[colback=Chocolate!5!white,colframe=Ai2PinkLight,title=Example Trajectory for OpenScholar SFT Data Part II]
\textbf{Functional Calling}\\
\begin{VerbatimWrap}
<tool_output>
...
<snippet id="S_4yWB0UY">Title: The AI Revolution in Education: Will AI Replace or Assist Teachers in Higher Education?
AI in education has evolved from computer-assisted instruction to intelligent tutoring systems (ITS) and adaptive learning technologies, with key themes including personalized feedback, natural language processing, and innovative tools like chatbots and virtual reality, enhancing student engagement and facilitating knowledge co-creation between 2020 and 2025.</snippet>
<snippet id="S_KAymhlA">Title: AI Language Models as Educational Allies: Enhancing Instructional Support in Higher Education
AI in education enhances instructional techniques, assessment methods, and personalized learning through intelligent tutoring systems and adaptive learning, while also necessitating further knowledge on technology to avoid misinformation, indicating a dynamic area of research from 2020 to 2025.</snippet>
...
</tool_output>

\end{VerbatimWrap}
\textbf{Assistant}\\
\begin{VerbatimWrap}
<think>
Reflection on first query results:
- The initial search returned broad reviews touching on AI branches (ML, DL, NLP) and applications (ITS, adaptive learning, automated assessment) with some claims about personalization and teacher support (S_c8C9s9m, S_NTanrK5, S_N8QWms7). These are relevant but high-level and light on explicit connections to learning theories and quantitative evidence.
- S_tuhIPDv explicitly mentions ITS and NLP integrating behaviorism, constructivism, and connectivism, which is directly pertinent, though I need stronger, peer-reviewed sources that articulate these theory-to-tool mappings.
- For models/frameworks, I also need authoritative, recent evidence on ITS effectiveness, adaptive learning via knowledge tracing, learning analytics frameworks (predictive models, dashboards), and NLP applications (AWE, ASR, chatbots, LLMs) with impacts on personalization, teacher efficiency, and engagement.

Gaps:
- Explicit theoretical grounding linking behaviorism, constructivism, and connectivism to AI tools with scholarly backing.
- Quantitative or meta-analytic evidence for ITS/adaptive systems.
- Up-to-date learning analytics frameworks and outcomes.
- NLP in education evidence, including automated feedback and LLM-based tutoring.

Next query:
Search for papers that explicitly connect behaviorism, constructivism, and connectivism to AI/ITS/adaptive learning frameworks in education, 2020–2025, in education and CS venues.
</think>
<call_tool name="snippet_search" year="2020-2025" limit="10" fieldsOfStudy="Education,Computer Science">behaviorism constructivism connectivism artificial intelligence in education theoretical framework</call_tool>

\end{VerbatimWrap}
\textbf{Functional Calling}\\
\begin{VerbatimWrap}
...
\end{VerbatimWrap}
\end{tcolorbox}
\caption{\textbf{Example Trajectory for OpenScholar SFT Data Part II.}}
\label{fig:example_openscholar_2}
\end{figure}

\begin{figure}
\begin{tcolorbox}[colback=Chocolate!5!white,colframe=Ai2PinkLight,title=Example Trajectory for OpenScholar SFT Data Part III]
\textbf{Assistant}\\
\begin{VerbatimWrap}
<think>
We have initial broad reviews but need stronger, specific evidence tying learning theories to AI designs and quantifying effects for ITS, adaptive learning, learning analytics, and NLP in education. Key gaps: (1) explicit mapping of behaviorism/constructivism/connectivism to AI paradigms with scholarly backing; (2) quantitative evidence for ITS/adaptive learning (e.g., learning gains, knowledge tracing); (3) learning analytics frameworks and teacher-facing outcomes; and (4) NLP applications (AWE/ASR/chatbots/LLM tutors) with impacts on personalization, efficiency, and engagement.

Next, I will search for meta-analyses or systematic reviews on ITS effectiveness and adaptive learning from 2020–2025, then for learning analytics frameworks and NLP-in-education evaluations.
</think>
<call_tool name="snippet_search" year="2020-2025" limit="10" fieldsOfStudy="Education,Computer Science">intelligent tutoring systems meta-analysis effectiveness learning gains 2020..2025 educational data mining knowledge tracing adaptive learning</call_tool>

\end{VerbatimWrap}
\textbf{Functional Calling}\\
\begin{VerbatimWrap}
...

\end{VerbatimWrap}
\textbf{Assistant}\\
\begin{VerbatimWrap}
<think>
We have gathered broad reviews on AI in education and some mentions connecting pedagogical theories to AI paradigms. We also have meta-analytic evidence for ITS effectiveness summarized in recent reviews, but we still need: (a) clearer, recent sources on adaptive learning mechanisms like knowledge tracing; (b) specific learning analytics frameworks and teacher-facing applications; and (c) NLP-in-education evidence including automated feedback and LLM-based tutoring impacts. We have performed four searches; we can do up to four more if needed. However, current snippets include: a systematic review describing AI branches and applications in education (S_c8C9s9m); papers situating AI paradigms within behaviorist, constructivist, and connectivist frames (S_YRwbcNE, S_JCPVjqk, S_6KKZmn3); reviews summarizing ITS/adaptivity and meta-analyses with effect sizes (S_lUcsvG5, S_sUEgQk4, S_xkpGPkk); and reviews of NLP applications in language learning (S_KVwSGwG) and integrated AI+learning analytics perspectives (S_RBEFb7h). While we could add more targeted evidence on knowledge tracing and dashboards, the current evidence suffices to construct a theory-to-tool mapping and to detail features/applications and contributions across ITS, adaptive learning, learning analytics, and NLP with citations. I will structure the answer with sections: (1) Foundations: behaviorism, constructivism, connectivism and their implications for AI tools; (2) Intelligent Tutoring Systems; (3) Adaptive Learning Systems; (4) Learning Analytics; (5) NLP in Education; (6) Contributions to personalized learning, teacher efficiency, and engagement; (7) Emerging trends and considerations. I will ensure claims are supported by retrieved snippets.
</think>
\end{VerbatimWrap}
\end{tcolorbox}
\caption{\textbf{Example trajectory for OpenScholar SFT data Part III.}}
\label{fig:example_openscholar_3}
\end{figure}

\begin{figure}
\begin{tcolorbox}[colback=Chocolate!5!white,colframe=Ai2PinkLight,title=Example Trajectory for OpenScholar SFT Data Part IV]
\textbf{Assistant (Final Answer)}\\
\begin{VerbatimWrap}
<answer>
# Overview

Artificial Intelligence (AI) in education encompasses techniques such as machine learning, deep learning, and natural language processing that analyze learner data to personalize instruction, generate feedback, and support pedagogical decision-making across intelligent tutoring systems, adaptive platforms, and learning analytics tools <cite id="S_c8C9s9m">AI has many branches and sub-branches, such as machine learning, deep learning, and natural language processing; in education, AI supports intelligent tutoring systems, intelligent agents, and intelligent collaborative learning systems, and supports teachers' decision making by reporting real-time class statuses and personalized learning platforms</cite>. Recent syntheses emphasize AI’s rapid expansion into personalized feedback, chatbots, and virtual environments for engagement and co-creation, while underscoring the need for interdisciplinary integration with learning sciences <cite id="S_4yWB0UY">AI in education has evolved from computer-assisted instruction to intelligent tutoring systems and adaptive learning technologies, with key themes including personalized feedback, natural language processing, and innovative tools like chatbots and virtual reality, enhancing student engagement and facilitating knowledge co-creation</cite><cite id="S_RBEFb7h">The integration of AI, particularly through learning analytics and NLP, is revolutionizing education by enhancing personalized learning experiences and engagement, with advancements like large language models improving performance in examinations and emphasizing collaboration between AI developers and educators</cite>.

# Foundational Theories and How They Inform AI in Education
Behaviorism, cognitivism, constructivism, and connectivism offer complementary lenses for designing AI-enabled learning experiences. Behaviorism emphasizes observable behavior change through stimuli, repetition, and reinforcement, aligning with drill-and-practice, mastery learning, and immediate feedback mechanisms common in early computer-assisted instruction and some ITS components <cite id="S_6KKZmn3">The earliest learning theory, behaviorism, focuses on human behaviors based on repetition and response to stimuli, enabling learning to be observed and quantified</cite>. Constructivism positions learners as active constructors of knowledge through authentic, collaborative, and reflective activity, motivating AI tools that scaffold inquiry, support problem-based learning, and adaptively guide students through complex tasks <cite id="S_6KKZmn3">Constructivism is learner-centered, emphasizing collaborative, active, and reflective learning where students act autonomously and self-directed</cite><cite id="S_XrZdksg">Constructivism encompasses situated cognition, activity theory, experiential learning, anchored instruction, and authentic learning as frameworks for educational technology, organized under a constructivist umbrella</cite>. Connectivism reframes learning as networked, distributed across people and technologies, shaping AI designs that augment learner agency through networked resources, recommendation, and collaborative tools typical of MOOCs and social learning environments <cite id="S_JCPVjqk">Connectivism, emerging as a response to technology's role, emphasizes learning through external connections and technology performing cognitive tasks</cite><cite id="S_TzKH9RC">Connectivism is grounded in interactions within networks, incorporating aspects of behaviorism, cognitivism, and social constructivism, and has been influential as an epistemological basis for MOOCs</cite>.
... (Response truncated for brevity; see our full dataset for complete prompts and examples.)
\end{VerbatimWrap}
\end{tcolorbox}
\caption{\textbf{Example trajectory for OpenScholar SFT data Part IV.}}
\label{fig:example_openscholar_4}
\end{figure}

\begin{figure}
\begin{tcolorbox}[colback=Chocolate!5!white,colframe=Ai2PinkLight,title=Example Trajectory of DR Tulu on ResearchQA Part I]
% ...

\textbf{User}\\
\begin{VerbatimWrap}
How do the narrow and broad approaches to CSR reporting differ in terms of their scope and the stakeholders they address?

\end{VerbatimWrap}
\textbf{Assistant}\\
\begin{VerbatimWrap}
<think>
Decompose the question: We need to define and differentiate the "narrow" versus "broad" approaches to CSR reporting, specifically along two dimensions: (1) scope (what is reported) and (2) stakeholders addressed (who benefits/engages with the report). Assumptions: This terminology is used in CSR/ sustainability reporting frameworks and literature (e.g., AA1000AS, GRI, SASB, integrated reporting). Plan: 
1) Search for authoritative definitions of "narrow" vs "broad" stakeholder engagement in CSR. 
2) Find sources that explicitly define "narrow scope" CSR (e.g., compliance-only, triple bottom line metrics) vs "broad scope" (stakeholder dialogue, materiality, governance, long-term value). 
3) Retrieve evidence on who the reports address (shareholders vs a wider stakeholder community), including practitioner frameworks (e.g., AA1000AS stakeholder engagement principle). 
4) Cross-check with standards like GRI (comprehensive, wide stakeholder audience) and SASB (focused, investor-material), to map scope vs audience. 
First query: "narrow and broad approaches to CSR reporting definition stakeholders scope".
</think>
<call_tool name="google_search" num="5" gl="us" hl="en">narrow and broad approaches to CSR reporting definition stakeholders scope</call_tool>

\end{VerbatimWrap}
\textbf{Functional Calling}\\
\begin{VerbatimWrap}
<tool_output><snippet id=d7607bb2-0>
Title: Mandatory CSR and sustainability reporting: economic analysis and ...
URL: https://pmc.ncbi.nlm.nih.gov/articles/PMC8319709/
Snippet: Whether one chooses a narrow or a broad approach to CSR reporting depends on—among other things—normative views about the intended scope and target audience.
</snippet>
<snippet id=d7607bb2-1>
Title: Stakeholders, stakeholder theory and Corporate Social ...
URL: https://jcsr.springeropen.com/articles/10.1186/s40991-024-00094-y
Snippet: This paper shows how stakeholder theory proactively moderates the strength of CSR in social interactions, environmental protection, and sustainable development.
</snippet>
<snippet id=d7607bb2-2>
Title: Sustainability, corporate social responsibility, non-financial reporting ...
URL: https://www.sciencedirect.com/science/article/pii/S2352550922003116
Snippet: CSR is a corporate management approach that applies sustainability values in business to promote social welfare within a company and outside it, employs ethical ...
</snippet>
...
\end{VerbatimWrap}
\end{tcolorbox}
\caption{\textbf{Example trajectory of \modelname on ResearchQA Part I.}}
\label{fig:example_researchqa_1}
\end{figure}

\begin{figure}
\begin{tcolorbox}[colback=Chocolate!5!white,colframe=Ai2PinkLight,title=Example Trajectory of \modelname on ResearchQA Part II]
\textbf{Assistant}\\
\begin{VerbatimWrap}
<think>Goal: Define and differentiate "narrow" vs "broad" approaches to CSR reporting by (a) scope of what is reported and (b) stakeholders addressed. Strategy: 
1) Find authoritative definitions and contrasts (preferably from academic papers and reputable frameworks).
2) Use the first Google search results to identify key sources, then browse pages for precise language on scope and audience.
3) Corroborate with recognized reporting frameworks (AA1000AS stakeholder engagement principle; GRI vs SASB scope differences).
4) Ensure at least four searches: initial Google search done; next, targeted searches for AA1000AS, GRI/SASB comparison, and a paper explicitly discussing narrow vs broad CSR reporting scope.

Reflection on first search: 
- PMC article (S_d7607bb2-0) mentions that approach choice depends on intended scope and target audience—relevant framing.
- Sustainability Management Wiki (S_d7607bb2-3) claims narrow vs broad scope/target group—potentially useful but I need more authoritative backing.
- Harvard Law corporate governance post (S_d7607bb2-5) discusses CSR report informing stakeholders about addressing their concerns—useful for audience/stakeholder dialogue.
- NBER working paper (S_d7607bb2-6) gives an operational definition of narrowing scope to securities-law-required info—useful for narrow scope exemplar.

What's missing: Explicit framework language (AA1000AS principle 1: stakeholder engagement as a condition for materiality) and standard contrasts (GRI comprehensive vs SASB issue-focused/investor-oriented). Next query: "AA1000AS stakeholder engagement principle 1 materiality definition".</think>
<call_tool name="google_search" num="5" gl="us" hl="en">AA1000AS stakeholder engagement principle 1 materiality definition</call_tool>

\end{VerbatimWrap}
\textbf{Functional Calling}\\
\begin{VerbatimWrap}
<tool_output><snippet id=59a6bf6a-0>
Title: Understanding the AA1000 Assurance Standard - LinkedIn
URL: https://www.linkedin.com/pulse/understanding-aa1000-assurance-standard-enhancing-credibility-
vn1tf
Snippet: Materiality: Organizations must determine the relevance and significance of an issue to stakeholders and the business, ensuring that all ...
</snippet>
<snippet id=59a6bf6a-1>
Title: AA1000 Stakeholder Engagement Standard - Accountability.org
URL: https://accountability.org/standards/aa1000-stakeholder-engagement
Snippet: The AA1000SES sets the global standard for conducting high-quality stakeholder engagement by offering a simple, practical framework for empowering stakeholders.
</snippet>
<snippet id=59a6bf6a-2>
Title: [PDF] AA1000 STAKEHOLDER ENGAGEMENT STANDARD 2011
URL: https://cdn2.hubspot.net/hubfs/2642721/Recursos/Guias\%20y\%20Estandares/AA1000/AA1000\%20Stake
holder\%20engagement\%20standard.pdf
Snippet: Inclusivity, materiality and responsiveness require a defined process of stakeholder engagement that provides comprehensive and balanced involvement and ...
</snippet>
...
\end{VerbatimWrap}
\end{tcolorbox}
\caption{\textbf{Example trajectory of \modelname on ResearchQA Part II.}}
\label{fig:example_researchqa_2}
\end{figure}

\begin{figure}
\begin{tcolorbox}[colback=Chocolate!5!white,colframe=Ai2PinkLight,title=Example Trajectory of \modelname on ResearchQA Part III]
\textbf{Assistant (Final Answer)}\\
\begin{VerbatimWrap}
<think>
Plan: We need authoritative definitions and contrasts of "narrow" versus "broad" CSR reporting, specifically along (1) reporting scope and (2) stakeholders addressed. I will:
- Use the initial Google results to identify key academic framing (scope and audience).
- Add AA1000AS to ground a "broad" approach via stakeholder engagement and materiality.
- Use GRI vs SASB comparisons to concretely illustrate broad (comprehensive, multi-stakeholder) versus narrow (focused, investor-material) scope and audiences.
I have already run three searches and have relevant snippets: a peer‑reviewed article noting that approach choice depends on intended scope and target audience; examples of narrowing scope to legally required information; and multiple GRI–SASB sources indicating GRI's broad scope and audience versus SASB's focused/investor materiality. I will synthesize these, noting where evidence is explicit (scope and audience differences) and where it is inferential (mapping "broad" to AA1000AS stakeholder engagement/materiality). I will then produce a structured answer with citations on each claim.

Remaining gaps: A direct paper explicitly defining "narrow" vs "broad" CSR reporting terms would be ideal, but the PMC article and other sources suffice to explain the distinction in scope and audience. I will proceed to synthesize.

Structure for answer:
- Definitions and framing: what "narrow" vs "broad" mean in scope and audience.
- Scope differences: compliance/required info vs comprehensive material issues.
- Stakeholders addressed: narrow (selected investors/regulators/creditors) vs broad (all materially affected stakeholders including communities, customers, employees, suppliers, NGOs).
- Standards mapping: AA1000AS (stakeholder engagement/materiality), GRI (broad), SASB (narrow/investor materiality) as illustrative examples.

</think>
<answer>
# Short answer

- Scope: A narrow CSR reporting approach confines disclosure to a limited set of issues, often legally required or compliance-oriented; a broad approach covers a wider set of material issues identified through stakeholder engagement across economic, environmental, and social dimensions. <cite id="d7607bb2-6">One way to narrow the scope of CSR reporting is to confine it to information required by securities law.</cite> <cite id="4a55f4d8-4">GRI requires identifying material topics to establish the scope and included issues covered by a company's report.</cite> <cite id="59a6bf6a-2">AA1000's principles of inclusivity, materiality, and responsiveness require a defined process of stakeholder engagement to determine what to report.</cite>
- Stakeholders addressed: A narrow approach targets a limited audience (e.g., investors, regulators, or specific creditors) and signals compliance; a broad approach addresses a wider community of stakeholders (including employees, communities, customers, suppliers, NGOs) through dialogue and reporting on issues that materially affect them. <cite id="d7607bb2-0">Whether one chooses a narrow or broad approach depends on normative views about the intended scope and target audience.</cite> <cite id="d7607bb2-5">CSR reports can inform stakeholders about how a company has addressed their concerns, reflecting stakeholder engagement.</cite> <cite id="4a55f4d8-1">GRI has a larger scope and broader audience, while SASB enables deeper, sector‑specific information for a more focused audience.</cite>

# What "narrow" CSR reporting entails
\end{VerbatimWrap}
\end{tcolorbox}
\caption{\textbf{Example trajectory of \modelname on ResearchQA Part III.}}
\label{fig:example_researchqa_4}
\end{figure}

\begin{figure}
\begin{tcolorbox}[colback=Chocolate!5!white,colframe=Ai2PinkLight,title=Example Trajectory of \modelname on ResearchQA Part IV]
\textbf{Assistant (Final Answer)}\\
\begin{VerbatimWrap}
- Reporting scope: Restricted to a small set of issues, typically where regulation or minimal standards dictate disclosure; this constrains the breadth of topics and the depth of information. <cite id="d7607bb2-6">Confining CSR reporting to information required under securities laws is a concrete way to narrow scope.</cite>
- Stakeholders addressed: Primarily those with formal information rights or enforcement powers (e.g., investors, regulators, creditors) rather than a broad stakeholder community; the report often signals compliance rather than engaging a wider set of interest groups. <cite id="d7607bb2-0">Choice of a narrow approach is linked to a narrower target audience.</cite> <cite id="d7607bb2-5">A narrow CSR report addresses concerns raised by a more limited set of stakeholders compared to a broad report.</cite>

# What "broad" CSR reporting entails

- Reporting scope: Comprehensive coverage of material issues determined through inclusive stakeholder engagement and materiality assessments across environmental, social, and governance domains; this typically includes balanced, context‑sensitive issues with depth where material. <cite id="59a6bf6a-2">Materiality and inclusivity require a defined stakeholder engagement process to guide what is included.</cite> <cite id="4a55f4d8-4">Materiality is used to establish the report's scope and included issues.</cite> <cite id="4a55f4d8-6">GRI Standards are designed to be used by organizations in any sector, implying broader coverage.</cite>
- Stakeholders addressed: A wide range of stakeholders who can affect or are affected by the organization, including employees, communities, customers, suppliers, NGOs, and regulators, with reporting intended to inform and facilitate dialogue. <cite id="d7607bb2-0">A broad approach targets a broader audience.</cite> <cite id="d7607bb2-5">CSR reports in a broad approach inform stakeholders about how their concerns have been addressed, reflecting engagement.</cite> <cite id="4a55f4d8-2">GRI is designed to provide information to a broad audience, while SASB dives deeper for a more specific one.</cite>

# How standards illustrate the difference

- AA1000AS (Accountability Principles): Embeds materiality in stakeholder engagement (principles of inclusivity, materiality, and responsiveness), operationalizing a broad approach that defines scope through dialogue with those affected and allocating limited resources to material topics. <cite id="59a6bf6a-2">AA1000 emphasizes inclusivity, materiality, and responsiveness via structured stakeholder engagement.</cite> <cite id="59a6bf6a-8">A material topic is one that will substantively influence assessments, decisions, actions, and performance of the organization and/or stakeholders.</cite>
- GRI: Broad, global framework that covers a wide range of sustainability topics and guides scope via materiality assessments, addressing diverse stakeholder groups. <cite id="4a55f4d8-6">GRI Standards are sector‑applicable and designed for broad use, covering a wide range of sustainability topics.</cite> <cite id="4a55f4d8-4">GRI requires identifying material topics to establish scope.</cite> <cite id="4a55f4d8-1">GRI's scope is larger and intended for a broader audience than SASB.</cite>
- SASB: Narrower, investor‑oriented framework that focuses on industry‑specific, financially material issues, enabling deeper coverage for a focused audience (primarily investors). <cite id="4a55f4d8-1">SASB enables deeper, sector‑specific information for a more focused audience than GRI.</cite> <cite id="4a55f4d8-8">SASB is focused on financially material ESG issues in specific industries, i.e., a narrower scope and audience.</cite>
… (Response truncated for brevity; see our full example in our static demo.)
\end{VerbatimWrap}
\end{tcolorbox}
\caption{\textbf{Example trajectory of \modelname on ResearchQA Part IV.}}
\label{fig:example_researchqa_5}
\end{figure}

\end{document}